\RequirePackage{latexml}
\documentclass[10pt,a4paper]{article}
\iflatexml
  \usepackage{amsmath,amssymb,graphicx,booktabs,array,tabularx}
\usepackage[authoryear,round]{natbib}
\usepackage{hyperref}
\hypersetup{hidelinks}

\else
  \usepackage[a4paper,margin=25mm]{geometry}
\usepackage[T1]{fontenc}
\usepackage{tgpagella,inconsolata}
\usepackage{amsmath,amssymb,graphicx,booktabs,array,tabularx}
\usepackage[authoryear,round]{natbib}
\usepackage{microtype,xurl,hyperref,placeins,needspace}
\usepackage[font=small,labelfont=bf,skip=5pt]{caption}
\hypersetup{colorlinks=true,citecolor=blue,linkcolor=black,urlcolor=black,pdftitle={OceanCoder: A Verification-Guided Large Language Model for Automated Simulation Modelling in Ocean Engineering},pdfauthor={Can Xue, Changhong He, Ziyi Sun, Anliang Ge, Shimin Yu, Bingchen Liang}}
\fi
\begin{document}
\iflatexml
  \title{\textbf{OceanCoder: A Verification-Guided Large Language Model\\for Automated Simulation Modelling in Ocean Engineering}}
\date{}
\maketitle
\begin{center}
\begin{tabular}{ccc}
\textbf{Can Xue}\textsuperscript{\textdagger} &
\textbf{Changhong He}\textsuperscript{\textdagger,\textsection} &
\textbf{Ziyi Sun}\\
Ocean University of China & Ocean University of China & Ocean University of China\\
& Beihang University & \\[1em]
\textbf{Anliang Ge}\textsuperscript{*} &
\textbf{Shimin Yu}\textsuperscript{*} &
\textbf{Bingchen Liang}\\
Ocean University of China & Ocean University of China & Ocean University of China
\end{tabular}

\textsuperscript{\textdagger}Equal contribution\quad
\textsuperscript{\textsection}Project leader\quad
\textsuperscript{*}Corresponding authors
\end{center}

\section*{Abstract}
Numerical simulation is important in ocean engineering, but model construction requires expert knowledge. Large language models support automated modelling, but code syntax and successful execution alone do not establish whether generated models and results meet engineering requirements. We present OceanCoder, a Qwen3-8B-based large language model for automated simulation modelling in ocean engineering. To address specialised programming requirements and scarce training data, we develop an extensible data construction pipeline producing 12,000 verified supervised fine-tuning (SFT) examples across 12 engineering task categories. A three-layer verification framework checks model inputs against engineering parameters and task constraints, verifies completion of required simulation analyses, and assesses the engineering plausibility of computed outputs. Its outcomes provide graded rewards for group relative policy optimisation (GRPO), guiding the model towards engineering task completion. We also construct the first benchmark for automated modelling in ocean engineering, covering out-of-distribution tasks and real-world engineering cases. Experiments and ablation studies demonstrate the effectiveness of the data pipeline, verification framework and reward design. OceanCoder outperforms both base models and the closed-source comparison, with Pass@1 increasing from 0\% for the base model to 46.77\% after SFT and 51.43\% after GRPO. These results support engineering verification as a training signal for automated simulation modelling.

\noindent\textbf{Keywords:} Ocean engineering; Automated simulation modelling; Large language models; Engineering verification; Reinforcement learning.\par

  \section{Introduction}
\label{sec:introduction}

Numerical simulation helps engineers assess loads and motions in mooring systems, marine risers and subsea installation operations \citep{Hall2015MooringValidation,Jae2022MooOpT4FPSO,Kallinger2026HolisticMooring}. Building these models still requires specialist knowledge. Users must define component connections and physical properties, set environmental conditions, and choose the analysis steps and results to extract. A change in structural configuration or operating sequence can require corresponding changes to the program, even when an existing script is available. Generating simulation programs from engineering task descriptions could reduce this preparation effort and make specialised software easier to use.

Large language models (LLMs) offer a way to translate such descriptions into programs. Studies in ship design, finite-element analysis and computational fluid dynamics (CFD) show how natural-language requirements can be converted into design parameters or simulation workflows \citep{Choi2026LLMasUI,Hou2025AutoFEA,Fan2026ChatCFD}. However, an executable program may still perform a different analysis from the one requested. A solver can run successfully after a program has changed a line property or extracted the wrong response quantity. Assessing automated modelling therefore requires checking whether the model and its results meet the engineering requirements, in addition to checking code syntax and execution.

We present OceanCoder, a Qwen3-8B-based large language model for automated simulation modelling in ocean engineering. OceanCoder takes a natural-language description of engineering conditions and analysis requirements and generates a complete OrcaFlex program to build the model, run the analyses and extract the results. We develop this capability by first training on verified examples and then using engineering verification of generated programs to provide further training feedback.

Training OceanCoder requires paired examples of an engineering task and a program that correctly implements it. Such public data for specialised ocean engineering simulation software are scarce. Seed-based generation can expand training examples \citep{Wei2024Magicoder,Wei2024SelfCodeAlign}, but engineering parameters must remain consistent. For example, water depth, endpoint coordinates and line lengths in a catenary model must describe a compatible geometry. We therefore build a data construction pipeline that expands expert-designed seed cases within engineering constraints and generates each task description and reference program from the same physical parameter record. Each pair is verified before entering the training corpus. The resulting corpus contains 12,000 supervised fine-tuning (SFT) examples across 12 categories of static analysis, dynamic response and offshore operations. The corpus can be extended by adding seed cases with their parameter constraints and verification rules.

Training on verified examples cannot prevent all mistakes in new programs. In Figure~\ref{fig:motivation}, an SFT-generated lowering program changes the prescribed line properties. It completes the simulation, but its error in predicting the first seabed-contact time exceeds the task tolerance. To identify errors of this kind, we introduce a three-layer engineering verification framework covering inputs, processes and outputs. It checks the constructed model against the given parameters and task constraints, confirms that the required analyses have been completed, and checks the reported response quantities and their numerical values against task-specific criteria. These checks assess whether the computed results are consistent with the specified engineering problem. A successful run is accepted as a completed task only when it passes the engineering checks.

The verification results also provide feedback for improving the model. Earlier code-generation studies have used program outcomes and execution feedback as reinforcement-learning signals \citep{Le2022CodeRL,Gehring2025RLEF}. After SFT, we use the engineering verification outcomes as rewards for online group relative policy optimisation (GRPO). Programs that fail to run, run but fail engineering checks, or complete the task receive different rewards. This graded feedback retains information about partial progress while reserving full reward for complete task acceptance. During training, generated programs run in OrcaFlex, and their verification results supply the rewards used to update the model.

We also construct the first benchmark for automated modelling in ocean engineering, covering out-of-distribution tasks and real-world engineering cases to assess generalisation and practical applicability. Its 665 tasks cover core parameter variations, structural variations and engineering cases adapted from public project descriptions and technical materials. Each model generates one program per task without using simulator feedback to repair it. All programs are evaluated using the same engineering acceptance criteria. This protocol assesses the modelling capability acquired through training. The benchmark and its evaluation resources will be released together.

The experiments compare OceanCoder with base and leading closed-source models under this protocol. Independent manual audits compare the verifier's decisions with human inspection of generated programs and results. Controlled reward ablations examine the contribution of intermediate verification feedback to training. OceanCoder outperforms the evaluated comparison models, with the proportion of tasks completed on the first attempt (Pass@1) increasing from 46.77\% after SFT to 51.43\% after GRPO.

\par\begingroup
\refstepcounter{figure}
\begin{center}
\includegraphics[width=\linewidth,alt={Reference and SFT-generated lowering models with different line properties and seabed-contact times.}]{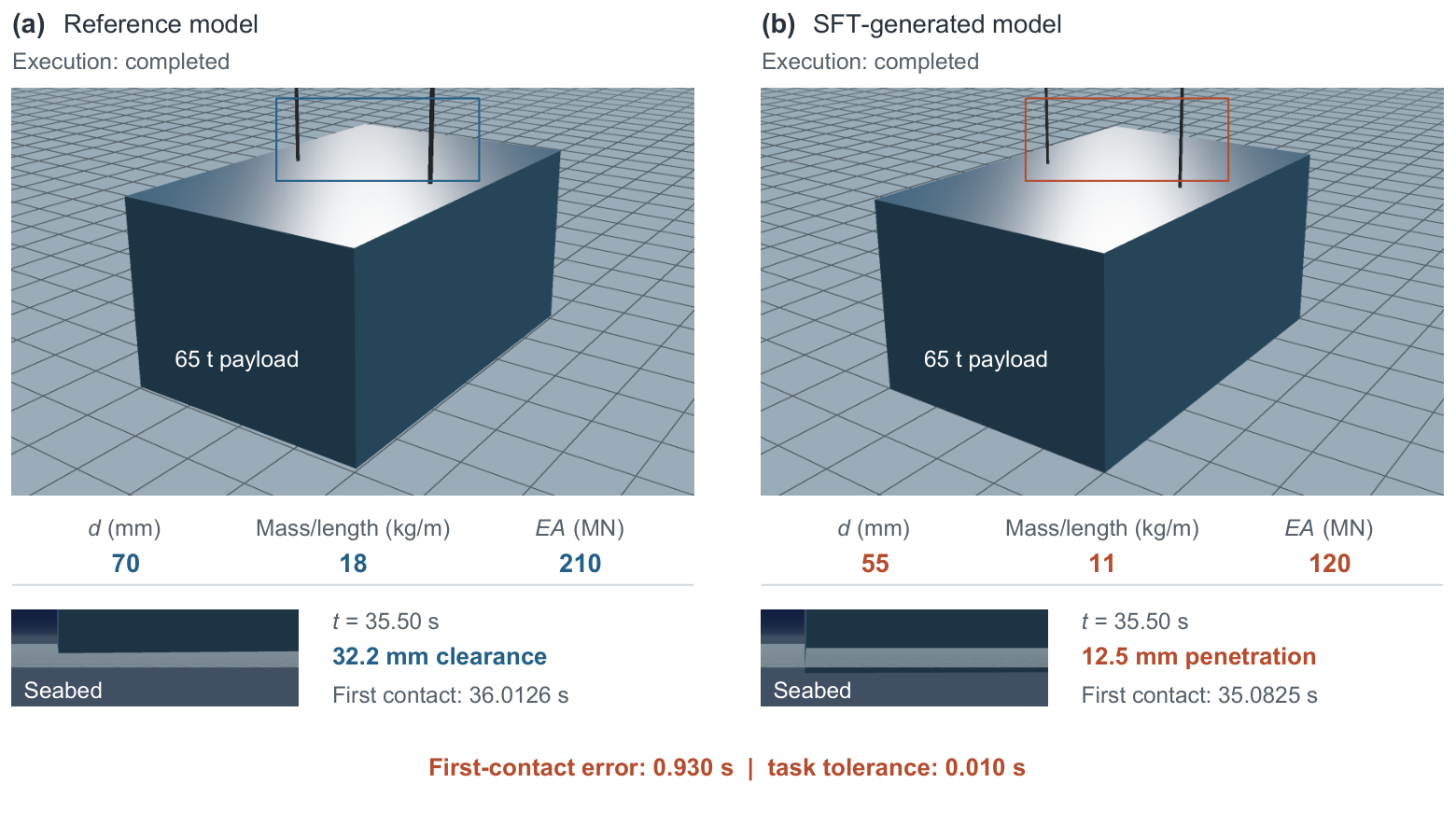}
\end{center}

\noindent\textbf{Figure~\thefigure:} The verification gap between executable code and complete engineering task acceptance. In a two-point lowering task involving a 65~t payload, the reference and OceanCoder programs after SFT both execute, but their line properties and seabed-contact responses differ. At $t=35.50$~s, the reference retains 32.2~mm clearance while the SFT model has 12.5~mm penetration. The predicted first-contact times differ by 0.9301~s, exceeding the 0.010~s task tolerance. Here, $d$ denotes line diameter and $EA$ axial stiffness; the reference is a checked simulation solution.\par

\label{fig:motivation}
\par\endgroup

The main contributions are as follows.
\begin{enumerate}
\item \textbf{OceanCoder for automated simulation modelling.} A Qwen3-8B-based model translates natural-language engineering requirements into complete OrcaFlex programs for model construction, analysis execution and result extraction.
\item \textbf{Verified and extensible training data.} A seed-based pipeline preserves engineering parameter relationships and agreement between task descriptions and reference programs, providing 12,000 verified examples across 12 task categories.
\item \textbf{Engineering verification as training feedback.} A framework checks model inputs, analysis completion and computed outputs, and turns the outcomes into graded GRPO rewards that guide model training.
\item \textbf{Benchmark and evaluation protocol.} A 665-task benchmark applies common engineering checks to all evaluated models. Manual audits and controlled reward ablations assess verifier reliability and the contribution of training feedback.
\end{enumerate}

\section{Related work}
\label{sec:related_work}

\subsection{Large language models for engineering simulation}
\label{subsec:rw_llm_sim}

Automating engineering simulation involves translating natural-language requirements into solver inputs and coordinating numerical analyses. Language models support these tasks through structured interfaces, retrieval and agent coordination. LLM-as-UI converts user requests into parameters for modular ship design \citep{Choi2026LLMasUI}. AutoFEA combines planning and example retrieval to generate finite element solver inputs \citep{Hou2025AutoFEA}. For computational fluid dynamics (CFD), OpenFOAMGPT and ChatCFD use domain knowledge and iterative correction to configure simulation cases \citep{Pandey2025OpenFOAMGPT,Fan2026ChatCFD}. MetaOpenFOAM 2.0 and Foam-Agent organise simulation and post-processing through task decomposition and agent coordination \citep{Chen2025MetaOpenFOAM2,Yue2026FoamAgent}. LLM-based residual dynamics also augment marine-vessel predictive control \citep{Zeng2026LLMMPC}. These studies assign language models specific roles in interpreting requirements, constructing cases and supporting engineering decisions.

Domain-specific supervision supports different engineering tasks. OceanGPT uses ocean-domain instruction data \citep{Bi2024OceanGPT}. ALL-FEM fine-tunes models on verified FEniCS scripts for finite element workflows \citep{Deotale2026ALLFEM}. Simulation-generated data also train neural networks to predict riser fatigue \citep{Sivaprasad2023RiserFatigue}. Program generation requires reliable pairs of engineering requirements and executable implementations. Self-Instruct generates and filters instructions, Magicoder constructs tasks from source-code snippets, and SelfCodeAlign selects generated solutions through execution tests \citep{Wang2023SelfInstruct,Wei2024Magicoder,Wei2024SelfCodeAlign}. For engineering simulation, execution testing must be accompanied by consistency between physical parameters, task descriptions and program implementations. When a task contains coupled parameters, independent changes to the description or program can make an otherwise executable example inconsistent with the specified engineering problem. To preserve this consistency during data expansion, OceanCoder generates descriptions and reference OrcaFlex programs from shared physical parameter records, subject to engineering constraints.

\subsection{Engineering verification and learning from feedback}
\label{subsec:rw_verification_rl}

Successful execution does not establish whether a program satisfies its intended requirements. The test-oracle problem concerns how to judge the correctness of observed program behaviour \citep{Barr2015OracleProblem}. EvalPlus shows that expanded tests detect functional errors missed by the original test suites \citep{Liu2023EvalPlus}. In optimisation modelling, even agreement with a reference objective value can conceal an incorrect constraint set. Constraint injection addresses this problem by checking feasible and constraint-violating solutions alongside objective values \citep{Luo2026ConstraintInjection}. Computational verification addresses both code implementation and numerical-solution accuracy \citep{Oberkampf2002VV}. These distinctions motivate checking the engineering specification as well as program behaviour.

The same verification criteria can guide learning as well as evaluation. Constraint injection uses verification outcomes for synthetic-data filtering and reinforcement-learning rewards \citep{Luo2026ConstraintInjection}. CodeRL combines unit-test outcomes, graded rewards and a learned critic \citep{Le2022CodeRL}. RLEF trains models to use execution feedback during iterative refinement \citep{Gehring2025RLEF}. CODERL+ uses variable-level execution feedback \citep{Jiang2026CodeRLPlus}. OraclePhys examines how different forms of finite-element-based supervision and verification feedback affect learning in structural mechanics \citep{Li2026OraclePhys}. The appropriate feedback depends on the task. Unit-test outcomes, formulation constraints and structural responses each assess different requirements. For OrcaFlex modelling, the relevant checks concern the constructed physical system, completion of the required analyses and extraction of the requested responses. OceanCoder's three-layer verification framework performs these checks and distinguishes execution progress from complete task acceptance. The resulting verification states provide graded rewards for online GRPO, linking model updates to engineering task completion.

\subsection{Benchmarks for automated simulation modelling}
\label{subsec:rw_benchmarks}

Benchmarks differ in the tasks they assign and the evidence they use to judge success. DS-1000 evaluates programming with scientific Python libraries, and BigCodeBench tests diverse function calls and complex instructions \citep{Lai2023DS1000,Zhuo2025BigCodeBench}. SciCode assesses research programming problems curated by scientists \citep{Tian2024SciCode}. Simulation benchmarks extend this evaluation to numerical software and its outputs. FEABench checks COMSOL-based multiphysics models, physical settings and numerical results \citep{Mudur2025FEABench}. CFDLLMBench covers CFD knowledge, solver implementation and OpenFOAM workflows \citep{Somasekharan2026CFDLLMBench}. SimulCost evaluates simulation parameter tuning under accuracy and computational-cost constraints \citep{Cao2026SimulCost}. LiveCodeBench additionally evaluates code generation, self-repair, execution and output prediction using time-segmented problems \citep{Jain2025LiveCodeBench}. These settings assess distinct programming and simulation requirements.

Interaction budgets further affect what a benchmark measures. A single-attempt setting assesses the initial response, whereas feedback-based settings also assess error diagnosis and correction. To measure the modelling capability acquired through training, the OceanCoder benchmark fixes the interaction budget. Each model makes one attempt to generate a complete OrcaFlex program for each task, without simulator-feedback repair. During evaluation, verification outcomes determine acceptance without being returned to the model for another generation attempt. Its 665 tasks cover parameter variations, structural variations and engineering contexts. Common checks of model inputs, analysis completion and requested outputs define acceptance for every model. The resulting comparisons separate performance across task variations while retaining a consistent definition of engineering task completion.

\section{OceanCoder framework and training procedure}
\subsection{Task definition and framework overview}
\label{subsec:framework_overview}

\par\begingroup
\refstepcounter{figure}
\begin{center}
\includegraphics[width=\linewidth,alt={OceanCoder data construction, supervised fine-tuning, verification-guided GRPO, and independent benchmark evaluation.}]{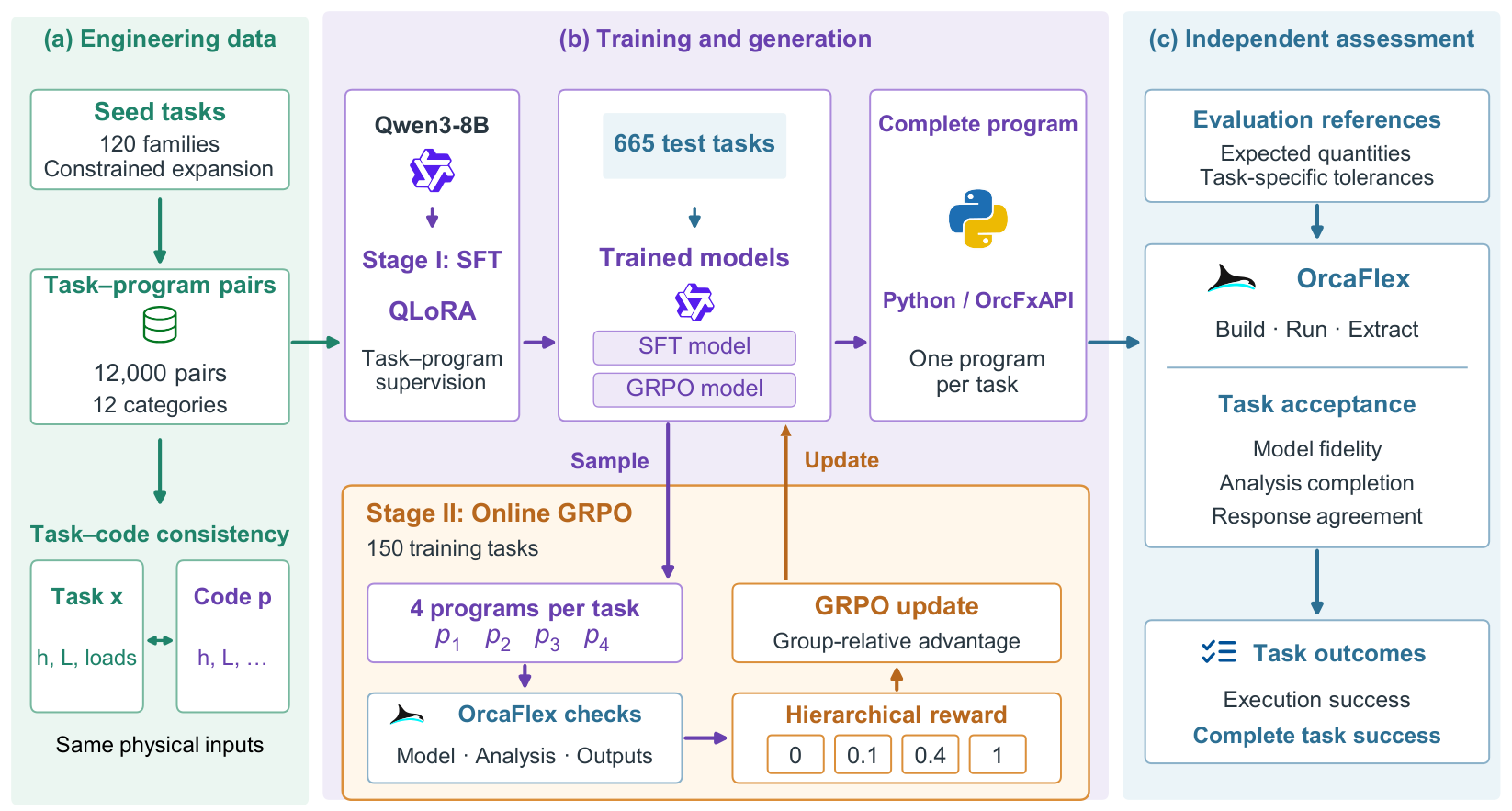}
\end{center}

\noindent\textbf{Figure~\thefigure:} Overview of the OceanCoder framework for engineering-constrained data construction, supervised fine-tuning, online GRPO and independent evaluation.\par

\label{fig:framework}
\par\endgroup

OceanCoder addresses the task of generating complete OrcaFlex programs from natural-language specifications of physical systems, required analyses and requested responses. Each program uses OrcFxAPI, the Python interface to OrcaFlex, to build the model, run the analyses and extract the results \citep{OrcinaPythonInterface}. Task completion requires agreement with the engineering specification throughout this process, including the model configuration, analysis steps and reported responses.

Figure~\ref{fig:framework} presents the data construction, training and evaluation framework. We expand expert-designed seed cases within engineering constraints and use the same parameters to generate each task description and reference program. Verified pairs are used for supervised fine-tuning (SFT) of Qwen3-8B. Since programs generated after SFT can still alter specified inputs or extract unintended responses, subsequent training uses engineering verification of generated programs as feedback. Three-layer verification checks model inputs, analysis completion and outputs. The verification outcomes determine graded rewards for online group relative policy optimisation (GRPO). The trained OceanCoder checkpoints and comparison models are evaluated on a separate benchmark using the same engineering acceptance criteria. Each model makes one attempt to generate a complete program for each task, without simulator-feedback repair.

\subsection{Engineering-constrained data construction}
\label{subsec:data_construction}

We construct training examples by expanding expert-designed seed cases, as shown in Figure~\ref{fig:data_pipeline}. A seed defines the physical objects and their connections, the analysis procedure and the response quantities to extract. These elements establish the task family within which additional cases are generated. The corpus covers 12 categories grouped into static analysis, dynamic response and offshore operations, as listed in Table~\ref{tab:task_categories}. Each category contains ten seed families, with 100 expanded examples per family, producing 1,000 examples per category and 12,000 examples in total. The seed therefore provides both an engineering specification and a complete program structure that can implement admissible variations of that specification.

\begin{table}[htbp]
\caption{Task coverage of the supervised training corpus (1,000 pairs per category; 12,000 pairs across 12 categories). SCR denotes a steel catenary riser.}

\label{tab:task_categories}
 \centering
\begin{tabularx}{\linewidth}{@{}l l >{\raggedright\arraybackslash}X@{}}
\toprule
Group & ID & Engineering task \\
\midrule
Static & A1 & Single-catenary static equilibrium \\
analysis & A2 & Four-point mooring static equilibrium \\
 & A3 & SCR static configuration and touchdown \\
 & A4 & Pipeline lateral stability \\
\midrule
Dynamic & B1 & Single catenary under regular waves \\
response & B2 & Coupled four-point mooring dynamics \\
 & B3 & SCR response to motion, waves and current \\
 & B4 & Free-span pipeline wave--current response \\
\midrule
Offshore & C1 & Static S-lay pipeline installation \\
operations & C2 & Dynamic laying of a flexible cable \\
 & C3 & Single-point lifting and lowering \\
 & C4 & Two-point lowering and seabed contact \\
\bottomrule
\end{tabularx}
\par
\end{table}

\par\begingroup
\refstepcounter{figure}
\begin{center}
\includegraphics[width=\linewidth,alt={Engineering-constrained expansion of seed cases into verified task descriptions and reference programs.}]{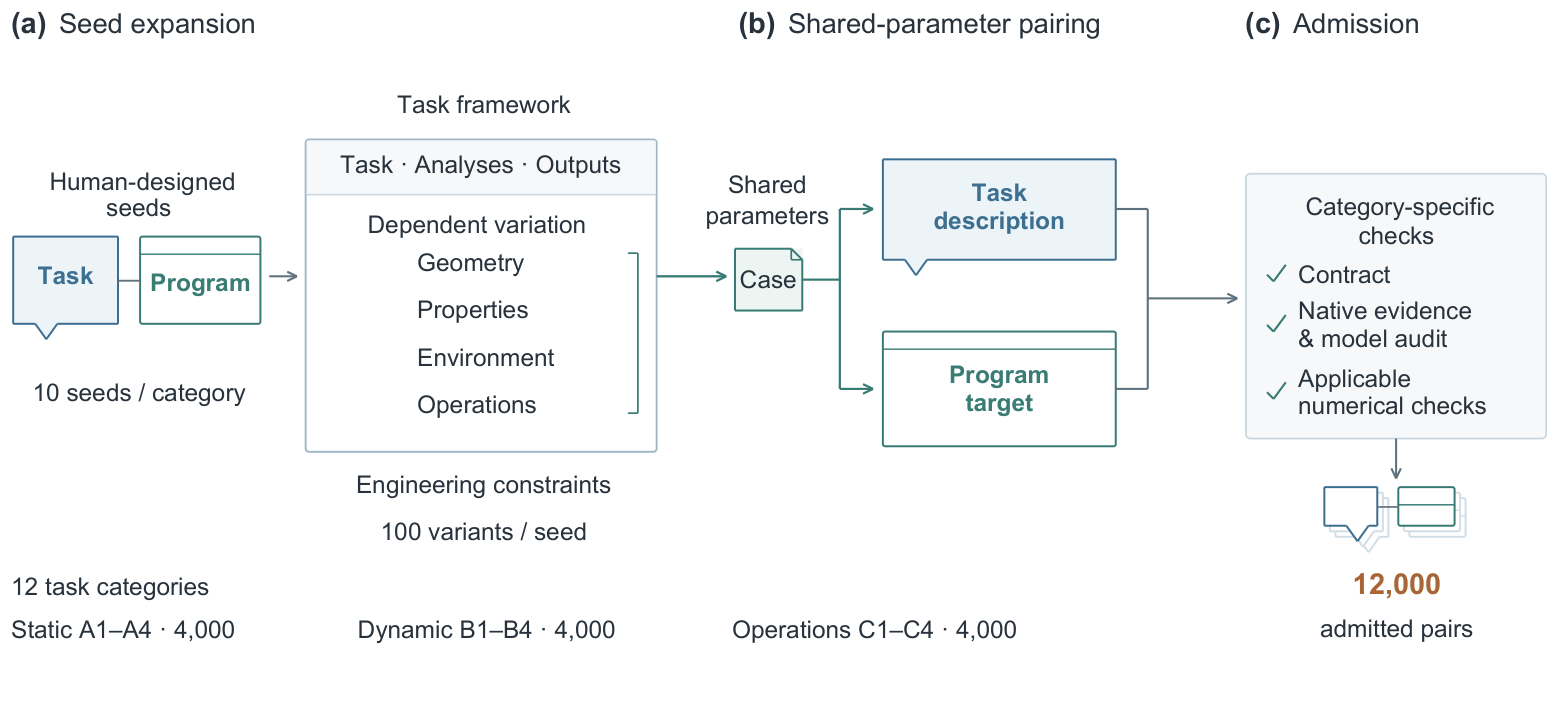}
\end{center}

\noindent\textbf{Figure~\thefigure:} Engineering-constrained data construction pipeline for expanding expert-designed seed cases, generating task descriptions and reference programs from shared parameters, and verifying examples before training.\par

\label{fig:data_pipeline}
\par\endgroup

Parameter variation preserves the relationships required by each task. For a catenary case, water depth, endpoint positions and line length jointly determine the geometry to be modelled. Line diameter, unit mass and axial stiffness follow the combinations allowed by the seed rules. For a lowering task, the payout rate and stage durations must agree with the prescribed operation. These dependencies restrict which parameter combinations can be used together. The ranges in Table~\ref{tab:parameter_ranges} describe the marginal coverage of the resulting corpus. They are obtained from constrained cases and coupled component-property packages.

A shared parameter record connects each expanded specification to its reference implementation. Once a case is selected, the same values populate the task description and the corresponding model assignments in the program. The expanded A1 case in Table~\ref{tab:a1_example}, for example, specifies a water depth of 100~m and a line length of 200~m. These values appear in the description and are assigned to the native OrcaFlex model. The record also determines the endpoint coordinates, line properties and contact settings. The analysis and extraction steps retain the seed's task definitions, including the requested response quantities and their locations. This construction keeps the language requirement, physical model and output requirements consistent as the parameters change.

\begin{table}[htbp]
\caption{Marginal ranges of principal physical, environmental and operational parameters across the 12,000-pair SFT training corpus (1,000 pairs per category). Ranges reflect constrained within-category variations and coupled component property packages rather than independent Cartesian combinations.}

\label{tab:parameter_ranges}
\begingroup
 \centering
\begin{tabularx}{\linewidth}{@{}l l r >{\raggedright\arraybackslash}X@{}}
\toprule
Input group & Parameter (unit) & Corpus range & Engineering scope and coupling constraints \\
\midrule
Geometry & Water depth, $h$ ($\mathrm{m}$) & 15--1,975 & Shallow-to-intermediate pipelines, moorings and installation tasks (15--380~m); deepwater steel catenary risers (450--1,975~m). \\
 & Total line / pipeline length, $L$ ($\mathrm{m}$) & 45--3,015 & Scaled with water depth and hang-off geometry across lifting slings (45--75~m), catenary moorings and pipelay (130--1,031~m) and deepwater SCRs (920--3,015~m). \\
Structure & Outer / nominal diameter, $d$ ($\mathrm{mm}$) & 44--610 & Coupled within component property packages: hoisting wires and subsea cables (44--160~mm), mooring chains and ropes (70--360~mm) and offshore steel pipelines/risers (219.1--610~mm). \\
 & Bare unit mass, $m_{\mathrm{bare}}$ ($\mathrm{kg/m}$) & 6.5--366.2 & Bare structural mass across cables (6.5--60~kg/m), mooring lines (52--286.6~kg/m) and bare steel pipes (70.8--366.2~kg/m); in coated pipeline stability tasks (A4), adding concrete weight coating yields total coated mass $m_{\mathrm{coated}}$ of 109.7--1,393.7~kg/m. \\
 & Axial stiffness, $EA$ ($\mathrm{MN}$) & 60--10,049 & Assigned by structural member type to preserve realistic axial elongation across slings and cables (60--400~MN), mooring lines (82--1,674~MN) and thick-walled steel pipes (1,894--10,049~MN). \\
Environment & Wave height, $H$ / $H_s$ ($\mathrm{m}$) & 0--8.0 & Still water (0~m) in static equilibrium and controlled lowering analyses; regular waves (1.08--8.0~m) and irregular JONSWAP seas ($H_s=2.52$--3.30~m) in dynamic tasks. \\
 & Wave period, $T$ / $T_p$ ($\mathrm{s}$) & 6.5--15.0 & Regular wave periods (6.5--15.0~s) and spectral peak periods ($T_p=10$--11~s) paired with wave height and water depth to maintain realistic wave steepness. \\
 & Current speed, $V_c$ ($\mathrm{m/s}$) & 0.00--1.20 & Zero in still-water analyses; steady currents of 0.20--1.20~m/s across incident directions ($0^\circ$--$355^\circ$) in combined wave--current loading scenarios. \\
Seabed \& ops & Seabed normal stiffness, $k_n$ ($\mathrm{kN/m^3}$) & 400--7,792 & Linear elastic seabed contact stiffness (with Coulomb friction coefficients of 0.20--0.69), complemented by nonlinear soil resistance models in deepwater riser touchdown analyses. \\
 & Payloads, tensions \& speeds & Task-specific & Lifted payloads (30--115~t), moored displacement vessels (1,170--15,970~t), S-lay tensioner capacities (120--850~kN) and winch/lay speeds (0.03--0.89~m/s) matched to stage durations. \\
\bottomrule
\end{tabularx}
\par
\endgroup
\end{table}

\begin{table}[htbp]
\caption{Comparison of an original A1 seed (\texttt{seed\_01}) and an expanded supervised training instance (\texttt{A1\_seed\_01\_v001}). English summaries retain the exact physical inputs of the Chinese task statements.}

\label{tab:a1_example}
\begingroup
 \centering
\begin{tabularx}{\linewidth}{@{}l >{\raggedright\arraybackslash}X >{\raggedright\arraybackslash}X@{}}
\toprule
Component & Original seed: A1, \texttt{seed\_01} & Expanded pair: \texttt{A1\_seed\_01\_v001} \\
\midrule
Task & Construct one catenary, solve static equilibrium, save the model and extract the specified responses. & The same task family and analysis and output requirements, with expanded physical inputs. \\
Geometry & Water depth 80 m; fixed A at $(0,0,-8)$ m; anchored B at $(110,0)$ m and zero height above the seabed; line length 145 m. & Water depth 100 m; fixed A at $(0,0,-12)$ m; anchored B at $(145,0)$ m and zero height above the seabed; line length 200 m. \\
Line properties & Equivalent outer diameter 90 mm; mass 95 kg/m; axial stiffness 800 MN. & Equivalent outer diameter 160.7 mm; mass 158.2 kg/m; axial stiffness 729.7 MN. \\
Contact and mesh & Default contact diameter (inherited from outer diameter) and standard axial compression; target segment length 2 m. & Parameterized contact diameter (306.0 mm) and compression-limit switch activated within the seed's admissible parameter space; target segment length 2 m. \\
Common conditions & \multicolumn{2}{p{0.77\linewidth}@{}}{Still water, zero current, flat elastic seabed and normal stiffness $1000\ \mathrm{kN/m^3}$. Unspecified inputs retain the stated empty-model defaults.} \\
Program excerpts & {\ttfamily model.environment.WaterDepth = 80.0\par
line.Length = (145.000,)\par
model.CalculateStatics()} & {\ttfamily model.environment.WaterDepth = 100.0\par
line.Length = (200.0,)\par
model.CalculateStatics()} \\
Outputs & \multicolumn{2}{p{0.77\linewidth}@{}}{\texttt{model.dat} and \texttt{results.json}: end effective tensions; minimum elevation and its arc length; maximum curvature and its arc length; model state and warnings.} \\
Verified response & Static equilibrium solved and verified via model reload (73 nodes; top effective tension 103.144 kN; touchdown contact confirmed on the elastic seabed). & Static equilibrium solved and verified via model reload (101 nodes; top effective tension 152.703 kN; touchdown contact confirmed on the elastic seabed). \\
\bottomrule
\end{tabularx}
\par
\endgroup
\end{table}

Each pair undergoes verification before admission to the corpus. Static checks use the program's abstract syntax tree (AST) to inspect its structure and required operations. Native OrcaFlex execution then establishes whether the prescribed analyses complete and the required files are produced. The saved model is reloaded so that its actual settings can be compared with the task specification. Applicable responses are independently recomputed from the saved results and checked against the reported outputs. These stages inspect complementary evidence about the program, the model it constructs and the quantities it reports. Appendix~\ref{app:admission_verification} details the checks for each category.

An admitted example contains a system instruction, a natural-language task description and a complete Python program. The system instruction specifies the program format and the execution and output requirements. Its seed family and parameter values are retained to document its construction, and Appendix~\ref{app:training_records} summarises the corpus composition. Additional task categories can follow the same procedure once their seed cases, parameter constraints and verification rules have been defined.

\subsection{Stage I: supervised fine-tuning}
\label{subsec:sft}

We fine-tune Qwen3-8B \citep{Yang2025Qwen3} on the verified task--program pairs to establish its ability to generate complete OrcaFlex programs. Each target response includes model construction, analysis execution and result extraction. Training minimises the negative log-likelihood of the reference program tokens, averaged over the supervised response tokens. System instructions, user requirements and padding tokens are excluded from the loss. This supervision trains the model to generate programs that implement the specified model configuration, analysis steps and result extraction.

Fine-tuning uses quantised low-rank adaptation (QLoRA) with 4-bit NormalFloat (NF4) quantisation \citep{Dettmers2023QLoRA}. Low-rank adapters in the attention and feed-forward layers are trained while the base weights remain frozen \citep{Hu2022LoRA}. The selected SFT checkpoint initialises GRPO. A separate copy of this checkpoint, including its trained adapters, remains fixed and supplies the reference token probabilities used to compute the Kullback--Leibler (KL) penalty in Section~\ref{subsec:grpo} (see Appendix~\ref{app:executed_hyperparameters} for training logs and compute resources).

\subsection{Three-layer engineering verification}
\label{subsec:verification}

Engineering verification determines whether a generated program fulfils its task by examining the saved OrcaFlex model, analysis records and numerical outputs. The input layer checks the physical system constructed by the program. Geometry, component properties, object connections, environmental loads and boundary conditions are compared with the engineering specification. The saved model records the settings that the solver actually used. Its inspection covers assignments to individual line segments, connection positions and restraint settings, as well as properties shared across a complete component. This allows the checks to distinguish a valid calculation from a calculation performed with an incorrectly configured system.

The process layer establishes whether the required analyses have been completed. Static tasks require a solved equilibrium state. Dynamic tasks additionally require completion of the prescribed simulation period, while operational tasks require the specified sequence of analysis stages and motions. The verifier uses execution records and saved analysis states to establish completion. This connects the analysis requirement to the actual solution state, including tasks in which several stages must be completed before a requested response is available.

The output layer checks both the meaning and the numerical value of each response. A result must represent the requested physical quantity at the specified spatial position and over the prescribed time interval. For example, a directional restoring force requires the correct projection and sign, and a line response requires the correct arc-length location. The verifier recomputes applicable quantities from the saved results before comparing them with the reference values. A correctly configured model can therefore still fail acceptance if its program reports the wrong quantity or extracts a response from an unintended location or interval.

Numerical comparison accounts for equivalent representations and numerical precision. Periodic angles are normalised before comparison, and non-physical identifiers are excluded from numerical checks. Thus, equivalent directions around the $0^\circ/360^\circ$ boundary and different metadata labels can be handled consistently. Absolute or relative tolerances are specified for individual response quantities. Where the task specification includes an absolute threshold near zero, that threshold governs comparisons of small values. Mesh and time-step sensitivity checks on the seed cases inform the selected tolerances, as documented in Appendix~\ref{app:seed_examples}. Complete acceptance requires all applicable input, process and output checks to pass. This joint decision preserves the relationship between the prescribed physical problem and the reported response.

The lowering task in Figure~\ref{fig:motivation} illustrates how the layers work together. Input checks examine the prescribed hoisting-line properties and connections, while process checks establish completion of the lowering sequence. First seabed contact is identified from the geometric zero crossing and confirmed by contact force over a 0.05~s window. The peak upward contact force is evaluated within the following two-second window. The first-contact time is compared with the reference using an absolute tolerance of 0.010~s, while peak contact force and penetration use the task's 5\% relative tolerance. The SFT program has a first-contact-time error of 0.9301~s, exceeding the prescribed tolerance despite successful execution. These checks connect the specified model to the analysis performed and the responses reported. The engineering acceptance criteria are used for GRPO feedback and benchmark evaluation. Verification of reference training pairs before SFT is described separately in Section~\ref{subsec:data_construction}. Category-specific checks are detailed in Appendix~\ref{app:admission_verification}.

\par\begingroup
\refstepcounter{figure}
\begin{center}
\includegraphics[width=\linewidth,alt={Four candidate programs are executed and verified to supply graded rewards for GRPO updates.}]{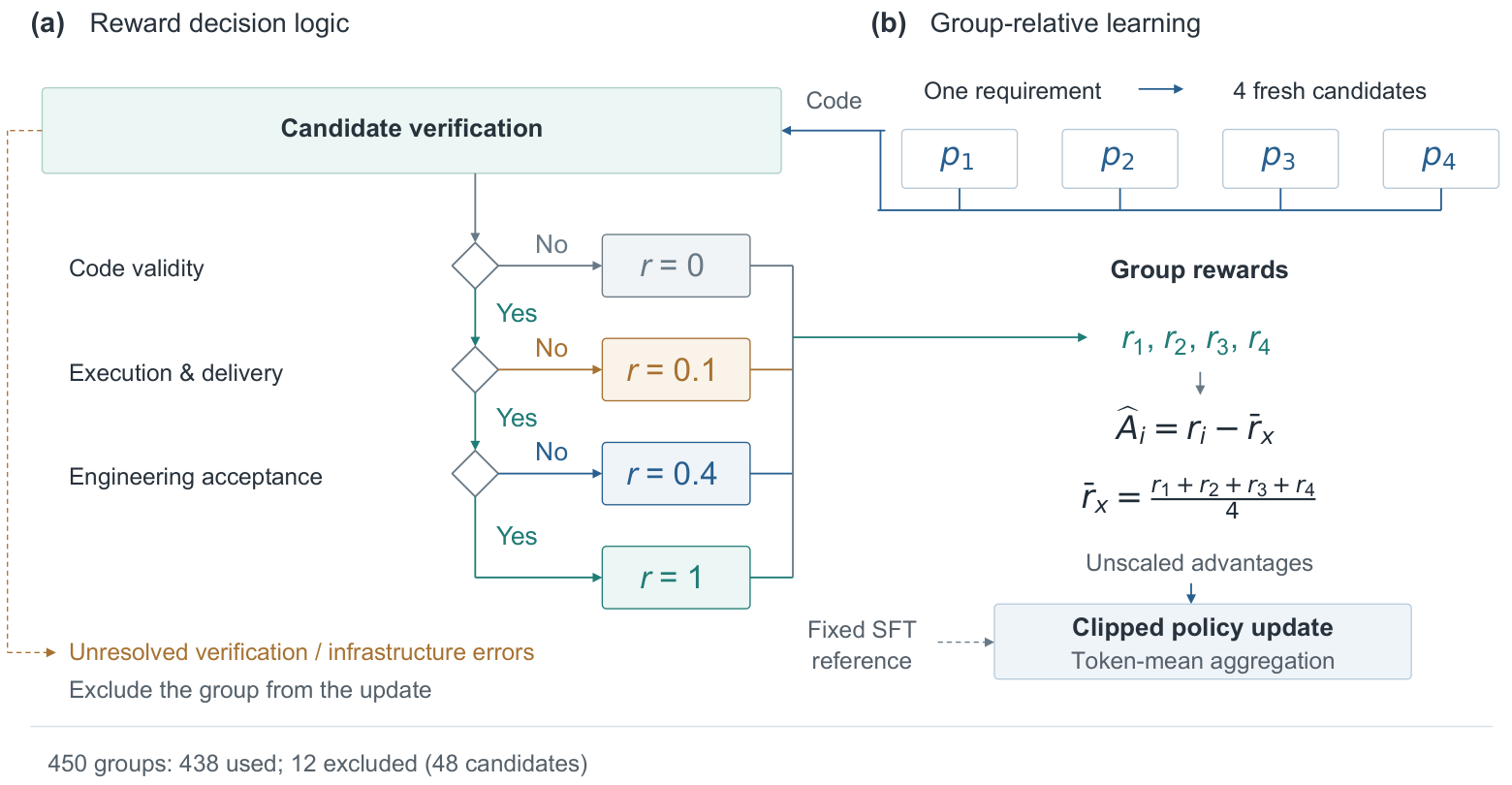}
\end{center}

\noindent\textbf{Figure~\thefigure:} Verification-guided GRPO training in OceanCoder. Four candidate programs are generated for each engineering requirement. Execution and engineering verification provide graded rewards for policy updates.\par

\label{fig:reward_grpo}
\par\endgroup

\subsection{Stage II: closed-loop reinforcement learning}
\label{subsec:grpo}

Online GRPO connects program generation to engineering verification through repeated rounds of sampling, simulation and model updating, as shown in Figure~\ref{fig:reward_grpo}. Each round begins with a batch of engineering requirements. The current model generates four candidate programs for each requirement and remains fixed while those candidates are evaluated. Candidate programs first undergo format, syntax and static safety checks. Those that pass proceed to OrcaFlex execution and engineering verification. The resulting verification states determine the rewards in Table~\ref{tab:reward_levels}. This feedback addresses errors produced by the current model on the engineering tasks used for training.

\begin{table}[htbp]
\caption{Four-level rewards based on program execution and engineering task acceptance.}

\label{tab:reward_levels}
 \centering
\begin{tabularx}{\linewidth}{@{}cX@{}}
\toprule
Reward & Established outcome \\
\midrule
0 & Format, syntax or static safety checks fail. \\
0.1 & Code checks pass, but execution, file delivery or model reloading fails. \\
0.4 & Execution, delivery and reloading pass, but engineering acceptance is incomplete. \\
1 & All applicable engineering acceptance checks pass. \\
\bottomrule
\end{tabularx}
\par
\end{table}

The reward hierarchy preserves progress towards complete task acceptance. Within each four-candidate group, programs may fail at different stages, so intermediate rewards can provide a learning signal even when no candidate completes the task. Following the group-relative comparison in GRPO \citep{Shao2024DeepSeekMath}, we compare each reward with the group mean to obtain the advantage in Eq.~\eqref{eq:group_advantage},
\begin{equation}
\widehat{A}_i
= r_i - \frac{1}{4}\sum_{j=1}^{4}r_j,
\label{eq:group_advantage}
\end{equation}
where $r_i$ is the reward of candidate $i$, and $\widehat{A}_i$ is its advantage relative to the group mean. The advantages are not divided by the group standard deviation, following the corresponding unscaled-advantage formulation \citep{Liu2025DrGRPO}. Groups with unresolved verification outcomes are excluded from optimisation.

The policy update combines a clipped probability-ratio objective \citep{Schulman2017PPO} with a KL penalty relative to the fixed SFT policy. Three policies have distinct roles in this calculation. The sampling policy supplies the candidate programs and their original token probabilities. The current policy is updated using those candidates. The reference policy retains the selected SFT checkpoint, including its trained adapters, and provides the regularisation target. Table~\ref{tab:policy_notation} summarises the notation used below.

For token $p_{i,t}$ conditioned on $q_{i,t}$, Eq.~\eqref{eq:policy_ratio} gives the ratio between the current and sampling policies,
\begin{equation}
\rho_{i,t}=\frac{\pi_\theta(p_{i,t}\mid q_{i,t})}
{\pi_{\mathrm{old}}(p_{i,t}\mid q_{i,t})}.
\label{eq:policy_ratio}
\end{equation}
The group advantage is shared across the response tokens of each candidate. The loss in Eq.~\eqref{eq:grpo_objective} is averaged over the usable response tokens, using token-level aggregation \citep{Yu2025DAPO},
\begin{equation}
\begin{aligned}
\mathcal L_{\mathrm{GRPO}}&=\frac{1}{M}\sum_{i,t}\ell_{i,t},\\
\ell_{i,t}&=-\min\!\left(\rho_{i,t}\widehat A_i,
\widetilde\rho_{i,t}\widehat A_i\right)+\beta D_{i,t}.
\end{aligned}
\label{eq:grpo_objective}
\end{equation}
Here, $\widetilde\rho_{i,t}=\operatorname{clip}(\rho_{i,t},1-\epsilon,1+\epsilon)$. The fixed reference defines $\delta_{i,t}=\log\pi_{\mathrm{ref}}(p_{i,t}\mid q_{i,t})-\log\pi_\theta(p_{i,t}\mid q_{i,t})$ and $D_{i,t}=\exp(\delta_{i,t})-\delta_{i,t}-1$. We use $\epsilon=0.2$ and $\beta=0.02$. Clipping limits the contribution of large probability-ratio changes, while the KL term penalises departures from the SFT reference.

\begin{table}[htbp]
\caption{Notation used in the group-relative policy update.}

\label{tab:policy_notation}
 \centering
\begin{tabularx}{\linewidth}{@{}lX@{}}
\toprule
Symbol & Meaning \\
\midrule
$p_{i,t}$ & Token $t$ of candidate program $i$. \\
$q_{i,t}$ & Prompt and preceding tokens of that candidate. \\
$\pi_\theta$, $\pi_{\mathrm{old}}$ & Current and sampling policies, respectively. \\
$\pi_{\mathrm{ref}}$ & Fixed SFT reference policy. \\
$r_i$, $\widehat A_i$ & Candidate reward and group-centred advantage. \\
$\rho_{i,t}$, $\widetilde\rho_{i,t}$ & Unclipped and clipped probability ratios. \\
$\delta_{i,t}$ & Reference minus current log probability. \\
$D_{i,t}$ & Sampled KL regularisation term. \\
$M$ & Total number of usable response tokens. \\
$\epsilon$, $\beta$ & Clipping parameter and KL coefficient. \\
\bottomrule
\end{tabularx}
\par
\end{table}

Only candidates with completed verification outcomes contribute to an update. If a solver run is interrupted by a licence or system timeout, the affected group is excluded so that an incomplete run is not treated as an engineering failure. Equal-reward groups have zero centred task advantage under Eq.~\eqref{eq:group_advantage}, while the reference-policy KL term remains defined. Once the usable groups have supplied an update, the current policy generates fresh programs for the next batch. Native execution and verification are repeated for these new candidates, maintaining feedback from the policy's current behaviour (Appendix~\ref{app:grpo_diagnostics} details the epoch-wise group reward diagnostics).

\section{Experimental setup}
\label{sec:experimental_setup}

Section 3 describes how the outcomes of OrcaFlex simulations guide the training of OceanCoder. We now assess the modelling capability acquired through training. We evaluate the models on a common set of held-out tasks that range from changes in individual parameters to complete engineering scenarios.

\subsection{Benchmark tasks and engineering scenarios}
\label{subsec:benchmark_tasks}

The benchmark comprises 665 tasks. Each task specifies the physical system to model, the analyses to perform and the engineering responses to report. All test instances are distinct from those used for supervised fine-tuning (SFT) or reinforcement learning. We organise the tasks into three strata to assess performance under parameter changes, structural modifications and engineering scenarios. Table~\ref{tab:test_tasks} summarises their composition.

\begin{table}[htbp]
\caption{Composition of the 665-task evaluation set.}

\label{tab:test_tasks}
\centering
\begin{tabularx}{\linewidth}{@{}>{\raggedright\arraybackslash}X r r@{}}
\toprule
Task stratum & Tasks & Share (\%) \\
\midrule
Core parameter variation & 352 & 52.93 \\
Structural variation & 237 & 35.64 \\
Engineering context & 76 & 11.43 \\
\midrule
Total & 665 & 100.00 \\
\bottomrule
\end{tabularx}
\par
\end{table}

The 352 core parameter variations retain a familiar modelling procedure while changing its physical or operating parameters. They test whether a generated program correctly applies the specified values within that procedure. The 237 structural variations require changes to the model configuration. A task may prescribe different properties for a local line segment, change the rotational restraint at a line end or alter the number of supports. The program must implement the change in the relevant part of the model and extract the requested response at the specified location.

The categories and strata describe different aspects of the benchmark. Categories identify the engineering analysis family, while strata identify the change that a generated implementation must accommodate. This permits comparisons within a common task family as well as across different kinds of requirement changes. The category-by-stratum counts in Appendix~\ref{app:category_results} document this coverage. Evaluation tasks retain their own specifications, reference evidence and output requirements, so acceptance is determined for each complete task.

The remaining 76 tasks are based on documented engineering scenarios. We adapted 18 tasks from published offshore project descriptions and 58 from industry verification examples and technical documentation. Each scenario is expressed as a self-contained specification of the configuration, operating conditions, analysis sequence and required outputs. These tasks assess whether a model can combine the specified requirements in a complete OrcaFlex program. Table~\ref{tab:engineering_sources} summarises the source groups; Appendix~\ref{app:case_sources} provides the source catalogue and task mappings.

Constructing these tasks requires converting source descriptions into explicit numerical specifications. Representative load cases, coordinate frames, component properties and installation geometries are defined so that each problem can be implemented and assessed independently. Reference programs are reviewed against the resulting specifications, including units, connections, boundary conditions, analysis stages and response definitions. Native execution, model reloading and output recomputation provide the corresponding reference evidence. The source catalogue and task mappings retain the connection between the public engineering material and the self-contained benchmark problem.

\begin{table}[htbp]
\caption{Provenance of the 76 engineering-context evaluation tasks. Public sources provide engineering scenario and modelling context for the self-contained benchmark specifications.}

\label{tab:engineering_sources}
\centering
\begin{tabularx}{\linewidth}{@{}>{\raggedright\arraybackslash}X r >{\raggedright\arraybackslash}X >{\raggedright\arraybackslash}X@{}}
\toprule
Source class & Tasks & Examples of source context & Relation to the evaluated task \\
\midrule
Official examples, documentation, validation or user material & 58 & Moorings, production risers, deployment, contact and line feeding. & Scenario and modelling-procedure basis, formulated with explicit task parameters and required outputs. \\
Published offshore engineering projects & 18 & Fensfjorden, Browse trunkline, Veja Mate, Baltic Sea inter-array cables and Project Natick. & Self-contained simulation specifications adapted from published offshore project descriptions and operating conditions. \\
\midrule
Total & 76 & Source-card and case mappings in Appendix~\ref{app:case_sources}. & Native simulation references and task-specific acceptance checks. \\
\bottomrule
\end{tabularx}
\par
\end{table}

\subsection{Compared models and training configurations}
\label{subsec:compared_models}

We compare OceanCoder with Qwen3-8B, Qwen3-235B-A22B-Instruct-2507 (hereafter Qwen3-235B) and ChatGPT-6 Sol. Qwen3-8B establishes performance at the starting point of OceanCoder training. Qwen3-235B provides a larger baseline from the same model family, and ChatGPT-6 Sol provides a closed-source comparison.

Three OceanCoder configurations allow us to examine how performance changes during training. OceanCoder-SFT is trained on the 12,000 verified task--program pairs described in Section~\ref{subsec:data_construction}. SFT runs for five epochs, comprising 3,750 updates. We select the checkpoint at update 2,250, corresponding to the end of the third epoch, for evaluation and subsequent reinforcement learning. Its comparison with the Qwen3-8B base model measures the change in task completion after domain supervision.

Both reinforcement learning configurations start from the selected SFT checkpoint. They use the same 150 training requirements, candidate sampling procedure and policy update settings. GRPO-Binary assigns a reward of one to a fully accepted program and zero to all other outcomes. Four-level GRPO uses the graded rewards defined in Section~\ref{subsec:grpo}. Comparing four-level GRPO with OceanCoder-SFT measures the additional change after reinforcement learning. Comparing it with GRPO-Binary tests the contribution of intermediate rewards. The training and evaluation settings are summarised in Table~\ref{tab:training_settings}.

\begin{table}[htbp]
\caption{OceanCoder training and evaluation settings.}

\label{tab:training_settings}
 \centering
\begin{tabularx}{\linewidth}{@{}l >{\raggedright\arraybackslash}X >{\raggedright\arraybackslash}X@{}}
\toprule
Setting & Stage I: SFT & Stage II: GRPO \\
\midrule
Initial model & Qwen3-8B & SFT (update 2,250) \\
Training data & 12,000 pairs (12 cats.) & 150 requirements \\
Schedule & 5 epochs (3,750 steps) & 3 epochs (57 steps) \\
Learning rate & $2\times 10^{-4}$ & $5\times 10^{-6}$ \\
Batching & 1/GPU, grad.\ accum.\ 16 & 8 prompts/batch \\
Sampling & -- & $G=4$, $T=1.0$, $p=0.95$ \\
Rewards & -- & $\{0,0.1,0.4,1\}$ \\
Checkpoint & Epoch 3 (step 2,250) & Epoch 3 (step 57) \\
Evaluation & 3 rounds $\times$ 665 tasks & Same as SFT \\
\bottomrule
\end{tabularx}
\par
\par
 SFT uses 4-bit NF4 QLoRA ($r=16$, $\alpha=32$, dropout 0.05). GRPO-Binary uses identical settings with binary rewards $\{0,1\}$.
\end{table}

Each evaluated model is assessed across three evaluation rounds on the 665 benchmark tasks, with one generation attempt per task in each round. We report the mean and sample standard deviation of the evaluation metrics across the three rounds. Task-level records from the first evaluation round are used for stage-wise failure analysis and representative engineering case studies.

\subsection{Single-attempt generation and engineering assessment}
\label{subsec:evaluation_protocol}

Each model receives the task description and a common system instruction specifying the program and output requirements. Reference programs and expected numerical results are withheld during generation. Each model makes one attempt to generate a complete OrcaFlex program for each task. The evaluator assesses the resulting submission without simulator-feedback repair. The model receives no simulator feedback during generation and has no opportunity to revise its submission after execution.

Generation and assessment use fixed software and output requirements. The simulations use OrcaFlex and OrcFxAPI 11.6. Evaluation uses a decoding temperature of 0 and allows up to 16,384 completion tokens within the model's context window. Responses that reach the generation limit remain in the assigned-task denominator. Appendix~\ref{app:generation_review} records the decoding and termination rules used in the common protocol.

Assessment begins with checks on response format, Python syntax and static safety requirements. Programs that pass these checks are executed in OrcaFlex. After execution, the evaluator checks that the required outputs have been delivered and that the saved native files can be reloaded. It then applies the engineering criteria from Section~\ref{subsec:verification} to the saved model, analysis records and exported responses. A task is accepted only when all applicable checks pass, including the numerical tolerances specified for that task.

The assessment retains the stage at which a submission first fails. A format or code failure prevents a valid simulation from being established. A runtime failure interrupts the prescribed calculation. Delivery and reloading checks establish whether the required model and output files have been saved and can be inspected. Subsequent model, analysis and response checks determine complete engineering acceptance. These recorded outcomes support the cumulative counts in Appendix~\ref{app:acceptance_records} and the failure analysis in Section~\ref{sec:results_failures}. The same sequence is applied to every model and evaluation round.

\subsection{Evaluation metrics}
\label{subsec:evaluation_metrics}

The primary metric is Pass@1, the percentage of assigned tasks successfully completed in a single generation attempt under the engineering acceptance criteria in Section~\ref{subsec:evaluation_protocol}. The execution success rate (ESR) is the percentage for which the program runs to completion in OrcaFlex without a runtime error. Both metrics use all assigned tasks as the denominator, including submissions that reach the generation limit or fail a code check. The denominator is 665 for the overall comparison and the corresponding number of assigned tasks for each stratum.

Each metric is calculated separately for each evaluation round. The reported mean and sample standard deviation are then computed across the three rounds, with 665 tasks in the denominator of each overall rate.

Every program counted in Pass@1 also contributes to ESR. The difference between ESR and Pass@1, expressed in percentage points, gives the proportion of assigned tasks that execute successfully but fail a subsequent delivery, reload or engineering check. Reporting the two rates together distinguishes changes in the ability to generate runnable programs from changes in the ability to complete the specified engineering task.

\begin{table}[htbp]
\caption{Overall and stratified performance on the 665-task benchmark. ESR denotes execution success rate, and Pass@1 measures complete task success. Values with standard deviations are reported as mean $\pm$ sample standard deviation across three evaluation rounds. Stratum sizes are given in parentheses.}

\label{tab:main_results}
\begingroup
 \centering
\begin{tabularx}{\linewidth}{@{}>{\raggedright\arraybackslash}X r r@{}}
\toprule
Model & ESR (\%) & Pass@1 (\%) \\
\midrule
Qwen3-8B
& 0.00 & 0.00 \\
Qwen3-235B
& 0.00 & 0.00 \\
ChatGPT-6 Sol
& $27.52 \pm 1.52$ & $7.82 \pm 1.27$ \\
OceanCoder-SFT
& $69.17 \pm 1.40$ & $46.77 \pm 1.08$ \\
OceanCoder-GRPO-Binary
& $67.37 \pm 1.63$ & $46.47 \pm 1.43$ \\
OceanCoder-GRPO
& $\mathbf{70.53 \pm 1.08}$ & $\mathbf{51.43 \pm 0.94}$ \\
\midrule
Task stratum
& SFT Pass@1 (\%) & GRPO Pass@1 (\%) \\
\midrule
Core parameter variation (352)
& $80.78 \pm 1.34$ & $\mathbf{88.83 \pm 1.18}$ \\
Structural variation (237)
& $10.69 \pm 0.88$ & $\mathbf{11.25 \pm 0.64}$ \\
Engineering context (76)
& $1.75 \pm 0.76$ & $\mathbf{3.51 \pm 0.76}$ \\
\bottomrule
\end{tabularx}
\par
\endgroup
\end{table}

\section{Results and analysis}
\label{sec:results}

The evaluation in Section~\ref{sec:experimental_setup} measures both successful execution and completion of the specified engineering task. We first compare overall and stratified model performance, then assess the reliability of engineering verification and the contribution of graded training feedback. Finally, failure patterns and representative engineering cases are examined to explain the observed capabilities and remaining limitations.

\subsection{Overall and stratified task performance}
\label{sec:results_overall}

OceanCoder achieves higher complete-task success rates than the evaluated base and closed-source models under the common single-attempt protocol. As shown in Table~\ref{tab:main_results}, neither Qwen base model completes any benchmark task, while ChatGPT-6 Sol achieves a Pass@1 of 7.82\%. Supervised fine-tuning on verified engineering examples increases Pass@1 to 46.77\%, and subsequent GRPO training raises it to 51.43\%. The improvement is concentrated in core parameter variations, where Pass@1 increases from 80.78\% to 88.83\%, a gain of 8.05 percentage points. By comparison, structural variations and engineering-context tasks reach only 11.25\% and 3.51\%, respectively, indicating that changes in structural configuration and broader engineering requirements remain substantially more difficult than parameter variations within familiar task families.

The distribution of gains helps identify the capability improved by engineering feedback. Figure~\ref{fig:stage_performance}(c) shows that 28 of the 31 additional accepted tasks in the first evaluation round occur in the core stratum, with category-level counts detailed in Appendix~\ref{app:category_results}. These tasks retain a familiar modelling procedure and require the program to apply different physical or operating values correctly. The concentration of gains in this stratum is consistent with refinement of previously learned modelling routines. Structural variations additionally require assigning changed properties, connections or restraints to the appropriate parts of the system. Their smaller improvement shows that applying a known procedure to new values and adapting that procedure to a changed configuration remain different generation requirements.

\par\begingroup
\refstepcounter{figure}
\begin{center}
\includegraphics[width=\linewidth,alt={Overall and stratified task success and category-level gains from SFT to GRPO.}]{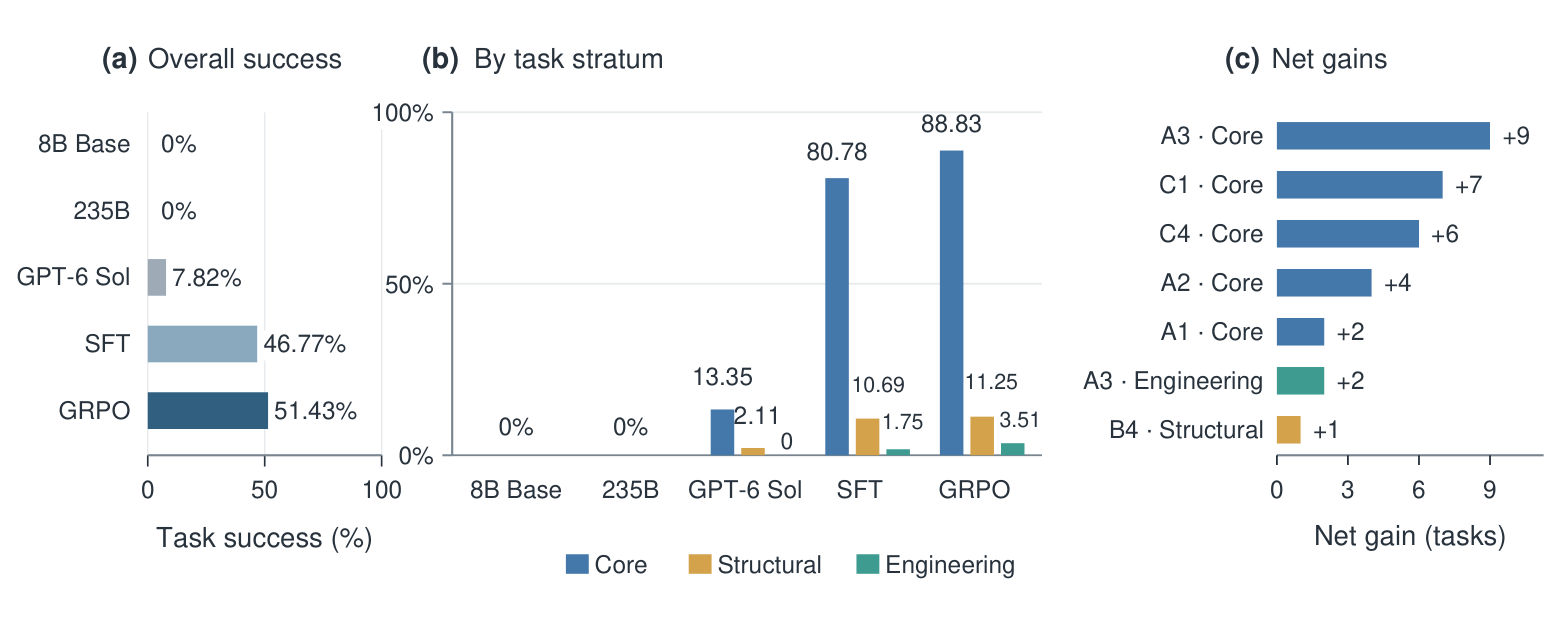}
\end{center}

\noindent\textbf{Figure~\thefigure:} Complete task success on the 665-task benchmark. (a) Overall Pass@1 for the compared models. (b) Pass@1 by task stratum. Panels (a) and (b) report means across three evaluation rounds. GPT-6 Sol denotes ChatGPT-6 Sol. (c) Category-level net increases in accepted tasks from SFT to GRPO in the first evaluation round. The gains sum to 31 tasks.\par

\label{fig:stage_performance}
\par\endgroup

Execution success does not necessarily translate into complete engineering acceptance. ChatGPT-6 Sol successfully executes 27.52\% of its programs, but only 7.82\% satisfy all task requirements. A similar distinction remains after domain adaptation. From SFT to GRPO, Pass@1 increases by 4.66 percentage points, whereas ESR increases by only 1.36 percentage points. Consequently, the gap between ESR and Pass@1 narrows from 22.40 to 19.10 percentage points. This indicates a reduction in the proportion of assigned tasks for which programs execute successfully but fail subsequent delivery, model reloading or engineering acceptance. The difference between these metrics motivates a closer examination of the verification criteria used to determine task completion.

\subsection{Reliability of engineering verification}
\label{sec:results_verifier_audit}

To examine whether engineering verification can distinguish executable programs from correct implementations, we manually audit 100 SFT-generated programs covering all 12 task categories. Independent inspection of task requirements, programs, execution records and numerical outputs identifies 48 valid and 52 invalid implementations. Table~\ref{tab:verification_audit} compares the automated decisions with these manual judgements. Execution checks alone achieve 78.00\% agreement, incorrectly accepting 22 programs that complete execution but fail the specified engineering task. Among the 70 executable programs, these failures account for 31.43\%. Sixteen involve incorrect model configurations, while six involve errors in response extraction. Adding native-model input checks identifies the configuration errors and increases agreement to 94.00\%; output verification detects the remaining six errors.

\begin{table}[htbp]
\caption{Verification decisions compared with manual inspection of 100 programs (48 valid, 52 invalid). False acceptances (FA) and false rejections (FR) are reported as counts.}

\label{tab:verification_audit}
 \centering
\begin{tabularx}{\linewidth}{@{}>{\raggedright\arraybackslash}X r r r@{}}
\toprule
Verification criterion & FA & FR & Agree. (\%) \\
\midrule
Execution checks only & 22 & 0 & 78.00 \\
Execution and model-input checks & 6 & 0 & 94.00 \\
Unnormalised raw output comparison & 0 & 13 & 87.00 \\
\textbf{Full three-layer verifier} & \textbf{0} & \textbf{0} & \textbf{100.00} \\
\bottomrule
\end{tabularx}
\par
\end{table}

The identified errors illustrate why model-input and output checks are both necessary. In some lowering tasks, generated programs assign a line diameter of 55~mm and an axial stiffness of 120~MN instead of the prescribed 70~mm and 210~MN. Although the solver completes the analysis, the constructed physical model differs from the specification. Conversely, a restoring-force program passes the model-input checks but reports a directional force of $+607.23$~kN when the required projection is $-607.23$~kN. This error arises during response extraction rather than model construction. These cases demonstrate that successful execution and correct physical configuration do not individually guarantee complete task acceptance.

Verification must also avoid rejecting physically equivalent results. Unnormalised output comparison falsely rejects 13 valid implementations: ten differ only in non-physical metadata labels, two represent equivalent directions across the $0^\circ/360^\circ$ boundary, and one contains a peak timestamp affected by numerical noise in a nearly zero yaw response. Filtering non-physical metadata, normalising periodic angles and applying the prescribed numerical tolerances eliminate these false rejections. The full three-layer verifier consequently accepts all 48 valid implementations and rejects all 52 invalid ones, achieving 100.00\% agreement with manual inspection on the audited sample.

False acceptances would reward programs that fail the engineering specification, while false rejections would penalise valid implementations. The audit therefore supports using both model-input and output verification to construct task-based training feedback.

\subsection{Effect of graded verification feedback}
\label{sec:results_ablation}
\label{sec:results_training}

\par\begingroup
\refstepcounter{figure}
\begin{center}
\includegraphics[width=\linewidth,alt={Engineering verification rewards and reward composition during GRPO training.}]{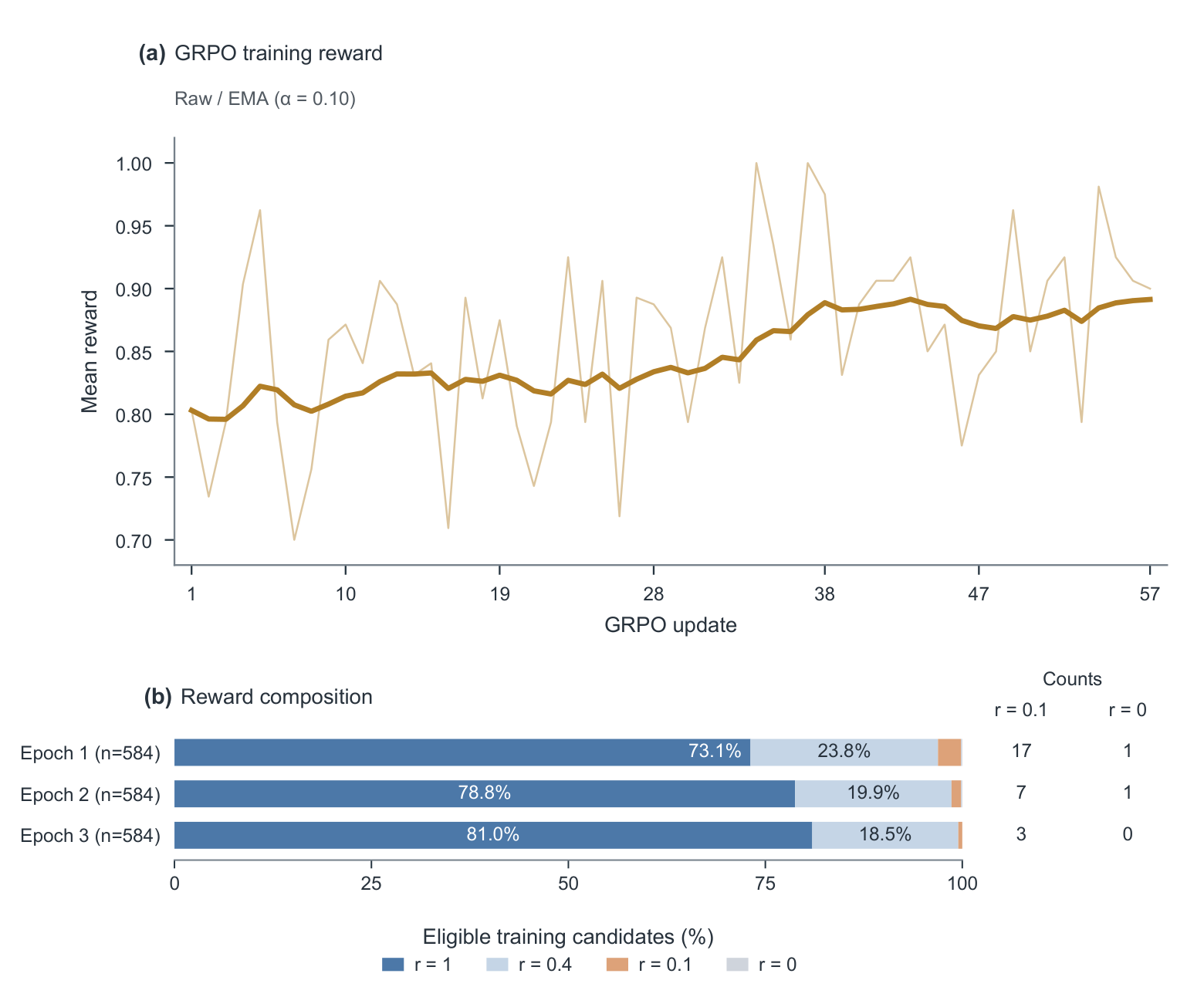}
\end{center}

\noindent\textbf{Figure~\thefigure:} Engineering verification outcomes during GRPO training. (a) Mean reward at each update and its exponential moving average with $\alpha=0.10$. (b) Reward composition across the three training epochs, with 584 eligible candidates per epoch. Full acceptance receives a reward of 1. Intermediate rewards distinguish execution or delivery failures from executable programs that fail engineering acceptance.\par

\label{fig:training_dynamics}
\par\endgroup

The reward ablation and training trajectory (Figure~\ref{fig:training_dynamics}) evaluate whether intermediate engineering verification outcomes improve policy learning beyond binary task acceptance. Under matched training settings, four-level GRPO achieves a mean Pass@1 of 51.43\%, compared with 46.47\% for binary rewards, an improvement of 4.96 percentage points. Binary-reward GRPO remains close to the SFT baseline of 46.77\%. The difference can be understood through group-relative advantage estimation. When all four candidates fail complete acceptance, binary rewards assign zero to every candidate, producing zero relative advantages regardless of their execution progress. Four-level rewards instead distinguish failures at different stages. In the training records, 142 of 438 usable groups (32.42\%) produce non-zero relative advantages under four-level rewards, compared with 118 groups (26.94\%) under binary rewards.

The frequency of intermediate outcomes further explains their contribution. Among 1,752 eligible training candidates, 363 pass execution, delivery and model reloading but fail engineering acceptance, receiving a reward of 0.4. Another 27 pass the code checks but fail execution, delivery or model reloading, receiving 0.1. Together, these 390 candidates account for 22.26\% of the sampled programs. Under a binary scheme, their progress would not be distinguished from earlier failures. The four-level hierarchy retains this information while reserving the maximum reward for complete engineering acceptance. The ablation results therefore support using verification states that reflect different stages of task completion.

The reward comparison also shows a difference in retained execution capability. Binary-reward GRPO records an ESR of 67.37\%, compared with 69.17\% for SFT and 70.53\% for four-level GRPO. Thus, the configuration that removes intermediate distinctions has lower execution success as well as lower complete-task success. The intermediate states describe progress that is observable before final acceptance, including delivery of reloadable models. Their ordering supplies relative preferences among incomplete candidates generated for the same requirement. Complete acceptance remains governed by the same engineering checks for every evaluated checkpoint. The improvement under graded rewards therefore accompanies more successful execution and a higher rate of satisfying the full specification.

The training dynamics are consistent with this interpretation. As shown in Figure~\ref{fig:training_dynamics}, mean reward increases from 0.8293 in the first epoch to 0.8844 in the third epoch. Over the same period, the full-acceptance rate rises from 73.12\% to 80.99\%, while intermediate outcomes decrease from 26.71\% to 19.01\%. These changes indicate that a larger proportion of programs sampled during training satisfy the complete engineering requirements. They describe training-set behaviour rather than independent generalisation; the separate benchmark results in Table~\ref{tab:main_results} establish the corresponding improvement on evaluation tasks.

\subsection{Failure modes and representative engineering cases}
\label{sec:results_failures}
\label{sec:results_cases}

\par\begingroup
\refstepcounter{figure}
\begin{center}
\includegraphics[width=\linewidth,alt={Execution success, engineering acceptance, and failure stages in benchmark evaluation.}]{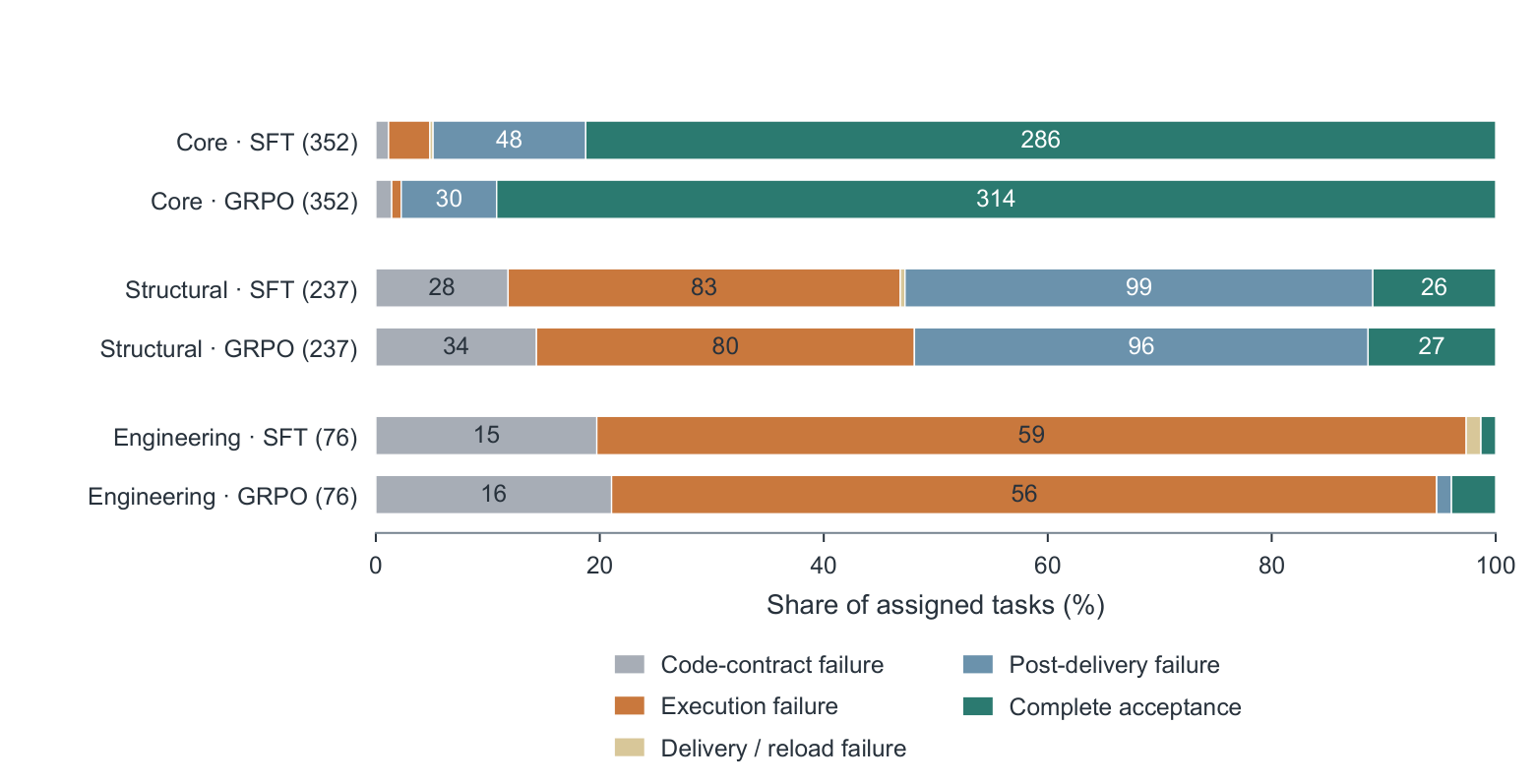}
\end{center}

\noindent\textbf{Figure~\thefigure:} Failure and acceptance outcomes for SFT and four-level GRPO across task strata in the first evaluation round. Each bar partitions the assigned tasks by the earliest failed stage or complete acceptance. The stages cover code-contract compliance, execution, delivery and model reloading, and subsequent engineering acceptance. Parentheses indicate stratum sizes.\par

\label{fig:failure_reward_analysis}
\par\endgroup

Despite the improvements in complete-task success, performance remains strongly dependent on task complexity. Figure~\ref{fig:failure_reward_analysis} shows the stage-wise outcomes from the first evaluation round. On structural variations, 123 GRPO programs execute successfully, but only 27 pass complete engineering acceptance. The remaining 96 executable programs fail subsequent acceptance checks, indicating persistent difficulties in preserving modified physical configurations or extracting the required responses. Engineering-context tasks exhibit a different failure pattern: only four programs execute successfully, three of which pass complete acceptance. Their principal difficulties therefore arise earlier, during program construction and execution. These patterns indicate that structural modifications and broader engineering scenarios challenge different stages of the modelling process.

These patterns give different meanings to failure in the two strata. A structural variant can retain most of a familiar program while changing a local segment, connection or restraint. The implementation must associate the modification with the correct object and preserve the requested response location. A solvable model can still contain an incorrect assignment. For engineering-context tasks, the low execution rate places the main observed difficulty earlier in the workflow. The model must assemble the required objects, API operations and analysis sequence into a complete executable program. The stage records consequently distinguish a failure to construct that workflow from a failure to implement its engineering details.

\par\begingroup
\refstepcounter{figure}
\begin{center}
\includegraphics[width=\linewidth,alt={Representative ocean engineering cases and comparison of generated simulation responses.}]{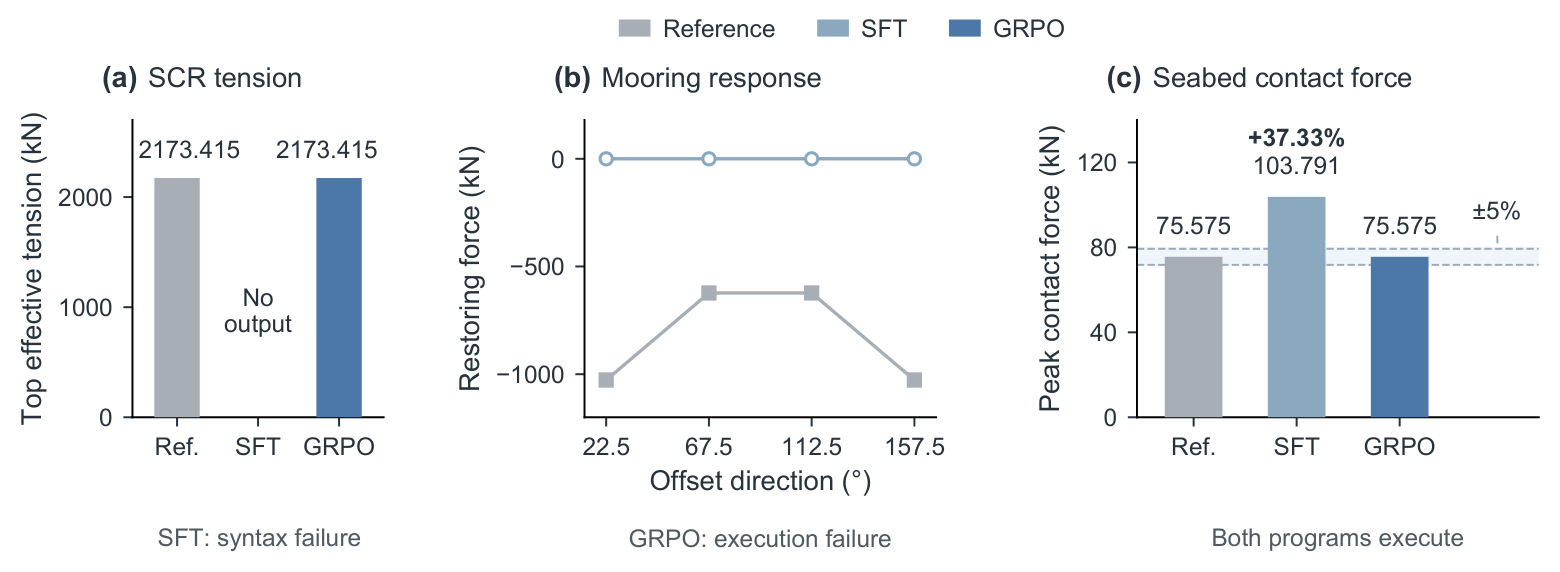}
\end{center}

\noindent\textbf{Figure~\thefigure:} Reference solutions and OceanCoder outputs for three representative engineering tasks. (a) Top effective tension in the steel catenary riser task. (b) Horizontal restoring forces at four vessel offset directions in the mooring task. (c) Peak upward seabed contact force in the two-point lowering task introduced in Figure~\ref{fig:motivation}. Syntax and execution failures are marked explicitly. The shaded band in panel (c) indicates the $\pm5\%$ task tolerance around the reference force.\par

\label{fig:engineering_cases}
\par\endgroup

Three representative cases from the first evaluation round illustrate these differences in greater detail (Appendix~\ref{app:case_evidence} provides the detailed case specifications and numerical comparisons). The static steel catenary riser task requires modelling a riser in 1,200~m water depth with a length of 2,220~m. The SFT-generated program contains a malformed statement and fails syntax validation. In contrast, GRPO constructs the riser--seabed system, completes the static equilibrium analysis and reports a top effective tension of 2,173.415~kN, matching the reference value to three decimal places (Figure~\ref{fig:engineering_cases}(a)). This case demonstrates an improvement in program construction and execution, where the SFT model fails before a valid analysis can begin.

Structural modifications present a different challenge. In the four-point mooring task, a short terminal line segment with reduced axial stiffness is added, and horizontal restoring forces are required at four vessel offsets of 12~m. The SFT program executes but incorrectly assigns the multi-segment line collection, producing restoring forces close to zero rather than the reference range of $-1{,}026.96$ to $-622.85$~kN. The GRPO program attempts to configure the required segment arrays but encounters a Python tuple-unpacking error during execution. As shown in Figure~\ref{fig:engineering_cases}(b), neither program completes the task, although their failures occur at different stages. This case illustrates the remaining difficulty of generating programs for modified structural configurations, consistent with the low acceptance rate observed in the structural-variation stratum.

The two-point lowering case further demonstrates why complete engineering verification is necessary even when programs execute successfully. A 65~t subsea payload is lowered onto an elastic seabed using two winch-controlled lines. The prescribed line diameter, unit mass and axial stiffness are 70~mm, 18~kg/m and 210~MN, respectively. Both SFT and GRPO programs complete execution and export the required models and results. However, the SFT program assigns 55~mm, 11~kg/m and 120~MN, altering the physical system and consequently its contact response. It predicts first seabed contact at 35.0825~s rather than the reference 36.0126~s, an error of 0.9301~s that exceeds the 0.010~s tolerance. Its peak upward contact force is 103.791~kN, compared with the reference 75.575~kN, corresponding to a 37.33\% overprediction against a 5\% tolerance, and it overpredicts maximum foot penetration (0.09041~m compared with the reference 0.05918~m). The GRPO program preserves the prescribed line properties and lowering schedule, reproducing the reference contact time, peak force and penetration (Figure~\ref{fig:engineering_cases}(c)). This example illustrates how engineering verification distinguishes executable but physically incorrect programs from those that satisfy the complete task specification.

\section{Conclusions}
\label{sec:conclusions}

We have developed OceanCoder to generate complete OrcaFlex programs from natural-language ocean engineering requirements. Engineering-constrained data construction and supervised fine-tuning establish program generation for model construction, analysis execution and result extraction. A three-layer verifier then assesses the constructed model, completed analyses and reported responses. Its outcomes provide graded feedback for online GRPO and common acceptance criteria for the 665-task benchmark.

Across the benchmark, Pass@1 increases from 46.77\% after SFT to 51.43\% after four-level GRPO, compared with 46.47\% under binary rewards. The full verifier agrees with manual inspection on all 100 audited programs, including cases in which execution succeeds despite engineering errors. These results support the complementary roles of reliable domain supervision and engineering verification feedback. Most of the performance gain occurs in core parameter variations, while structural modifications and broader engineering scenarios remain difficult. Complete task acceptance therefore requires more than successful program execution.

The findings are limited to the task categories, parameter ranges and modelling assumptions represented in the corpus and benchmark. Engineering acceptance establishes consistency with the specified numerical problem and its required outputs; it does not by itself validate predictions against unobserved field conditions. Future work should address structural and engineering-scenario failures by expanding the configurations and analysis procedures represented in training, together with their verification criteria. Such extensions would broaden the modelling scope while retaining explicit links between requirements, generated programs and accepted results.

\section*{Data availability}

Representative seed specifications, source mappings, summaries of the 100-task verification audit and evaluation statistics are provided in the manuscript and appendices. The trained OceanCoder weights and the specifications of the 665-task benchmark will be released publicly upon publication.

\par
\appendix

\numberwithin{equation}{section}
\numberwithin{figure}{section}
\numberwithin{table}{section}

\renewcommand{\theHequation}{appendix.\arabic{section}.\arabic{equation}}
\renewcommand{\theHfigure}{appendix.\arabic{section}.\arabic{figure}}
\renewcommand{\theHtable}{appendix.\arabic{section}.\arabic{table}}

\section{Data construction and training records}
\label{app:training_records}

\par\begingroup
\refstepcounter{figure}
\begin{center}
\includegraphics[width=\linewidth,alt={Distributions of task-description and reference-program lengths in the supervised training corpus.}]{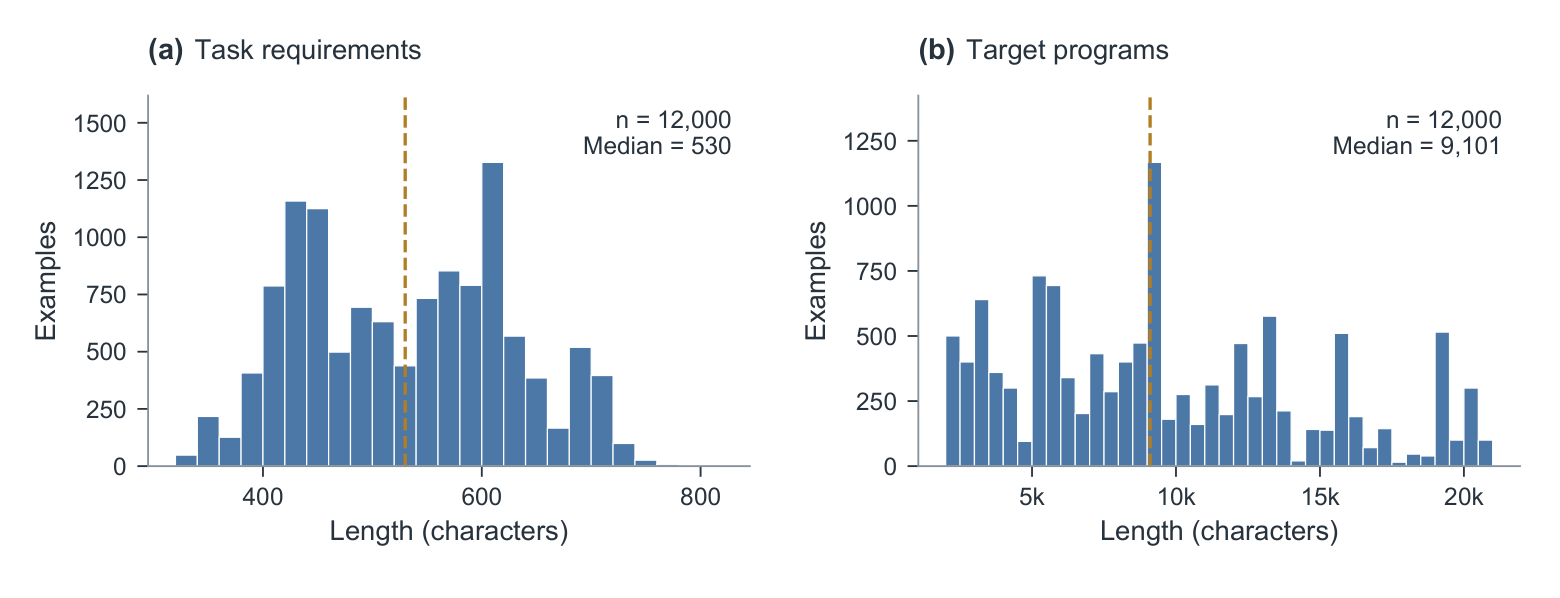}
\end{center}

\noindent\textbf{Figure~\thefigure:} Character-length distributions of the 12,000 SFT task requirements and target Python programs.\par

\label{fig:data_lengths}
\par\endgroup

\subsection{Corpus composition and example lengths}
\label{app:data_lengths}

The supervised corpus contains 12,000 instruction--program pairs spanning 12 task categories, with ten seed families per category and 100 parameter-expanded instances per seed. Each pair is generated from a distinct parameter set within its seed family.

Table~\ref{tab:app_data_lengths} and Figure~\ref{fig:data_lengths} report the character-length distributions of the task descriptions and target programs across the corpus.

\begin{table}[htbp]
\caption{Corpus records and character lengths (mean; median; range). GRPO requirements contain no supervised response.}

\label{tab:app_data_lengths}
 \centering
\begin{tabularx}{\linewidth}{@{}l r >{\raggedright\arraybackslash}X@{}}
\toprule
Resource & Records & Character lengths (mean; median; range) \\
\midrule
SFT requirements & 12,000 & 530.95; 530; 321--804 \\
SFT programs & 12,000 & 9,608.77; 9,101; 2,420--20,976 \\
GRPO requirements & 150 & 512.90; 502; 395--717 \\
\bottomrule
\end{tabularx}
\par
\end{table}

The SFT responses contain 67--462 lines, with a mean of 206.57 and a median of 221. The 150 GRPO requirements are sampled from the supervised training sources for online policy exploration, focusing on multi-component mooring and staged offshore-operation tasks (A2: 35; A3: 2; A4: 2; B1: 6; B2: 4; B4: 1; C1: 32; C2: 32; and C4: 36) where multi-step configuration and response-extraction errors are most frequent after SFT.

\subsection{Category-specific admission verification}
\label{app:admission_verification}

Each of the 12 categories in the supervised corpus contributes 1,000 verified task--program pairs. Candidate pairs undergo category-specific verification tailored to their physical modelling and numerical analysis requirements, combining static AST checks, native OrcaFlex solver execution, saved-model reload and recomputation, and output verification. At the seed-family level, manual review of model geometry and response curves is further combined with mesh and time-step sensitivity checks to establish baseline discretisation settings. Table~\ref{tab:app_admission} summarises the primary verification checks across the 12 categories.

\begin{table}[htbp]
\caption{Admission verification checks across the 12 SFT categories (1,000 pairs per category).}

\label{tab:app_admission}
 \centering
\begin{tabularx}{\linewidth}{@{}l >{\raggedright\arraybackslash}X@{}}
\toprule
Cat. & Primary physical and numerical verification checks \\
\midrule
A1 & Single-catenary static equilibrium, line-property and boundary coordinates, end tensions, minimum elevation and maximum curvature. \\
A2 & Multi-line mooring equilibrium, fairlead/anchor layout, segment properties and directional offset restoring forces. \\
A3 & Deepwater SCR static profile, hang-off and touchdown zone (TDZ) arc lengths, bending moments and effective tensions. \\
A4 & Pipeline lateral stability across stable and breakout regimes, hydrodynamic/coating parameters and lateral displacement. \\
B1 & Regular-wave catenary dynamics, build-up/steady-cycle window alignment, peak tension and dynamic excursion. \\
B2 & Coupled four-point mooring dynamics under waves and current, floater trajectory and fairlead tension envelopes. \\
B3 & Dynamic SCR response under top motion, waves and current, TDZ curvature, minimum tension and stress ranges. \\
B4 & Free-span pipeline wave--current response, span geometry, end restraints, mid-span oscillation and bending moments. \\
C1 & Static S-lay installation across stinger supports, overbend/sagbend curvature, tensioner load and touchdown reaction. \\
C2 & Dynamic cable-laying under vessel transit and payout, catenary evolution, seabed contact and bottom tension bounds. \\
C3 & Single-point lifting and splash-zone lowering, winch payout kinematics, hydrodynamic loading and hoist-line tensions. \\
C4 & Two-point subsea lowering under implicit dynamics, synchronous payout, first-contact timing, peak force and penetration. \\
\bottomrule
\end{tabularx}
\par
\end{table}

\subsection{Executed training settings and resources}
\label{app:executed_hyperparameters}

The principal hyperparameter settings for both stages are listed in Table~\ref{tab:training_settings}. During Stage I, the five-epoch SFT run (3,750 updates) required 44.92 h (161,724.67 s) in total, with a cumulative run-average training loss of 0.00807 across all updates. At the selected epoch-3 checkpoint (update 2,250, reached at 26.96 h; Figure~\ref{fig:sft_loss}), the logged step-level response-token cross-entropy loss had decreased from 0.8971 (at update 10) to 0.0022. Stage II (online GRPO) was executed for three epochs (57 updates; 19 updates per epoch) on two NVIDIA A800 (80 GB PCIe) GPUs running CUDA 12.4 and PyTorch 2.6.0.

\par\begingroup
\refstepcounter{figure}
\begin{center}
\includegraphics[width=\linewidth,alt={Supervised fine-tuning loss over training updates.}]{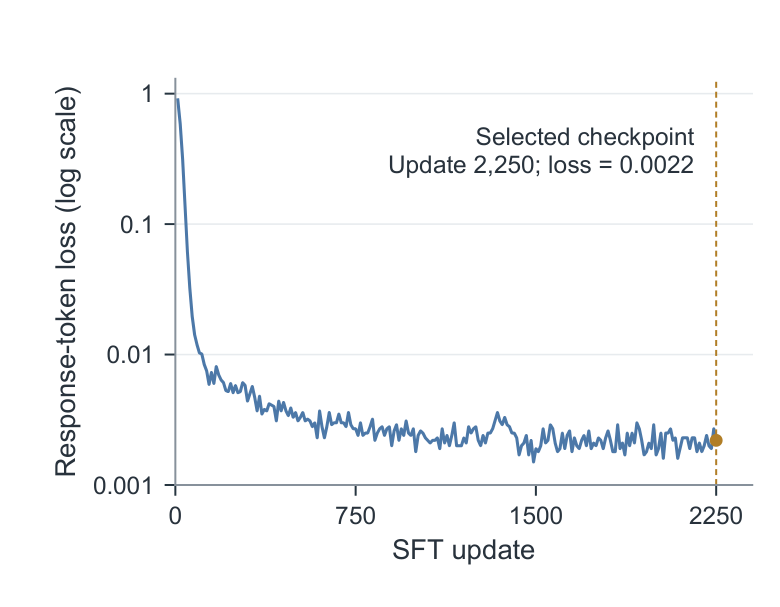}
\end{center}

\noindent\textbf{Figure~\thefigure:} SFT response-token loss trajectory through the selected epoch-3 checkpoint (update 2,250). All 225 logged loss values from updates 10 to 2,250 are shown on a logarithmic scale; the dashed line marks update 2,250, where the loss stabilises at 0.0022.\par

\label{fig:sft_loss}
\par\endgroup

\subsection{GRPO rewards, usable groups and partial progress}
\label{app:grpo_diagnostics}

Across three epochs, the training loop consumes 450 four-candidate groups and samples 1,800 nominal candidates. After the recorded verification exclusions, 438 groups and 1,752 candidates ($n=584$ per epoch) remain eligible for reward statistics and updates. Table~\ref{tab:app_epoch_rewards} retains the exact four-level counts. The mean training reward rises from 0.8293 to 0.8844, and the full-reward proportion rises from 73.12\% to 80.99\%. These quantities describe the reused training requirements and are distinct from success rates on evaluation tasks.

\par\begingroup
\refstepcounter{figure}
\begin{center}
\includegraphics[width=\linewidth,alt={GRPO reward groups, usable groups, and intermediate verification progress.}]{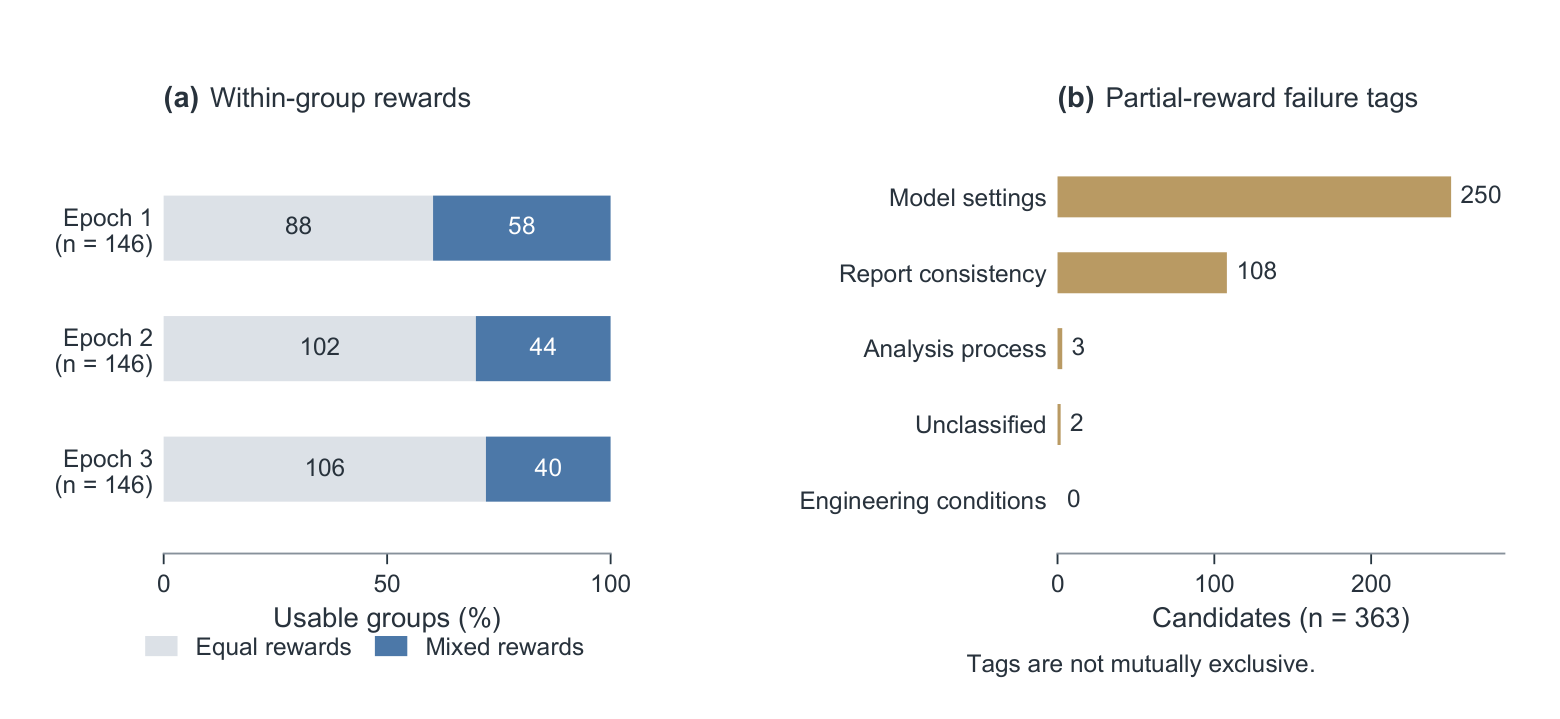}
\end{center}

\noindent\textbf{Figure~\thefigure:} Additional GRPO training diagnostics. The left panel separates equal-reward and mixed-reward usable groups by epoch; equal-reward groups have zero centred task advantage. The right panel shows recorded failure labels for the 363 reward-0.4 candidates.\par

\label{fig:reward_groups}
\par\endgroup

\begin{table}[htbp]
\caption{GRPO candidate reward counts by training epoch ($n=584$ eligible per epoch; 1,752 total). Partial denotes $r\in\{0.1,0.4\}$.}

\label{tab:app_epoch_rewards}
 \centering
\begin{tabularx}{\linewidth}{@{}l r r r r >{\raggedleft\arraybackslash}X r r@{}}
\toprule
Epoch & $r=0$ & $0.1$ & $0.4$ & $1$ & Mean & Full (\%) & Part. (\%) \\
\midrule
1 & 1 & 17 & 139 & 427 & 0.8293 & 73.12 & 26.71 \\
2 & 1 & 7 & 116 & 460 & 0.8683 & 78.77 & 21.06 \\
3 & 0 & 3 & 108 & 473 & 0.8844 & 80.99 & 19.01 \\
\midrule
Total & 2 & 27 & 363 & 1360 & 0.8607 & 77.63 & 22.26 \\
\bottomrule
\end{tabularx}
\par
\end{table}

Four groups are excluded in each of the three epochs due to solver-licence or timeout interruptions. The 48 affected candidates are omitted from gradient updates rather than assigned zero rewards.

Among usable groups, 88 of 146 have identical candidate rewards in epoch 1, 102 of 146 in epoch 2, and 106 of 146 in epoch 3, corresponding to 58, 44 and 40 mixed-reward groups, respectively (Figure~\ref{fig:reward_groups}a). Under the group-centred advantage formulation in Eq.~\eqref{eq:group_advantage}, equal-reward groups yield zero relative task advantage while still contributing through reference-policy KL regularisation.

Table~\ref{tab:app_partial_failures} and Figure~\ref{fig:reward_groups}(b) summarise the verification gates failed by the 363 candidates receiving $r=0.4$. Because a single candidate can trigger multiple verification flags, these categories represent diagnostic failure tags rather than mutually exclusive root causes.

\begin{table}[htbp]
\caption{Verification failure tags for the 363 GRPO training candidates with reward 0.4. Percentages use 363 as the denominator.}

\label{tab:app_partial_failures}
\begingroup
 \centering
\begin{tabularx}{\linewidth}{@{}>{\raggedright\arraybackslash}X r r@{}}
\toprule
Recorded label & Candidates & \% \\
\midrule
Model settings & 250 & 68.87 \\
Analysis process & 3 & 0.83 \\
Engineering conditions & 0 & 0.00 \\
Result/report consistency & 108 & 29.75 \\
Unclassified numerical difference & 2 & 0.55 \\
\bottomrule
\end{tabularx}
\par
\endgroup
\end{table}

Across epochs 1 to 3, physical model-setting violations decline steadily from 100 cases in epoch 1 to 80 in epoch 2 and 70 in epoch 3, indicating fewer violations of the specified engineering constraints on the sampled training candidates during reinforcement learning.

The training logs record reward summaries across all 57 updates on both GPU ranks. Figure~\ref{fig:training_dynamics}(a) plots their per-update means and exponential moving average with $\alpha=0.10$.

\section{Category-level evaluation outcomes}
\label{app:benchmark_provenance}
\label{app:category_results}

\begin{table}[htbp]
\caption{Complete task acceptance by category and stratum in the first evaluation round (SFT vs.\ GRPO accepted counts).}

\label{tab:appendix_category_counts}
 \centering
\begin{tabular}{@{}l r r r r r r r r r@{}}
\toprule
 & \multicolumn{3}{c}{Core} & \multicolumn{3}{c}{Structural} & \multicolumn{3}{c}{Engineering} \\
\cmidrule(lr){2-4}\cmidrule(lr){5-7}\cmidrule(l){8-10}
Cat. & $n$ & SFT & GRPO & $n$ & SFT & GRPO & $n$ & SFT & GRPO \\
\midrule
A1 & 28 & 25 & 27 & 20 & 0 & 0 & 10 & 0 & 0 \\
A2 & 30 & 10 & 14 & 20 & 0 & 0 & 11 & 0 & 0 \\
A3 & 30 & 18 & 27 & 20 & 0 & 0 & 10 & 1 & 3 \\
A4 & 30 & 27 & 27 & 17 & 9 & 9 & -- & -- & -- \\
B1 & 30 & 26 & 26 & 20 & 0 & 0 & 8 & 0 & 0 \\
B2 & 29 & 28 & 28 & 20 & 0 & 0 & 2 & 0 & 0 \\
B3 & 30 & 29 & 29 & 20 & 0 & 0 & 8 & 0 & 0 \\
B4 & 30 & 30 & 30 & 20 & 5 & 6 & 8 & 0 & 0 \\
C1 & 28 & 12 & 19 & 20 & 12 & 12 & 9 & 0 & 0 \\
C2 & 30 & 30 & 30 & 20 & 0 & 0 & 9 & 0 & 0 \\
C3 & 27 & 27 & 27 & 20 & 0 & 0 & 1 & 0 & 0 \\
C4 & 30 & 24 & 30 & 20 & 0 & 0 & -- & -- & -- \\
\midrule
Total & 352 & 286 & 314 & 237 & 26 & 27 & 76 & 1 & 3 \\
\bottomrule
\end{tabular}
\par
\end{table}

Table~\ref{tab:appendix_category_counts} reports complete task acceptance counts across all category--stratum combinations on the 665-task benchmark for the first evaluation round; see Section~\ref{subsec:benchmark_tasks} for benchmark composition. A dash indicates that the benchmark contains no task in that cell. Comparing task-level outcomes between the SFT and four-level GRPO checkpoints in the first evaluation round, the net gain of 31 accepted tasks comprises 28 core parameter-variation tasks, one structural-variation task and two engineering-context tasks.

\section{Stage-wise verification outcomes}
\label{app:acceptance_records}

Table~\ref{tab:appendix_acceptance_stages} reports cumulative pass counts across the five sequential evaluation gates analysed in Section~\ref{sec:results_failures}: syntactic validity, strict code-contract compliance, native solver execution, model delivery/reloading, and complete engineering acceptance. Unlike the mutually exclusive failure breakdown in Figure~\ref{fig:failure_reward_analysis}, each column in Table~\ref{tab:appendix_acceptance_stages} records the total number of tasks passing up to that verification gate out of the full stratum denominator. For the base Qwen3-8B and Qwen3-235B models, cumulative pass counts reach 0/665 (0.00\%) at the solver execution stage and all subsequent verification tiers (delivery and complete engineering acceptance), as invalid OrcFxAPI object or data-item references prevent the generated scripts from executing under the single-attempt evaluation protocol.

\begin{table}[htbp]
\caption{Cumulative stage pass counts in the first evaluation round across syntax (Syn.), strict code (Code), execution (Exec.), delivery/reload (Del.) and complete acceptance (Pass).}

\label{tab:appendix_acceptance_stages}
 \centering
\begin{tabularx}{\linewidth}{@{}X l r r r r r r@{}}
\toprule
Stratum & Policy & $n$ & Syn. & Code & Exec. & Del. & Pass \\
\midrule
All & SFT & 665 & 618 & 618 & 463 & 460 & 313 \\
 & GRPO & 665 & 622 & 610 & 471 & 471 & 344 \\
Core & SFT & 352 & 348 & 348 & 335 & 334 & 286 \\
 & GRPO & 352 & 347 & 347 & 344 & 344 & 314 \\
Structural & SFT & 237 & 209 & 209 & 126 & 125 & 26 \\
 & GRPO & 237 & 211 & 203 & 123 & 123 & 27 \\
Engineering & SFT & 76 & 61 & 61 & 2 & 1 & 1 \\
 & GRPO & 76 & 64 & 60 & 4 & 4 & 3 \\
\bottomrule
\end{tabularx}
\par
\end{table}

\section{Generation and verification protocol}
\label{app:generation_review}

All evaluated models use a decoding temperature of 0, with a maximum completion length of 16,384 tokens. Within each evaluation round, each model generates one program per task without simulator-feedback repair or access to reference solutions. Responses that exceed the token limit remain in the evaluation denominator and are recorded as code-check failures. The same model, analysis and output checks are applied to all evaluated models.

For the 100-task verification reliability audit (Section~\ref{sec:results_verifier_audit}, Table~\ref{tab:verification_audit}), tasks were selected prior to score inspection via proportional stratified sampling over the 665-task benchmark, producing 53 core parameter-variation tasks (four to five per category across A1--C4), 36 structural-variation tasks (three per category across A1--C4) and 11 engineering-context tasks (spanning A1, A2, A3, B1, B3, B4, C1 and C2). Across these three strata, solver execution succeeds on 51/53 (96.23\%), 19/36 (52.78\%) and 0/11 (0.00\%) tasks (70/100 overall), whereas complete engineering verification accepts 41/53 (77.36\%), 7/36 (19.44\%) and 0/11 (0.00\%) tasks (48/100 overall), matching the manual code-and-output audit on every stratum.

\section{Engineering sources, seeds and case evidence}
\label{app:case_sources}

\begin{table}[htbp]
\caption{Archived source catalogue for the 76 engineering-context tasks. Counts indicate the number of tasks citing each source; one task may cite multiple sources.}

\label{tab:source_catalogue_1}
\label{tab:source_catalogue_2}
 \centering
\begin{tabularx}{\linewidth}{@{}l >{\raggedright\arraybackslash}p{0.28\linewidth} r >{\raggedright\arraybackslash}X@{}}
\toprule
ID & Source title & Tasks & Public URL \\
\midrule
G03 & Line contact -- modelling & 1 & \url{https://www.orcina.com/webhelp/OrcaFlex/Content/html/Linecontact\%2CModelling.htm} \\
G04 & Modelling line feeding & 1 & \url{https://www.orcina.com/webhelp/OrcaFlex/Content/html/Modellinglinefeeding.htm} \\
G11 & C06 CALM buoy & 2 & \url{https://www.orcina.com/wp-content/uploads/examples/c/c06/C06\%20CALM\%20buoy.pdf} \\
G15 & Production riser examples & 1 & \url{https://www.orcina.com/Resources/examples/} \\
G16 & Mooring examples & 2 & \url{https://www.orcina.com/resources/examples/?key=c} \\
G26 & Line results -- contact and clearance & 2 & \url{https://www.orcina.com/webhelp/OrcaFlex/Content/html/Lineresults\%2CContact.htm} \\
G30 & OrcFxAPI and Python documentation & 2 & \url{https://www.orcina.com/resources/documentation/} \\
S01 & OrcaFlex examples -- Production risers & 24 & \url{https://www.orcina.com/Resources/examples/} \\
S03 & OrcaFlex examples -- Moorings & 23 & \url{https://www.orcina.com/resources/examples/?key=c} \\
S05 & OrcaFlex examples -- Pipelay and recovery & 9 & \url{https://www.orcina.com/resources/examples/?key=e} \\
S07 & OrcaFlex examples -- Deployment & 2 & \url{https://www.orcina.com/resources/examples/?key=g} \\
S13 & OrcaFlex User Group Meeting material & 9 & \url{https://www.orcina.com/Resources/ugm-material/} \\
S14 & OrcaFlex validation documents & 4 & \url{https://www.orcina.com/resources/validation/} \\
S15 & OrcaFlex QA, Testing and Validation & 3 & \url{https://www.orcina.com/wp-content/uploads/resources/validation/99-005-OrcaFlex-QA-Testing-and-Validation.pdf} \\
S17 & Browse Trunkline Installation Analysis & 5 & \url{https://atteris.com/resources/case-studies/concept-engineering-browse-pipelay/} \\
S18 & Veja Mate OWF cable installation study & 8 & \url{https://longitude-engineering.com/case-studies/veja-mate-owf-siem-aimery-cable-installation-study/} \\
S23 & Baltic Sea OWF inter-array cable study & 8 & \url{https://apollo.engineer/case-studies/installation-engineering-for-inter-array-cables-of-a-baltic-sea-offshore-wind-farm/} \\
S24 & The Fensfjorden Project & 5 & \url{https://bein.no/en/the-fensfjorden-project-nexans/} \\
S25 & Project Natick deployment and cable hook-up & 1 & \url{https://greenmarineuk.com/engineering-services/deployment-procedures-and-cable-hook-up/} \\
\bottomrule
\end{tabularx}
\par
\end{table}

\subsection{Scope and source provenance}

The 76 engineering-context tasks comprise 18 tasks adapted from published offshore engineering project descriptions and 58 inspired by official examples, modelling documentation, validation reports and user-group materials. These two classes are mutually exclusive at the task level and span static moorings and risers, dynamic wave--current responses, free-span pipelines, S-lay and cable-laying operations, and subsea lowering scenarios.

To form self-contained benchmark problems suitable for automated verification, the task specifications standardise coordinate frames, representative load cases, structural properties and installation geometries so that each case defines a complete numerical problem. Table~\ref{tab:source_catalogue_1} lists the archived sources, and Table~\ref{tab:engineering_case_source_mapping} maps each task to its sources.

\begin{table}[htbp]
\caption{Case-to-source mapping for the 76 engineering-context tasks (Table~\ref{tab:source_catalogue_1}).}

\label{tab:engineering_case_source_mapping}
 \centering
\begin{tabularx}{\linewidth}{@{}
l >{\raggedright\arraybackslash}X
l >{\raggedright\arraybackslash}X
l >{\raggedright\arraybackslash}X@{}}
\toprule
Case ID & Source IDs & Case ID & Source IDs & Case ID & Source IDs \\
\midrule
RE2-001 & S03, S07 & RE2-027 & G11 & RE2-053 & S13, S24 \\
RE2-002 & S03 & RE2-028 & G11, G30 & RE2-054 & S13 \\
RE2-003 & S03 & RE2-029 & G15 & RE2-055 & S13 \\
RE2-004 & S03 & RE2-030 & G03, G26 & RE2-056 & S13 \\
RE2-005 & S03, S07 & RE2-031 & G16 & RE2-057 & S13 \\
RE2-006 & S03 & RE2-032 & G04 & RE2-058 & S13, S24 \\
RE2-007 & S03 & RE2-033 & S01, S03 & RE2-059 & S05, S17 \\
RE2-008 & S03 & RE2-034 & S01, S03 & RE2-060 & S05 \\
RE2-009 & S03 & RE2-035 & S01 & RE2-061 & S05, S17 \\
RE2-010 & S03 & RE2-036 & S01, S03 & RE2-062 & S05 \\
RE2-011 & S03 & RE2-037 & S01, S03 & RE2-063 & S05 \\
RE2-012 & S03 & RE2-038 & S01 & RE2-064 & S05, S17 \\
RE2-013 & S03 & RE2-039 & S01, S13 & RE2-065 & S05 \\
RE2-014 & S03 & RE2-040 & S01, S03 & RE2-066 & S05, S17 \\
RE2-015 & S03 & RE2-041 & S03 & RE2-067 & S05, S17 \\
RE2-016 & S03 & RE2-042 & S03 & RE2-068 & S18, S23 \\
RE2-017 & S01, S15 & RE2-043 & S01, S14 & RE2-069 & S18, S23 \\
RE2-018 & S01 & RE2-044 & S01 & RE2-070 & S18, S23 \\
RE2-019 & S01, S14 & RE2-045 & S01 & RE2-071 & S18, S24 \\
RE2-020 & S01 & RE2-046 & S01 & RE2-072 & S18, S23 \\
RE2-021 & S01 & RE2-047 & S01, S14 & RE2-073 & S18, S23 \\
RE2-022 & S01, S15 & RE2-048 & S01 & RE2-074 & S23, S25 \\
RE2-023 & S01 & RE2-049 & S01, S14 & RE2-075 & S18, S23 \\
RE2-024 & S01, S15 & RE2-050 & S01 & RE2-076 & S18, S23 \\
RE2-025 & G26 & RE2-051 & S13, S24 & & \\
RE2-026 & G16, G30 & RE2-052 & S13, S24 & & \\
\bottomrule
\end{tabularx}
\par
\end{table}

\subsection{Archived seed examples and review coverage}
\label{app:seed_examples}

Table~\ref{tab:archived_seed_specifications} summarises three representative seed specifications: A1 single-catenary statics, B1 regular-wave catenary dynamics and C4 two-point subsea lowering. Each seed family undergoes manual modelling review, native solver execution, saved-model reload and independent output recomputation. For A1 \texttt{seed\_01} (73 nodes), static equilibrium yields effective tensions spanning 40.43--103.14~kN along the catenary profile.

To establish baseline discretisation settings at the seed-family level, mesh-refinement checks on B1 confirm a maximum tension variation of 0.035\% and a maximum displacement variation of 1.98\%, while time-step sensitivity tests on C4 across $\Delta t \in \{0.005, 0.0025, 0.00125\}$~s record a maximum first-contact timing variation of 0.00548~s. These discretisation baselines inform the numerical acceptance thresholds ($\pm 5\%$ for dynamic force and displacement quantities and $\pm 0.010$~s for contact timing) applied to expanded instances.

\begin{table}[htbp]
\caption{Key specifications of three representative seed examples (A1, B1 and C4).}

\label{tab:archived_seed_specifications}
 \centering
\begin{tabularx}{\linewidth}{@{}l >{\raggedright\arraybackslash}X@{}}
\toprule
Seed & Physical specification and requested analysis \\
\midrule
A1 \texttt{seed\_01} & Water depth 80~m; ends $(0,0,-8)$ and $(110,0,-80)$~m; length 145~m; $d=90$~mm; $m=95$~kg/m; $EA=800$~MN; 2~m segments. Still-water statics: end tensions, minimum elevation and maximum curvature with arc lengths. \\
B1 \texttt{seed\_01} & Water depth 80~m; 145~m chain ($d=80$~mm); regular wave ($H=2$~m, $T=8$~s, $0^\circ$); friction 0.50. Dynamic simulation (1 build-up + 15 steady periods): peak top tension and maximum horizontal excursion. \\
C4 \texttt{seed\_02} & Water depth 80~m; $6\times4\times3$~m payload (65~t, 24~m$^3$); two slings (2.5~m spacing); 3~s hover + 2~m payout at 0.06~m/s ($\Delta t=0.005$~s); seabed stiffness 1,000/300~kN/m$^3$. Outputs: first-contact time, peak force, penetration and line tensions. \\
\bottomrule
\end{tabularx}
\par
\end{table}

For the explicit A1 expansion example, record \texttt{A1\_seed\_01\_v001} changes the water depth to 100~m, the upper end to $(0,0,-12)$~m, the anchor horizontal distance to 145~m and the line length to 200~m. The outer diameter, mass and axial stiffness become 160.7~mm, 158.2~kg/m and 729.7~MN; the outer contact diameter is 306~mm, and compression is limited. The 2~m target segment length and 1,000~kN/m$^3$ seabed normal stiffness are retained. The following lines are an excerpt of the archived expanded response; the surrounding object creation, connections and result extraction remain in the complete record.
\begin{verbatim}
model.environment.WaterDepth = 100.0
line_type_1.OD = 0.1607
line_type_1.MassPerUnitLength = 0.1582
line_type_1.EA = 729700.0
line.Length = (200.0,)
model.CalculateStatics()
\end{verbatim}

\subsection{Representative-case specifications and numerical evidence}
\label{app:case_evidence}

The case records retain the task specification, generated Python script, execution status, verification outcomes and saved simulation files. Reference values are obtained from verified OrcaFlex simulations. Table~\ref{tab:case_response_evidence} summarises the numerical comparisons for the three representative cases analysed in Section~\ref{sec:results_cases}.

\begin{table}[htbp]
\caption{Recorded responses for the three representative cases (--- denotes failure; C4-039 denotes C4-FC1000-seed\_02-039).}

\label{tab:case_response_evidence}
 \centering
\begin{tabularx}{\linewidth}{@{}l >{\raggedright\arraybackslash}X r r r@{}}
\toprule
Case & Quantity & Ref. & SFT & GRPO \\
\midrule
RE2-018 & Top tension (kN) & 2,173.42 & --- & 2,173.42 \\
A2-B06 & Force at $22.5^\circ$ (kN) & $-1,026.96$ & $\simeq0$ & --- \\
 & Force at $67.5^\circ$ (kN) & $-622.85$ & $\simeq0$ & --- \\
 & Force at $112.5^\circ$ (kN) & $-622.85$ & $\simeq0$ & --- \\
 & Force at $157.5^\circ$ (kN) & $-1,026.96$ & $\simeq0$ & --- \\
C4-039 & First contact (s) & 36.0126 & 35.0825 & 36.0126 \\
 & Peak force (kN) & 75.5751 & 103.7909 & 75.5751 \\
 & Penetration (m) & 0.05918 & 0.09041 & 0.05918 \\
\bottomrule
\end{tabularx}
\par
\end{table}

RE2-018 specifies a 2,220~m uniform steel catenary riser in 1,200~m water depth, with a hinged top at $(0,0,-15)$~m and an anchored end at horizontal position $(1,350,0)$~m. The outer and inner diameters are 0.324 and 0.273~m, Young's modulus is 210~GPa, steel density is 7.85~t/m$^3$, and the internal and external water density is 1.025~t/m$^3$. The target segment length is 5~m, the seabed normal stiffness is 1,000~kN/m$^3$, and there is no environmental loading or static tangential friction. The GRPO policy's reported top tension is checked against independent recomputation from its reloaded model together with the model-input and applicable physical checks. The SFT response fails syntax validation at line 63 and supplies no numerical result.

A2-B06 specifies a four-point mooring in 130.6~m water depth, with fairleads at the four sign combinations of $(\pm52,\pm13.6,-9)$~m and anchors at $(\pm460,\pm300,-130.6)$~m. Each line has total unstretched length 529.5~m, outer diameter 0.0766~m, mass 0.1134~t/m and nominal axial stiffness 753,600~kN. Its terminal short section occupies 3.25\% of the specified last segment and has 0.925 times the parent axial stiffness. The calculation evaluates the origin and four offsets of radius 12~m at directions $22.5^\circ$, $67.5^\circ$, $112.5^\circ$ and $157.5^\circ$. Evaluation uses a 5\% relative tolerance and checks the saved model inputs. The SFT program fails both the response comparison and the input check because the line-segment list is misassigned. The GRPO policy raises a tuple-unpacking \texttt{ValueError} and does not deliver a valid restoring-force series.

For C4-FC1000-seed\_02-039, the payload dimensions, mass, volume and lifting-point separation match the C4 seed family, while the available payout is 5~m, the landing payout is 2~m and the reserved length is 2~m. The seabed normal/shear stiffness is 400/100~kN/m$^3$ and its friction coefficient is 0.60. The task prescribes a 0.005~s implicit time step, geometric zero-crossing for first contact, confirmation by contact force over 0.05~s, and a two-second post-contact peak window. The stored comparisons apply an absolute tolerance of 0.010~s to first-contact times; the peak contact force and penetration comparisons use the applicable 5\% relative rule. These tolerances are defined by the task specification.

All numerical comparisons in Table~\ref{tab:case_response_evidence} and Figure~\ref{fig:engineering_cases} are extracted directly from the saved native \texttt{.dat}/\texttt{.sim} simulation files and structured output summaries generated during benchmark evaluation.

\bibliographystyle{plainnat}
\bibliography{references}

\else
  {\centering
{\fontsize{17}{20}\selectfont\bfseries OceanCoder: A Verification-Guided Large Language Model\\for Automated Simulation Modelling in Ocean Engineering\par}
\vspace{5mm}
{\centering
\fontsize{10}{12}\selectfont
\begin{minipage}[t]{0.31\textwidth}
\centering
\textbf{Can Xue}\textsuperscript{\textdagger}\par
{\footnotesize
Ocean University of China\par}
\end{minipage}\hfill
\begin{minipage}[t]{0.31\textwidth}
\centering
\textbf{Changhong He}\textsuperscript{\textdagger,\textsection}\par
{\footnotesize
Ocean University of China\par
Beihang University\par}
\end{minipage}\hfill
\begin{minipage}[t]{0.31\textwidth}
\centering
\textbf{Ziyi Sun}\par
{\footnotesize
Ocean University of China\par}
\end{minipage}\par
\vspace{4mm}
\begin{minipage}[t]{0.31\textwidth}
\centering
\textbf{Anliang Ge}\textsuperscript{*}\par
{\footnotesize
Ocean University of China\par}
\end{minipage}\hfill
\begin{minipage}[t]{0.31\textwidth}
\centering
\textbf{Shimin Yu}\textsuperscript{*}\par
{\footnotesize
Ocean University of China\par}
\end{minipage}\hfill
\begin{minipage}[t]{0.31\textwidth}
\centering
\textbf{Bingchen Liang}\par
{\footnotesize
Ocean University of China\par}
\end{minipage}\par
\vspace{2mm}
{\footnotesize
\textsuperscript{\textdagger}Equal contribution\quad
\textsuperscript{\textsection}Project leader\quad
\textsuperscript{*}Corresponding authors\par}
}

}
\vspace{4mm}
{\centering\large\bfseries Abstract\par}
\begin{quote}
Numerical simulation is important in ocean engineering, but model construction requires expert knowledge. Large language models support automated modelling, but code syntax and successful execution alone do not establish whether generated models and results meet engineering requirements. We present OceanCoder, a Qwen3-8B-based large language model for automated simulation modelling in ocean engineering. To address specialised programming requirements and scarce training data, we develop an extensible data construction pipeline producing 12,000 verified supervised fine-tuning (SFT) examples across 12 engineering task categories. A three-layer verification framework checks model inputs against engineering parameters and task constraints, verifies completion of required simulation analyses, and assesses the engineering plausibility of computed outputs. Its outcomes provide graded rewards for group relative policy optimisation (GRPO), guiding the model towards engineering task completion. We also construct the first benchmark for automated modelling in ocean engineering, covering out-of-distribution tasks and real-world engineering cases. Experiments and ablation studies demonstrate the effectiveness of the data pipeline, verification framework and reward design. OceanCoder outperforms both base models and the closed-source comparison, with Pass@1 increasing from 0\% for the base model to 46.77\% after SFT and 51.43\% after GRPO. These results support engineering verification as a training signal for automated simulation modelling.
\end{quote}
{\small\noindent\textbf{Keywords:} Ocean engineering; Automated simulation modelling; Large language models; Engineering verification; Reinforcement learning.\par}

  \section{Introduction}
\label{sec:introduction}

Numerical simulation helps engineers assess loads and motions in mooring systems, marine risers and subsea installation operations \citep{Hall2015MooringValidation,Jae2022MooOpT4FPSO,Kallinger2026HolisticMooring}. Building these models still requires specialist knowledge. Users must define component connections and physical properties, set environmental conditions, and choose the analysis steps and results to extract. A change in structural configuration or operating sequence can require corresponding changes to the program, even when an existing script is available. Generating simulation programs from engineering task descriptions could reduce this preparation effort and make specialised software easier to use.

Large language models (LLMs) offer a way to translate such descriptions into programs. Studies in ship design, finite-element analysis and computational fluid dynamics (CFD) show how natural-language requirements can be converted into design parameters or simulation workflows \citep{Choi2026LLMasUI,Hou2025AutoFEA,Fan2026ChatCFD}. However, an executable program may still perform a different analysis from the one requested. A solver can run successfully after a program has changed a line property or extracted the wrong response quantity. Assessing automated modelling therefore requires checking whether the model and its results meet the engineering requirements, in addition to checking code syntax and execution.

We present OceanCoder, a Qwen3-8B-based large language model for automated simulation modelling in ocean engineering. OceanCoder takes a natural-language description of engineering conditions and analysis requirements and generates a complete OrcaFlex program to build the model, run the analyses and extract the results. We develop this capability by first training on verified examples and then using engineering verification of generated programs to provide further training feedback.

Training OceanCoder requires paired examples of an engineering task and a program that correctly implements it. Such public data for specialised ocean engineering simulation software are scarce. Seed-based generation can expand training examples \citep{Wei2024Magicoder,Wei2024SelfCodeAlign}, but engineering parameters must remain consistent. For example, water depth, endpoint coordinates and line lengths in a catenary model must describe a compatible geometry. We therefore build a data construction pipeline that expands expert-designed seed cases within engineering constraints and generates each task description and reference program from the same physical parameter record. Each pair is verified before entering the training corpus. The resulting corpus contains 12,000 supervised fine-tuning (SFT) examples across 12 categories of static analysis, dynamic response and offshore operations. The corpus can be extended by adding seed cases with their parameter constraints and verification rules.

Training on verified examples cannot prevent all mistakes in new programs. In Figure~\ref{fig:motivation}, an SFT-generated lowering program changes the prescribed line properties. It completes the simulation, but its error in predicting the first seabed-contact time exceeds the task tolerance. To identify errors of this kind, we introduce a three-layer engineering verification framework covering inputs, processes and outputs. It checks the constructed model against the given parameters and task constraints, confirms that the required analyses have been completed, and checks the reported response quantities and their numerical values against task-specific criteria. These checks assess whether the computed results are consistent with the specified engineering problem. A successful run is accepted as a completed task only when it passes the engineering checks.

The verification results also provide feedback for improving the model. Earlier code-generation studies have used program outcomes and execution feedback as reinforcement-learning signals \citep{Le2022CodeRL,Gehring2025RLEF}. After SFT, we use the engineering verification outcomes as rewards for online group relative policy optimisation (GRPO). Programs that fail to run, run but fail engineering checks, or complete the task receive different rewards. This graded feedback retains information about partial progress while reserving full reward for complete task acceptance. During training, generated programs run in OrcaFlex, and their verification results supply the rewards used to update the model.

We also construct the first benchmark for automated modelling in ocean engineering, covering out-of-distribution tasks and real-world engineering cases to assess generalisation and practical applicability. Its 665 tasks cover core parameter variations, structural variations and engineering cases adapted from public project descriptions and technical materials. Each model generates one program per task without using simulator feedback to repair it. All programs are evaluated using the same engineering acceptance criteria. This protocol assesses the modelling capability acquired through training. The benchmark and its evaluation resources will be released together.

The experiments compare OceanCoder with base and leading closed-source models under this protocol. Independent manual audits compare the verifier's decisions with human inspection of generated programs and results. Controlled reward ablations examine the contribution of intermediate verification feedback to training. OceanCoder outperforms the evaluated comparison models, with the proportion of tasks completed on the first attempt (Pass@1) increasing from 46.77\% after SFT to 51.43\% after GRPO.

\begin{figure}[htbp]
\centering
\includegraphics[width=\linewidth]{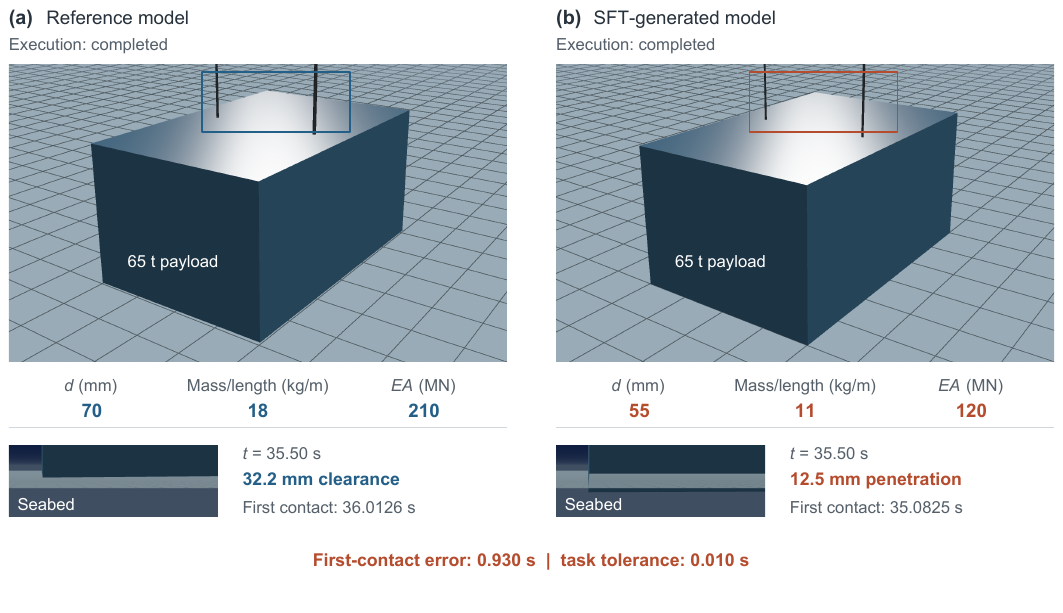}
\caption{The verification gap between executable code and complete engineering task acceptance. In a two-point lowering task involving a 65~t payload, the reference and OceanCoder programs after SFT both execute, but their line properties and seabed-contact responses differ. At $t=35.50$~s, the reference retains 32.2~mm clearance while the SFT model has 12.5~mm penetration. The predicted first-contact times differ by 0.9301~s, exceeding the 0.010~s task tolerance. Here, $d$ denotes line diameter and $EA$ axial stiffness; the reference is a checked simulation solution.}
\label{fig:motivation}
\end{figure}

The main contributions are as follows.
\begin{enumerate}
\item \textbf{OceanCoder for automated simulation modelling.} A Qwen3-8B-based model translates natural-language engineering requirements into complete OrcaFlex programs for model construction, analysis execution and result extraction.
\item \textbf{Verified and extensible training data.} A seed-based pipeline preserves engineering parameter relationships and agreement between task descriptions and reference programs, providing 12,000 verified examples across 12 task categories.
\item \textbf{Engineering verification as training feedback.} A framework checks model inputs, analysis completion and computed outputs, and turns the outcomes into graded GRPO rewards that guide model training.
\item \textbf{Benchmark and evaluation protocol.} A 665-task benchmark applies common engineering checks to all evaluated models. Manual audits and controlled reward ablations assess verifier reliability and the contribution of training feedback.
\end{enumerate}

\section{Related work}
\label{sec:related_work}

\subsection{Large language models for engineering simulation}
\label{subsec:rw_llm_sim}

Automating engineering simulation involves translating natural-language requirements into solver inputs and coordinating numerical analyses. Language models support these tasks through structured interfaces, retrieval and agent coordination. LLM-as-UI converts user requests into parameters for modular ship design \citep{Choi2026LLMasUI}. AutoFEA combines planning and example retrieval to generate finite element solver inputs \citep{Hou2025AutoFEA}. For computational fluid dynamics (CFD), OpenFOAMGPT and ChatCFD use domain knowledge and iterative correction to configure simulation cases \citep{Pandey2025OpenFOAMGPT,Fan2026ChatCFD}. MetaOpenFOAM 2.0 and Foam-Agent organise simulation and post-processing through task decomposition and agent coordination \citep{Chen2025MetaOpenFOAM2,Yue2026FoamAgent}. LLM-based residual dynamics also augment marine-vessel predictive control \citep{Zeng2026LLMMPC}. These studies assign language models specific roles in interpreting requirements, constructing cases and supporting engineering decisions.

Domain-specific supervision supports different engineering tasks. OceanGPT uses ocean-domain instruction data \citep{Bi2024OceanGPT}. ALL-FEM fine-tunes models on verified FEniCS scripts for finite element workflows \citep{Deotale2026ALLFEM}. Simulation-generated data also train neural networks to predict riser fatigue \citep{Sivaprasad2023RiserFatigue}. Program generation requires reliable pairs of engineering requirements and executable implementations. Self-Instruct generates and filters instructions, Magicoder constructs tasks from source-code snippets, and SelfCodeAlign selects generated solutions through execution tests \citep{Wang2023SelfInstruct,Wei2024Magicoder,Wei2024SelfCodeAlign}. For engineering simulation, execution testing must be accompanied by consistency between physical parameters, task descriptions and program implementations. When a task contains coupled parameters, independent changes to the description or program can make an otherwise executable example inconsistent with the specified engineering problem. To preserve this consistency during data expansion, OceanCoder generates descriptions and reference OrcaFlex programs from shared physical parameter records, subject to engineering constraints.

\subsection{Engineering verification and learning from feedback}
\label{subsec:rw_verification_rl}

Successful execution does not establish whether a program satisfies its intended requirements. The test-oracle problem concerns how to judge the correctness of observed program behaviour \citep{Barr2015OracleProblem}. EvalPlus shows that expanded tests detect functional errors missed by the original test suites \citep{Liu2023EvalPlus}. In optimisation modelling, even agreement with a reference objective value can conceal an incorrect constraint set. Constraint injection addresses this problem by checking feasible and constraint-violating solutions alongside objective values \citep{Luo2026ConstraintInjection}. Computational verification addresses both code implementation and numerical-solution accuracy \citep{Oberkampf2002VV}. These distinctions motivate checking the engineering specification as well as program behaviour.

The same verification criteria can guide learning as well as evaluation. Constraint injection uses verification outcomes for synthetic-data filtering and reinforcement-learning rewards \citep{Luo2026ConstraintInjection}. CodeRL combines unit-test outcomes, graded rewards and a learned critic \citep{Le2022CodeRL}. RLEF trains models to use execution feedback during iterative refinement \citep{Gehring2025RLEF}. CODERL+ uses variable-level execution feedback \citep{Jiang2026CodeRLPlus}. OraclePhys examines how different forms of finite-element-based supervision and verification feedback affect learning in structural mechanics \citep{Li2026OraclePhys}. The appropriate feedback depends on the task. Unit-test outcomes, formulation constraints and structural responses each assess different requirements. For OrcaFlex modelling, the relevant checks concern the constructed physical system, completion of the required analyses and extraction of the requested responses. OceanCoder's three-layer verification framework performs these checks and distinguishes execution progress from complete task acceptance. The resulting verification states provide graded rewards for online GRPO, linking model updates to engineering task completion.

\subsection{Benchmarks for automated simulation modelling}
\label{subsec:rw_benchmarks}

Benchmarks differ in the tasks they assign and the evidence they use to judge success. DS-1000 evaluates programming with scientific Python libraries, and BigCodeBench tests diverse function calls and complex instructions \citep{Lai2023DS1000,Zhuo2025BigCodeBench}. SciCode assesses research programming problems curated by scientists \citep{Tian2024SciCode}. Simulation benchmarks extend this evaluation to numerical software and its outputs. FEABench checks COMSOL-based multiphysics models, physical settings and numerical results \citep{Mudur2025FEABench}. CFDLLMBench covers CFD knowledge, solver implementation and OpenFOAM workflows \citep{Somasekharan2026CFDLLMBench}. SimulCost evaluates simulation parameter tuning under accuracy and computational-cost constraints \citep{Cao2026SimulCost}. LiveCodeBench additionally evaluates code generation, self-repair, execution and output prediction using time-segmented problems \citep{Jain2025LiveCodeBench}. These settings assess distinct programming and simulation requirements.

Interaction budgets further affect what a benchmark measures. A single-attempt setting assesses the initial response, whereas feedback-based settings also assess error diagnosis and correction. To measure the modelling capability acquired through training, the OceanCoder benchmark fixes the interaction budget. Each model makes one attempt to generate a complete OrcaFlex program for each task, without simulator-feedback repair. During evaluation, verification outcomes determine acceptance without being returned to the model for another generation attempt. Its 665 tasks cover parameter variations, structural variations and engineering contexts. Common checks of model inputs, analysis completion and requested outputs define acceptance for every model. The resulting comparisons separate performance across task variations while retaining a consistent definition of engineering task completion.

\section{OceanCoder framework and training procedure}
\subsection{Task definition and framework overview}
\label{subsec:framework_overview}

\begin{figure}[htbp]
\centering
\includegraphics[width=\linewidth]{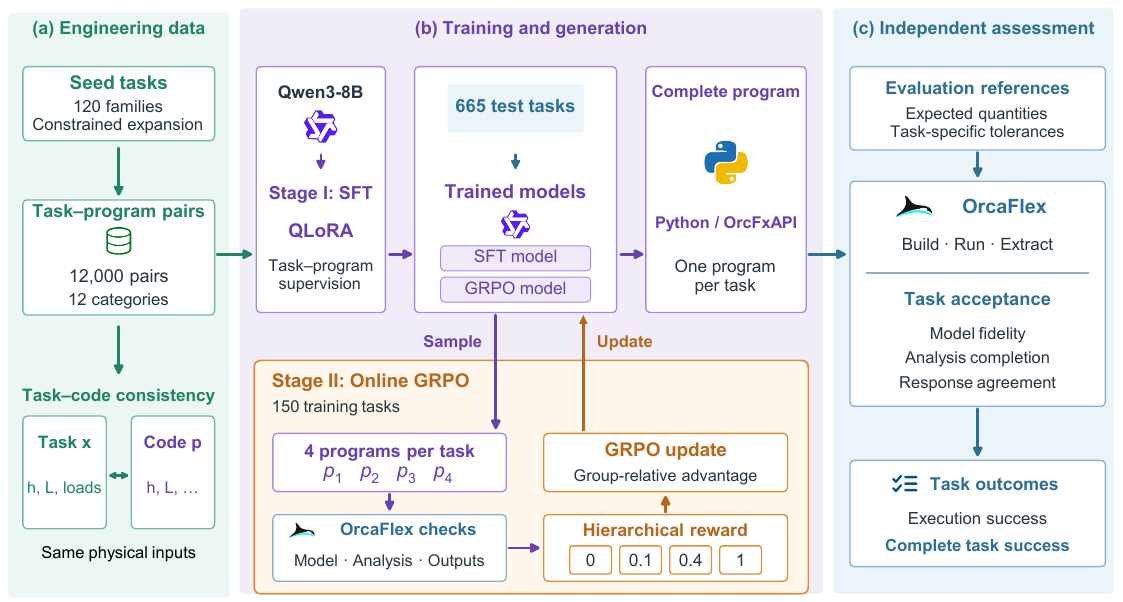}
\caption{Overview of the OceanCoder framework for engineering-constrained data construction, supervised fine-tuning, online GRPO and independent evaluation.}
\label{fig:framework}
\end{figure}

OceanCoder addresses the task of generating complete OrcaFlex programs from natural-language specifications of physical systems, required analyses and requested responses. Each program uses OrcFxAPI, the Python interface to OrcaFlex, to build the model, run the analyses and extract the results \citep{OrcinaPythonInterface}. Task completion requires agreement with the engineering specification throughout this process, including the model configuration, analysis steps and reported responses.

Figure~\ref{fig:framework} presents the data construction, training and evaluation framework. We expand expert-designed seed cases within engineering constraints and use the same parameters to generate each task description and reference program. Verified pairs are used for supervised fine-tuning (SFT) of Qwen3-8B. Since programs generated after SFT can still alter specified inputs or extract unintended responses, subsequent training uses engineering verification of generated programs as feedback. Three-layer verification checks model inputs, analysis completion and outputs. The verification outcomes determine graded rewards for online group relative policy optimisation (GRPO). The trained OceanCoder checkpoints and comparison models are evaluated on a separate benchmark using the same engineering acceptance criteria. Each model makes one attempt to generate a complete program for each task, without simulator-feedback repair.

\subsection{Engineering-constrained data construction}
\label{subsec:data_construction}

We construct training examples by expanding expert-designed seed cases, as shown in Figure~\ref{fig:data_pipeline}. A seed defines the physical objects and their connections, the analysis procedure and the response quantities to extract. These elements establish the task family within which additional cases are generated. The corpus covers 12 categories grouped into static analysis, dynamic response and offshore operations, as listed in Table~\ref{tab:task_categories}. Each category contains ten seed families, with 100 expanded examples per family, producing 1,000 examples per category and 12,000 examples in total. The seed therefore provides both an engineering specification and a complete program structure that can implement admissible variations of that specification.

\begin{table}[htbp]
\small
\caption{Task coverage of the supervised training corpus (1,000 pairs per category; 12,000 pairs across 12 categories). SCR denotes a steel catenary riser.}
\label{tab:task_categories}
 {\centering
\begin{tabularx}{\linewidth}{@{}l l >{\raggedright\arraybackslash}X@{}}
\toprule
Group & ID & Engineering task \\
\midrule
Static & A1 & Single-catenary static equilibrium \\
analysis & A2 & Four-point mooring static equilibrium \\
 & A3 & SCR static configuration and touchdown \\
 & A4 & Pipeline lateral stability \\
\midrule
Dynamic & B1 & Single catenary under regular waves \\
response & B2 & Coupled four-point mooring dynamics \\
 & B3 & SCR response to motion, waves and current \\
 & B4 & Free-span pipeline wave--current response \\
\midrule
Offshore & C1 & Static S-lay pipeline installation \\
operations & C2 & Dynamic laying of a flexible cable \\
 & C3 & Single-point lifting and lowering \\
 & C4 & Two-point lowering and seabed contact \\
\bottomrule
\end{tabularx}
\par}
\end{table}

\begin{figure}[htbp]
\centering
\includegraphics[width=\linewidth]{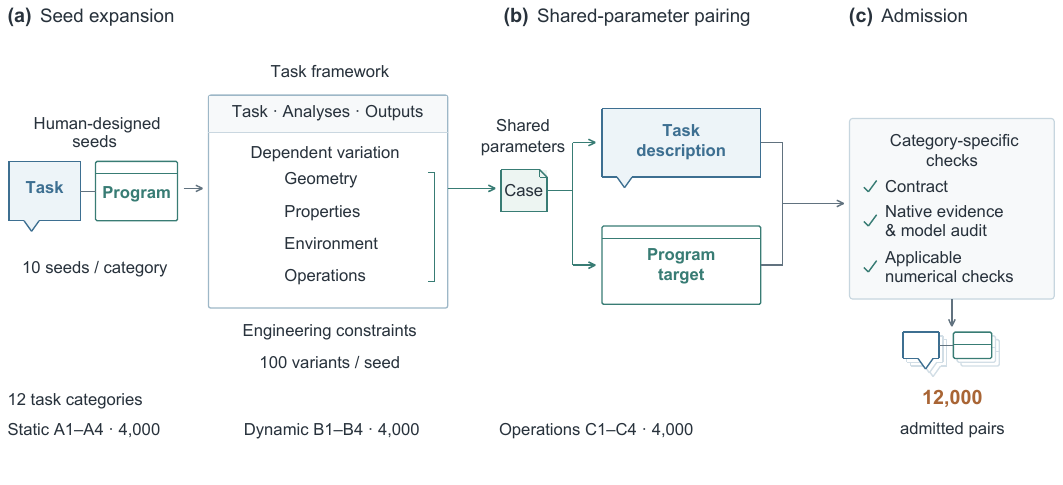}
\caption{Engineering-constrained data construction pipeline for expanding expert-designed seed cases, generating task descriptions and reference programs from shared parameters, and verifying examples before training.}
\label{fig:data_pipeline}
\end{figure}

Parameter variation preserves the relationships required by each task. For a catenary case, water depth, endpoint positions and line length jointly determine the geometry to be modelled. Line diameter, unit mass and axial stiffness follow the combinations allowed by the seed rules. For a lowering task, the payout rate and stage durations must agree with the prescribed operation. These dependencies restrict which parameter combinations can be used together. The ranges in Table~\ref{tab:parameter_ranges} describe the marginal coverage of the resulting corpus. They are obtained from constrained cases and coupled component-property packages.

A shared parameter record connects each expanded specification to its reference implementation. Once a case is selected, the same values populate the task description and the corresponding model assignments in the program. The expanded A1 case in Table~\ref{tab:a1_example}, for example, specifies a water depth of 100~m and a line length of 200~m. These values appear in the description and are assigned to the native OrcaFlex model. The record also determines the endpoint coordinates, line properties and contact settings. The analysis and extraction steps retain the seed's task definitions, including the requested response quantities and their locations. This construction keeps the language requirement, physical model and output requirements consistent as the parameters change.

\begin{table}[htbp]
\small
\caption{Marginal ranges of principal physical, environmental and operational parameters across the 12,000-pair SFT training corpus (1,000 pairs per category). Ranges reflect constrained within-category variations and coupled component property packages rather than independent Cartesian combinations.}
\label{tab:parameter_ranges}
\begingroup
 {\centering
\begin{tabularx}{\linewidth}{@{}>{\raggedright\arraybackslash}p{.12\linewidth} >{\raggedright\arraybackslash}p{.25\linewidth} >{\raggedleft\arraybackslash}p{.15\linewidth} >{\raggedright\arraybackslash}X@{}}
\toprule
Input group & Parameter (unit) & Corpus range & Engineering scope and coupling constraints \\
\midrule
Geometry & Water depth, $h$ ($\mathrm{m}$) & 15--1,975 & Shallow-to-intermediate pipelines, moorings and installation tasks (15--380~m); deepwater steel catenary risers (450--1,975~m). \\
 & Total line / pipeline length, $L$ ($\mathrm{m}$) & 45--3,015 & Scaled with water depth and hang-off geometry across lifting slings (45--75~m), catenary moorings and pipelay (130--1,031~m) and deepwater SCRs (920--3,015~m). \\
Structure & Outer / nominal diameter, $d$ ($\mathrm{mm}$) & 44--610 & Coupled within component property packages: hoisting wires and subsea cables (44--160~mm), mooring chains and ropes (70--360~mm) and offshore steel pipelines/risers (219.1--610~mm). \\
 & Bare unit mass, $m_{\mathrm{bare}}$ ($\mathrm{kg/m}$) & 6.5--366.2 & Bare structural mass across cables (6.5--60~kg/m), mooring lines (52--286.6~kg/m) and bare steel pipes (70.8--366.2~kg/m); in coated pipeline stability tasks (A4), adding concrete weight coating yields total coated mass $m_{\mathrm{coated}}$ of 109.7--1,393.7~kg/m. \\
 & Axial stiffness, $EA$ ($\mathrm{MN}$) & 60--10,049 & Assigned by structural member type to preserve realistic axial elongation across slings and cables (60--400~MN), mooring lines (82--1,674~MN) and thick-walled steel pipes (1,894--10,049~MN). \\
Environment & Wave height, $H$ / $H_s$ ($\mathrm{m}$) & 0--8.0 & Still water (0~m) in static equilibrium and controlled lowering analyses; regular waves (1.08--8.0~m) and irregular JONSWAP seas ($H_s=2.52$--3.30~m) in dynamic tasks. \\
 & Wave period, $T$ / $T_p$ ($\mathrm{s}$) & 6.5--15.0 & Regular wave periods (6.5--15.0~s) and spectral peak periods ($T_p=10$--11~s) paired with wave height and water depth to maintain realistic wave steepness. \\
 & Current speed, $V_c$ ($\mathrm{m/s}$) & 0.00--1.20 & Zero in still-water analyses; steady currents of 0.20--1.20~m/s across incident directions ($0^\circ$--$355^\circ$) in combined wave--current loading scenarios. \\
Seabed \& ops & Seabed normal stiffness, $k_n$ ($\mathrm{kN/m^3}$) & 400--7,792 & Linear elastic seabed contact stiffness (with Coulomb friction coefficients of 0.20--0.69), complemented by nonlinear soil resistance models in deepwater riser touchdown analyses. \\
 & Payloads, tensions \& speeds & Task-specific & Lifted payloads (30--115~t), moored displacement vessels (1,170--15,970~t), S-lay tensioner capacities (120--850~kN) and winch/lay speeds (0.03--0.89~m/s) matched to stage durations. \\
\bottomrule
\end{tabularx}
\par}
\endgroup
\end{table}

\begin{table}[htbp]
\small
\caption{Comparison of an original A1 seed (\texttt{seed\_01}) and an expanded supervised training instance (\texttt{A1\_seed\_01\_v001}). English summaries retain the exact physical inputs of the Chinese task statements.}
\label{tab:a1_example}
\begingroup
 {\centering
\begin{tabularx}{\linewidth}{@{}l >{\raggedright\arraybackslash}X >{\raggedright\arraybackslash}X@{}}
\toprule
Component & Original seed: A1, \texttt{seed\_01} & Expanded pair: \texttt{A1\_seed\_01\_v001} \\
\midrule
Task & Construct one catenary, solve static equilibrium, save the model and extract the specified responses. & The same task family and analysis and output requirements, with expanded physical inputs. \\
Geometry & Water depth 80 m; fixed A at $(0,0,-8)$ m; anchored B at $(110,0)$ m and zero height above the seabed; line length 145 m. & Water depth 100 m; fixed A at $(0,0,-12)$ m; anchored B at $(145,0)$ m and zero height above the seabed; line length 200 m. \\
Line properties & Equivalent outer diameter 90 mm; mass 95 kg/m; axial stiffness 800 MN. & Equivalent outer diameter 160.7 mm; mass 158.2 kg/m; axial stiffness 729.7 MN. \\
Contact and mesh & Default contact diameter (inherited from outer diameter) and standard axial compression; target segment length 2 m. & Parameterized contact diameter (306.0 mm) and compression-limit switch activated within the seed's admissible parameter space; target segment length 2 m. \\
Common conditions & \multicolumn{2}{p{0.77\linewidth}@{}}{Still water, zero current, flat elastic seabed and normal stiffness $1000\ \mathrm{kN/m^3}$. Unspecified inputs retain the stated empty-model defaults.} \\
Program excerpts & {\ttfamily model.environment.WaterDepth = 80.0\par
line.Length = (145.000,)\par
model.CalculateStatics()} & {\ttfamily model.environment.WaterDepth = 100.0\par
line.Length = (200.0,)\par
model.CalculateStatics()} \\
Outputs & \multicolumn{2}{p{0.77\linewidth}@{}}{\texttt{model.dat} and \texttt{results.json}: end effective tensions; minimum elevation and its arc length; maximum curvature and its arc length; model state and warnings.} \\
Verified response & Static equilibrium solved and verified via model reload (73 nodes; top effective tension 103.144 kN; touchdown contact confirmed on the elastic seabed). & Static equilibrium solved and verified via model reload (101 nodes; top effective tension 152.703 kN; touchdown contact confirmed on the elastic seabed). \\
\bottomrule
\end{tabularx}
\par}
\endgroup
\end{table}

Each pair undergoes verification before admission to the corpus. Static checks use the program's abstract syntax tree (AST) to inspect its structure and required operations. Native OrcaFlex execution then establishes whether the prescribed analyses complete and the required files are produced. The saved model is reloaded so that its actual settings can be compared with the task specification. Applicable responses are independently recomputed from the saved results and checked against the reported outputs. These stages inspect complementary evidence about the program, the model it constructs and the quantities it reports. Appendix~\ref{app:admission_verification} details the checks for each category.

An admitted example contains a system instruction, a natural-language task description and a complete Python program. The system instruction specifies the program format and the execution and output requirements. Its seed family and parameter values are retained to document its construction, and Appendix~\ref{app:training_records} summarises the corpus composition. Additional task categories can follow the same procedure once their seed cases, parameter constraints and verification rules have been defined.

\subsection{Stage I: supervised fine-tuning}
\label{subsec:sft}

We fine-tune Qwen3-8B \citep{Yang2025Qwen3} on the verified task--program pairs to establish its ability to generate complete OrcaFlex programs. Each target response includes model construction, analysis execution and result extraction. Training minimises the negative log-likelihood of the reference program tokens, averaged over the supervised response tokens. System instructions, user requirements and padding tokens are excluded from the loss. This supervision trains the model to generate programs that implement the specified model configuration, analysis steps and result extraction.

Fine-tuning uses quantised low-rank adaptation (QLoRA) with 4-bit NormalFloat (NF4) quantisation \citep{Dettmers2023QLoRA}. Low-rank adapters in the attention and feed-forward layers are trained while the base weights remain frozen \citep{Hu2022LoRA}. The selected SFT checkpoint initialises GRPO. A separate copy of this checkpoint, including its trained adapters, remains fixed and supplies the reference token probabilities used to compute the Kullback--Leibler (KL) penalty in Section~\ref{subsec:grpo} (see Appendix~\ref{app:executed_hyperparameters} for training logs and compute resources).

\subsection{Three-layer engineering verification}
\label{subsec:verification}

Engineering verification determines whether a generated program fulfils its task by examining the saved OrcaFlex model, analysis records and numerical outputs. The input layer checks the physical system constructed by the program. Geometry, component properties, object connections, environmental loads and boundary conditions are compared with the engineering specification. The saved model records the settings that the solver actually used. Its inspection covers assignments to individual line segments, connection positions and restraint settings, as well as properties shared across a complete component. This allows the checks to distinguish a valid calculation from a calculation performed with an incorrectly configured system.

The process layer establishes whether the required analyses have been completed. Static tasks require a solved equilibrium state. Dynamic tasks additionally require completion of the prescribed simulation period, while operational tasks require the specified sequence of analysis stages and motions. The verifier uses execution records and saved analysis states to establish completion. This connects the analysis requirement to the actual solution state, including tasks in which several stages must be completed before a requested response is available.

The output layer checks both the meaning and the numerical value of each response. A result must represent the requested physical quantity at the specified spatial position and over the prescribed time interval. For example, a directional restoring force requires the correct projection and sign, and a line response requires the correct arc-length location. The verifier recomputes applicable quantities from the saved results before comparing them with the reference values. A correctly configured model can therefore still fail acceptance if its program reports the wrong quantity or extracts a response from an unintended location or interval.

Numerical comparison accounts for equivalent representations and numerical precision. Periodic angles are normalised before comparison, and non-physical identifiers are excluded from numerical checks. Thus, equivalent directions around the $0^\circ/360^\circ$ boundary and different metadata labels can be handled consistently. Absolute or relative tolerances are specified for individual response quantities. Where the task specification includes an absolute threshold near zero, that threshold governs comparisons of small values. Mesh and time-step sensitivity checks on the seed cases inform the selected tolerances, as documented in Appendix~\ref{app:seed_examples}. Complete acceptance requires all applicable input, process and output checks to pass. This joint decision preserves the relationship between the prescribed physical problem and the reported response.

The lowering task in Figure~\ref{fig:motivation} illustrates how the layers work together. Input checks examine the prescribed hoisting-line properties and connections, while process checks establish completion of the lowering sequence. First seabed contact is identified from the geometric zero crossing and confirmed by contact force over a 0.05~s window. The peak upward contact force is evaluated within the following two-second window. The first-contact time is compared with the reference using an absolute tolerance of 0.010~s, while peak contact force and penetration use the task's 5\% relative tolerance. The SFT program has a first-contact-time error of 0.9301~s, exceeding the prescribed tolerance despite successful execution. These checks connect the specified model to the analysis performed and the responses reported. The engineering acceptance criteria are used for GRPO feedback and benchmark evaluation. Verification of reference training pairs before SFT is described separately in Section~\ref{subsec:data_construction}. Category-specific checks are detailed in Appendix~\ref{app:admission_verification}.

\begin{figure}[htbp]
\centering
\includegraphics[width=\linewidth]{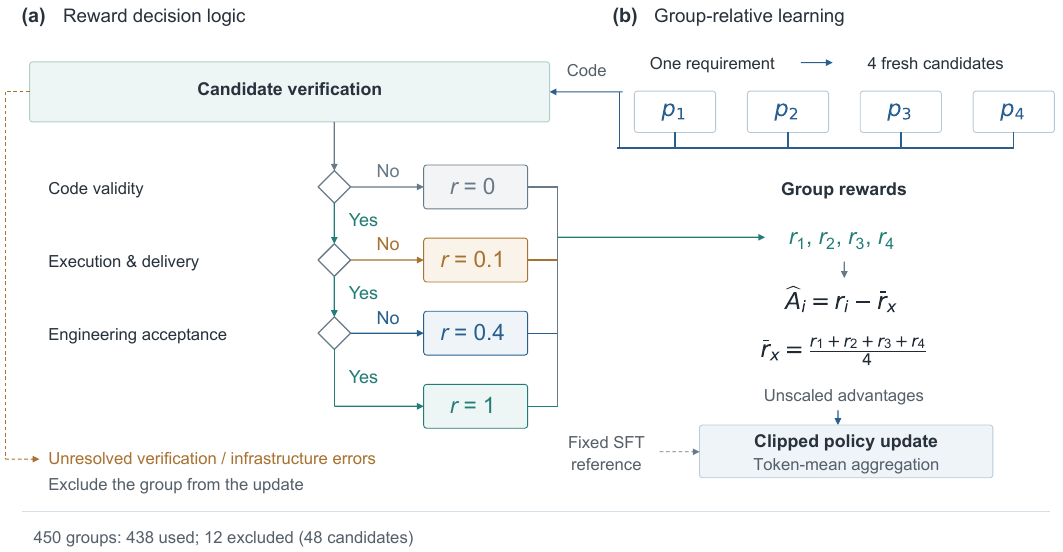}
\caption{Verification-guided GRPO training in OceanCoder. Four candidate programs are generated for each engineering requirement. Execution and engineering verification provide graded rewards for policy updates.}
\label{fig:reward_grpo}
\end{figure}

\subsection{Stage II: closed-loop reinforcement learning}
\label{subsec:grpo}

Online GRPO connects program generation to engineering verification through repeated rounds of sampling, simulation and model updating, as shown in Figure~\ref{fig:reward_grpo}. Each round begins with a batch of engineering requirements. The current model generates four candidate programs for each requirement and remains fixed while those candidates are evaluated. Candidate programs first undergo format, syntax and static safety checks. Those that pass proceed to OrcaFlex execution and engineering verification. The resulting verification states determine the rewards in Table~\ref{tab:reward_levels}. This feedback addresses errors produced by the current model on the engineering tasks used for training.

\begin{table}[htbp]
\small
\centering
\caption{Four-level rewards based on program execution and engineering task acceptance.}
\label{tab:reward_levels}
 {\centering
\begin{tabularx}{\linewidth}{@{}cX@{}}
\toprule
Reward & Established outcome \\
\midrule
0 & Format, syntax or static safety checks fail. \\
0.1 & Code checks pass, but execution, file delivery or model reloading fails. \\
0.4 & Execution, delivery and reloading pass, but engineering acceptance is incomplete. \\
1 & All applicable engineering acceptance checks pass. \\
\bottomrule
\end{tabularx}
\par}
\end{table}

The reward hierarchy preserves progress towards complete task acceptance. Within each four-candidate group, programs may fail at different stages, so intermediate rewards can provide a learning signal even when no candidate completes the task. Following the group-relative comparison in GRPO \citep{Shao2024DeepSeekMath}, we compare each reward with the group mean to obtain the advantage in Eq.~\eqref{eq:group_advantage},
\begin{equation}
\widehat{A}_i
= r_i - \frac{1}{4}\sum_{j=1}^{4}r_j,
\label{eq:group_advantage}
\end{equation}
where $r_i$ is the reward of candidate $i$, and $\widehat{A}_i$ is its advantage relative to the group mean. The advantages are not divided by the group standard deviation, following the corresponding unscaled-advantage formulation \citep{Liu2025DrGRPO}. Groups with unresolved verification outcomes are excluded from optimisation.

The policy update combines a clipped probability-ratio objective \citep{Schulman2017PPO} with a KL penalty relative to the fixed SFT policy. Three policies have distinct roles in this calculation. The sampling policy supplies the candidate programs and their original token probabilities. The current policy is updated using those candidates. The reference policy retains the selected SFT checkpoint, including its trained adapters, and provides the regularisation target. Table~\ref{tab:policy_notation} summarises the notation used below.

For token $p_{i,t}$ conditioned on $q_{i,t}$, Eq.~\eqref{eq:policy_ratio} gives the ratio between the current and sampling policies,
\begin{equation}
\rho_{i,t}=\frac{\pi_\theta(p_{i,t}\mid q_{i,t})}
{\pi_{\mathrm{old}}(p_{i,t}\mid q_{i,t})}.
\label{eq:policy_ratio}
\end{equation}
The group advantage is shared across the response tokens of each candidate. The loss in Eq.~\eqref{eq:grpo_objective} is averaged over the usable response tokens, using token-level aggregation \citep{Yu2025DAPO},
\begin{equation}
\begin{aligned}
\mathcal L_{\mathrm{GRPO}}&=\frac{1}{M}\sum_{i,t}\ell_{i,t},\\
\ell_{i,t}&=-\min\!\left(\rho_{i,t}\widehat A_i,
\widetilde\rho_{i,t}\widehat A_i\right)+\beta D_{i,t}.
\end{aligned}
\label{eq:grpo_objective}
\end{equation}
Here, $\widetilde\rho_{i,t}=\operatorname{clip}(\rho_{i,t},1-\epsilon,1+\epsilon)$. The fixed reference defines $\delta_{i,t}=\log\pi_{\mathrm{ref}}(p_{i,t}\mid q_{i,t})-\log\pi_\theta(p_{i,t}\mid q_{i,t})$ and $D_{i,t}=\exp(\delta_{i,t})-\delta_{i,t}-1$. We use $\epsilon=0.2$ and $\beta=0.02$. Clipping limits the contribution of large probability-ratio changes, while the KL term penalises departures from the SFT reference.

\begin{table}[htbp]
\small
\caption{Notation used in the group-relative policy update.}
\label{tab:policy_notation}
 {\centering
\begin{tabularx}{\linewidth}{@{}lX@{}}
\toprule
Symbol & Meaning \\
\midrule
$p_{i,t}$ & Token $t$ of candidate program $i$. \\
$q_{i,t}$ & Prompt and preceding tokens of that candidate. \\
$\pi_\theta$, $\pi_{\mathrm{old}}$ & Current and sampling policies, respectively. \\
$\pi_{\mathrm{ref}}$ & Fixed SFT reference policy. \\
$r_i$, $\widehat A_i$ & Candidate reward and group-centred advantage. \\
$\rho_{i,t}$, $\widetilde\rho_{i,t}$ & Unclipped and clipped probability ratios. \\
$\delta_{i,t}$ & Reference minus current log probability. \\
$D_{i,t}$ & Sampled KL regularisation term. \\
$M$ & Total number of usable response tokens. \\
$\epsilon$, $\beta$ & Clipping parameter and KL coefficient. \\
\bottomrule
\end{tabularx}
\par}
\end{table}

Only candidates with completed verification outcomes contribute to an update. If a solver run is interrupted by a licence or system timeout, the affected group is excluded so that an incomplete run is not treated as an engineering failure. Equal-reward groups have zero centred task advantage under Eq.~\eqref{eq:group_advantage}, while the reference-policy KL term remains defined. Once the usable groups have supplied an update, the current policy generates fresh programs for the next batch. Native execution and verification are repeated for these new candidates, maintaining feedback from the policy's current behaviour (Appendix~\ref{app:grpo_diagnostics} details the epoch-wise group reward diagnostics).

\section{Experimental setup}
\label{sec:experimental_setup}

Section 3 describes how the outcomes of OrcaFlex simulations guide the training of OceanCoder. We now assess the modelling capability acquired through training. We evaluate the models on a common set of held-out tasks that range from changes in individual parameters to complete engineering scenarios.

\subsection{Benchmark tasks and engineering scenarios}
\label{subsec:benchmark_tasks}

The benchmark comprises 665 tasks. Each task specifies the physical system to model, the analyses to perform and the engineering responses to report. All test instances are distinct from those used for supervised fine-tuning (SFT) or reinforcement learning. We organise the tasks into three strata to assess performance under parameter changes, structural modifications and engineering scenarios. Table~\ref{tab:test_tasks} summarises their composition.

\begin{table}[htbp]
\small
\caption{Composition of the 665-task evaluation set.}
\label{tab:test_tasks}
{\centering
\begin{tabularx}{\linewidth}{@{}>{\raggedright\arraybackslash}X r r@{}}
\toprule
Task stratum & Tasks & Share (\%) \\
\midrule
Core parameter variation & 352 & 52.93 \\
Structural variation & 237 & 35.64 \\
Engineering context & 76 & 11.43 \\
\midrule
Total & 665 & 100.00 \\
\bottomrule
\end{tabularx}
\par}
\end{table}

The 352 core parameter variations retain a familiar modelling procedure while changing its physical or operating parameters. They test whether a generated program correctly applies the specified values within that procedure. The 237 structural variations require changes to the model configuration. A task may prescribe different properties for a local line segment, change the rotational restraint at a line end or alter the number of supports. The program must implement the change in the relevant part of the model and extract the requested response at the specified location.

The categories and strata describe different aspects of the benchmark. Categories identify the engineering analysis family, while strata identify the change that a generated implementation must accommodate. This permits comparisons within a common task family as well as across different kinds of requirement changes. The category-by-stratum counts in Appendix~\ref{app:category_results} document this coverage. Evaluation tasks retain their own specifications, reference evidence and output requirements, so acceptance is determined for each complete task.

The remaining 76 tasks are based on documented engineering scenarios. We adapted 18 tasks from published offshore project descriptions and 58 from industry verification examples and technical documentation. Each scenario is expressed as a self-contained specification of the configuration, operating conditions, analysis sequence and required outputs. These tasks assess whether a model can combine the specified requirements in a complete OrcaFlex program. Table~\ref{tab:engineering_sources} summarises the source groups; Appendix~\ref{app:case_sources} provides the source catalogue and task mappings.

Constructing these tasks requires converting source descriptions into explicit numerical specifications. Representative load cases, coordinate frames, component properties and installation geometries are defined so that each problem can be implemented and assessed independently. Reference programs are reviewed against the resulting specifications, including units, connections, boundary conditions, analysis stages and response definitions. Native execution, model reloading and output recomputation provide the corresponding reference evidence. The source catalogue and task mappings retain the connection between the public engineering material and the self-contained benchmark problem.

\begin{table}[htbp]
\small
\caption{Provenance of the 76 engineering-context evaluation tasks. Public sources provide engineering scenario and modelling context for the self-contained benchmark specifications.}
\label{tab:engineering_sources}
{\centering
\begin{tabularx}{\linewidth}{@{}>{\raggedright\arraybackslash}X r >{\raggedright\arraybackslash}X >{\raggedright\arraybackslash}X@{}}
\toprule
Source class & Tasks & Examples of source context & Relation to the evaluated task \\
\midrule
Official examples, documentation, validation or user material & 58 & Moorings, production risers, deployment, contact and line feeding. & Scenario and modelling-procedure basis, formulated with explicit task parameters and required outputs. \\
Published offshore engineering projects & 18 & Fensfjorden, Browse trunkline, Veja Mate, Baltic Sea inter-array cables and Project Natick. & Self-contained simulation specifications adapted from published offshore project descriptions and operating conditions. \\
\midrule
Total & 76 & Source-card and case mappings in Appendix~\ref{app:case_sources}. & Native simulation references and task-specific acceptance checks. \\
\bottomrule
\end{tabularx}
\par}
\end{table}

\subsection{Compared models and training configurations}
\label{subsec:compared_models}

We compare OceanCoder with Qwen3-8B, Qwen3-235B-A22B-Instruct-2507 (hereafter Qwen3-235B) and ChatGPT-6 Sol. Qwen3-8B establishes performance at the starting point of OceanCoder training. Qwen3-235B provides a larger baseline from the same model family, and ChatGPT-6 Sol provides a closed-source comparison.

Three OceanCoder configurations allow us to examine how performance changes during training. OceanCoder-SFT is trained on the 12,000 verified task--program pairs described in Section~\ref{subsec:data_construction}. SFT runs for five epochs, comprising 3,750 updates. We select the checkpoint at update 2,250, corresponding to the end of the third epoch, for evaluation and subsequent reinforcement learning. Its comparison with the Qwen3-8B base model measures the change in task completion after domain supervision.

Both reinforcement learning configurations start from the selected SFT checkpoint. They use the same 150 training requirements, candidate sampling procedure and policy update settings. GRPO-Binary assigns a reward of one to a fully accepted program and zero to all other outcomes. Four-level GRPO uses the graded rewards defined in Section~\ref{subsec:grpo}. Comparing four-level GRPO with OceanCoder-SFT measures the additional change after reinforcement learning. Comparing it with GRPO-Binary tests the contribution of intermediate rewards. The training and evaluation settings are summarised in Table~\ref{tab:training_settings}.

\begin{table}[htbp]
\small
\caption{OceanCoder training and evaluation settings.}
\label{tab:training_settings}
 {\centering
\begin{tabularx}{\linewidth}{@{}l >{\raggedright\arraybackslash}X >{\raggedright\arraybackslash}X@{}}
\toprule
Setting & Stage I: SFT & Stage II: GRPO \\
\midrule
Initial model & Qwen3-8B & SFT (update 2,250) \\
Training data & 12,000 pairs (12 cats.) & 150 requirements \\
Schedule & 5 epochs (3,750 steps) & 3 epochs (57 steps) \\
Learning rate & $2\times 10^{-4}$ & $5\times 10^{-6}$ \\
Batching & 1/GPU, grad.\ accum.\ 16 & 8 prompts/batch \\
Sampling & -- & $G=4$, $T=1.0$, $p=0.95$ \\
Rewards & -- & $\{0,0.1,0.4,1\}$ \\
Checkpoint & Epoch 3 (step 2,250) & Epoch 3 (step 57) \\
Evaluation & 3 rounds $\times$ 665 tasks & Same as SFT \\
\bottomrule
\end{tabularx}
\par}
\par
 SFT uses 4-bit NF4 QLoRA ($r=16$, $\alpha=32$, dropout 0.05). GRPO-Binary uses identical settings with binary rewards $\{0,1\}$.
\end{table}

Each evaluated model is assessed across three evaluation rounds on the 665 benchmark tasks, with one generation attempt per task in each round. We report the mean and sample standard deviation of the evaluation metrics across the three rounds. Task-level records from the first evaluation round are used for stage-wise failure analysis and representative engineering case studies.

\subsection{Single-attempt generation and engineering assessment}
\label{subsec:evaluation_protocol}

Each model receives the task description and a common system instruction specifying the program and output requirements. Reference programs and expected numerical results are withheld during generation. Each model makes one attempt to generate a complete OrcaFlex program for each task. The evaluator assesses the resulting submission without simulator-feedback repair. The model receives no simulator feedback during generation and has no opportunity to revise its submission after execution.

Generation and assessment use fixed software and output requirements. The simulations use OrcaFlex and OrcFxAPI 11.6. Evaluation uses a decoding temperature of 0 and allows up to 16,384 completion tokens within the model's context window. Responses that reach the generation limit remain in the assigned-task denominator. Appendix~\ref{app:generation_review} records the decoding and termination rules used in the common protocol.

Assessment begins with checks on response format, Python syntax and static safety requirements. Programs that pass these checks are executed in OrcaFlex. After execution, the evaluator checks that the required outputs have been delivered and that the saved native files can be reloaded. It then applies the engineering criteria from Section~\ref{subsec:verification} to the saved model, analysis records and exported responses. A task is accepted only when all applicable checks pass, including the numerical tolerances specified for that task.

The assessment retains the stage at which a submission first fails. A format or code failure prevents a valid simulation from being established. A runtime failure interrupts the prescribed calculation. Delivery and reloading checks establish whether the required model and output files have been saved and can be inspected. Subsequent model, analysis and response checks determine complete engineering acceptance. These recorded outcomes support the cumulative counts in Appendix~\ref{app:acceptance_records} and the failure analysis in Section~\ref{sec:results_failures}. The same sequence is applied to every model and evaluation round.

\subsection{Evaluation metrics}
\label{subsec:evaluation_metrics}

The primary metric is Pass@1, the percentage of assigned tasks successfully completed in a single generation attempt under the engineering acceptance criteria in Section~\ref{subsec:evaluation_protocol}. The execution success rate (ESR) is the percentage for which the program runs to completion in OrcaFlex without a runtime error. Both metrics use all assigned tasks as the denominator, including submissions that reach the generation limit or fail a code check. The denominator is 665 for the overall comparison and the corresponding number of assigned tasks for each stratum.

Each metric is calculated separately for each evaluation round. The reported mean and sample standard deviation are then computed across the three rounds, with 665 tasks in the denominator of each overall rate.

Every program counted in Pass@1 also contributes to ESR. The difference between ESR and Pass@1, expressed in percentage points, gives the proportion of assigned tasks that execute successfully but fail a subsequent delivery, reload or engineering check. Reporting the two rates together distinguishes changes in the ability to generate runnable programs from changes in the ability to complete the specified engineering task.

\begin{table}[htbp]
\small
\caption{Overall and stratified performance on the 665-task benchmark. ESR denotes execution success rate, and Pass@1 measures complete task success. Values with standard deviations are reported as mean $\pm$ sample standard deviation across three evaluation rounds. Stratum sizes are given in parentheses.}
\label{tab:main_results}
\begingroup
 {\centering
\begin{tabularx}{\linewidth}{@{}>{\raggedright\arraybackslash}X r r@{}}
\toprule
Model & ESR (\%) & Pass@1 (\%) \\
\midrule
Qwen3-8B
& 0.00 & 0.00 \\
Qwen3-235B
& 0.00 & 0.00 \\
ChatGPT-6 Sol
& $27.52 \pm 1.52$ & $7.82 \pm 1.27$ \\
OceanCoder-SFT
& $69.17 \pm 1.40$ & $46.77 \pm 1.08$ \\
OceanCoder-GRPO-Binary
& $67.37 \pm 1.63$ & $46.47 \pm 1.43$ \\
OceanCoder-GRPO
& $\mathbf{70.53 \pm 1.08}$ & $\mathbf{51.43 \pm 0.94}$ \\
\midrule
Task stratum
& SFT Pass@1 (\%) & GRPO Pass@1 (\%) \\
\midrule
Core parameter variation (352)
& $80.78 \pm 1.34$ & $\mathbf{88.83 \pm 1.18}$ \\
Structural variation (237)
& $10.69 \pm 0.88$ & $\mathbf{11.25 \pm 0.64}$ \\
Engineering context (76)
& $1.75 \pm 0.76$ & $\mathbf{3.51 \pm 0.76}$ \\
\bottomrule
\end{tabularx}
\par}
\endgroup
\end{table}

\section{Results and analysis}
\label{sec:results}

The evaluation in Section~\ref{sec:experimental_setup} measures both successful execution and completion of the specified engineering task. We first compare overall and stratified model performance, then assess the reliability of engineering verification and the contribution of graded training feedback. Finally, failure patterns and representative engineering cases are examined to explain the observed capabilities and remaining limitations.

\subsection{Overall and stratified task performance}
\label{sec:results_overall}

OceanCoder achieves higher complete-task success rates than the evaluated base and closed-source models under the common single-attempt protocol. As shown in Table~\ref{tab:main_results}, neither Qwen base model completes any benchmark task, while ChatGPT-6 Sol achieves a Pass@1 of 7.82\%. Supervised fine-tuning on verified engineering examples increases Pass@1 to 46.77\%, and subsequent GRPO training raises it to 51.43\%. The improvement is concentrated in core parameter variations, where Pass@1 increases from 80.78\% to 88.83\%, a gain of 8.05 percentage points. By comparison, structural variations and engineering-context tasks reach only 11.25\% and 3.51\%, respectively, indicating that changes in structural configuration and broader engineering requirements remain substantially more difficult than parameter variations within familiar task families.

The distribution of gains helps identify the capability improved by engineering feedback. Figure~\ref{fig:stage_performance}(c) shows that 28 of the 31 additional accepted tasks in the first evaluation round occur in the core stratum, with category-level counts detailed in Appendix~\ref{app:category_results}. These tasks retain a familiar modelling procedure and require the program to apply different physical or operating values correctly. The concentration of gains in this stratum is consistent with refinement of previously learned modelling routines. Structural variations additionally require assigning changed properties, connections or restraints to the appropriate parts of the system. Their smaller improvement shows that applying a known procedure to new values and adapting that procedure to a changed configuration remain different generation requirements.

\begin{figure}[htbp]
\centering
\includegraphics[width=\linewidth]{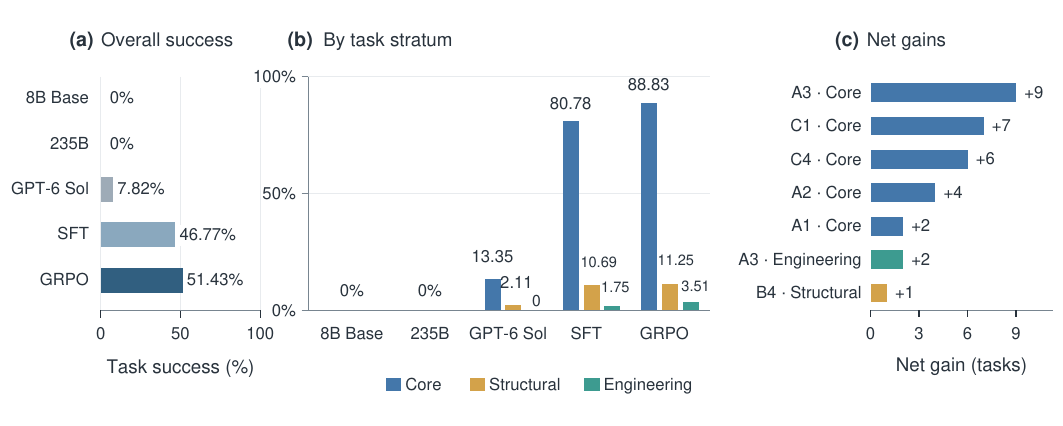}
\caption{Complete task success on the 665-task benchmark. (a) Overall Pass@1 for the compared models. (b) Pass@1 by task stratum. Panels (a) and (b) report means across three evaluation rounds. GPT-6 Sol denotes ChatGPT-6 Sol. (c) Category-level net increases in accepted tasks from SFT to GRPO in the first evaluation round. The gains sum to 31 tasks.}
\label{fig:stage_performance}
\end{figure}

Execution success does not necessarily translate into complete engineering acceptance. ChatGPT-6 Sol successfully executes 27.52\% of its programs, but only 7.82\% satisfy all task requirements. A similar distinction remains after domain adaptation. From SFT to GRPO, Pass@1 increases by 4.66 percentage points, whereas ESR increases by only 1.36 percentage points. Consequently, the gap between ESR and Pass@1 narrows from 22.40 to 19.10 percentage points. This indicates a reduction in the proportion of assigned tasks for which programs execute successfully but fail subsequent delivery, model reloading or engineering acceptance. The difference between these metrics motivates a closer examination of the verification criteria used to determine task completion.

\subsection{Reliability of engineering verification}
\label{sec:results_verifier_audit}

To examine whether engineering verification can distinguish executable programs from correct implementations, we manually audit 100 SFT-generated programs covering all 12 task categories. Independent inspection of task requirements, programs, execution records and numerical outputs identifies 48 valid and 52 invalid implementations. Table~\ref{tab:verification_audit} compares the automated decisions with these manual judgements. Execution checks alone achieve 78.00\% agreement, incorrectly accepting 22 programs that complete execution but fail the specified engineering task. Among the 70 executable programs, these failures account for 31.43\%. Sixteen involve incorrect model configurations, while six involve errors in response extraction. Adding native-model input checks identifies the configuration errors and increases agreement to 94.00\%; output verification detects the remaining six errors.

\begin{table}[htbp]
\small
\caption{Verification decisions compared with manual inspection of 100 programs (48 valid, 52 invalid). False acceptances (FA) and false rejections (FR) are reported as counts.}
\label{tab:verification_audit}
 {\centering
\begin{tabularx}{\linewidth}{@{}>{\raggedright\arraybackslash}X r r r@{}}
\toprule
Verification criterion & FA & FR & Agree. (\%) \\
\midrule
Execution checks only & 22 & 0 & 78.00 \\
Execution and model-input checks & 6 & 0 & 94.00 \\
Unnormalised raw output comparison & 0 & 13 & 87.00 \\
\textbf{Full three-layer verifier} & \textbf{0} & \textbf{0} & \textbf{100.00} \\
\bottomrule
\end{tabularx}
\par}
\end{table}

The identified errors illustrate why model-input and output checks are both necessary. In some lowering tasks, generated programs assign a line diameter of 55~mm and an axial stiffness of 120~MN instead of the prescribed 70~mm and 210~MN. Although the solver completes the analysis, the constructed physical model differs from the specification. Conversely, a restoring-force program passes the model-input checks but reports a directional force of $+607.23$~kN when the required projection is $-607.23$~kN. This error arises during response extraction rather than model construction. These cases demonstrate that successful execution and correct physical configuration do not individually guarantee complete task acceptance.

Verification must also avoid rejecting physically equivalent results. Unnormalised output comparison falsely rejects 13 valid implementations: ten differ only in non-physical metadata labels, two represent equivalent directions across the $0^\circ/360^\circ$ boundary, and one contains a peak timestamp affected by numerical noise in a nearly zero yaw response. Filtering non-physical metadata, normalising periodic angles and applying the prescribed numerical tolerances eliminate these false rejections. The full three-layer verifier consequently accepts all 48 valid implementations and rejects all 52 invalid ones, achieving 100.00\% agreement with manual inspection on the audited sample.

False acceptances would reward programs that fail the engineering specification, while false rejections would penalise valid implementations. The audit therefore supports using both model-input and output verification to construct task-based training feedback.

\subsection{Effect of graded verification feedback}
\label{sec:results_ablation}
\label{sec:results_training}

\begin{figure}[htbp]
\centering
\includegraphics[width=\linewidth]{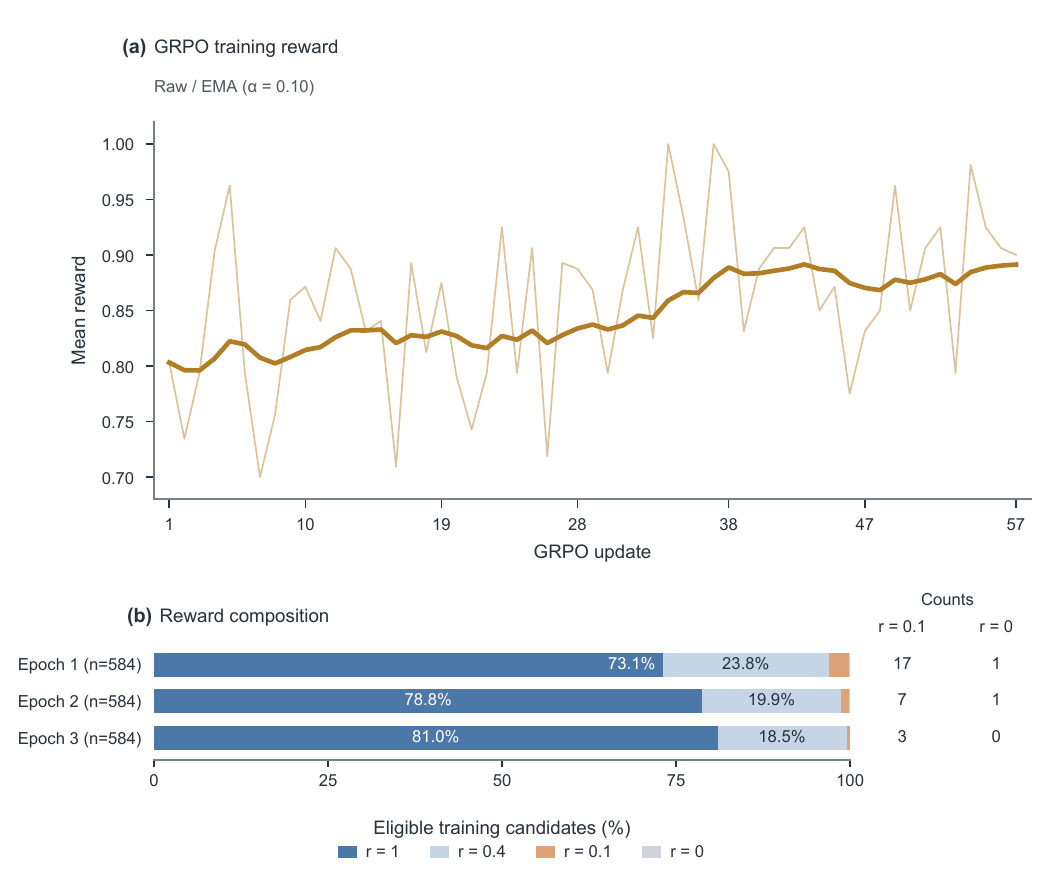}
\caption{Engineering verification outcomes during GRPO training. (a) Mean reward at each update and its exponential moving average with $\alpha=0.10$. (b) Reward composition across the three training epochs, with 584 eligible candidates per epoch. Full acceptance receives a reward of 1. Intermediate rewards distinguish execution or delivery failures from executable programs that fail engineering acceptance.}
\label{fig:training_dynamics}
\end{figure}

The reward ablation and training trajectory (Figure~\ref{fig:training_dynamics}) evaluate whether intermediate engineering verification outcomes improve policy learning beyond binary task acceptance. Under matched training settings, four-level GRPO achieves a mean Pass@1 of 51.43\%, compared with 46.47\% for binary rewards, an improvement of 4.96 percentage points. Binary-reward GRPO remains close to the SFT baseline of 46.77\%. The difference can be understood through group-relative advantage estimation. When all four candidates fail complete acceptance, binary rewards assign zero to every candidate, producing zero relative advantages regardless of their execution progress. Four-level rewards instead distinguish failures at different stages. In the training records, 142 of 438 usable groups (32.42\%) produce non-zero relative advantages under four-level rewards, compared with 118 groups (26.94\%) under binary rewards.

The frequency of intermediate outcomes further explains their contribution. Among 1,752 eligible training candidates, 363 pass execution, delivery and model reloading but fail engineering acceptance, receiving a reward of 0.4. Another 27 pass the code checks but fail execution, delivery or model reloading, receiving 0.1. Together, these 390 candidates account for 22.26\% of the sampled programs. Under a binary scheme, their progress would not be distinguished from earlier failures. The four-level hierarchy retains this information while reserving the maximum reward for complete engineering acceptance. The ablation results therefore support using verification states that reflect different stages of task completion.

The reward comparison also shows a difference in retained execution capability. Binary-reward GRPO records an ESR of 67.37\%, compared with 69.17\% for SFT and 70.53\% for four-level GRPO. Thus, the configuration that removes intermediate distinctions has lower execution success as well as lower complete-task success. The intermediate states describe progress that is observable before final acceptance, including delivery of reloadable models. Their ordering supplies relative preferences among incomplete candidates generated for the same requirement. Complete acceptance remains governed by the same engineering checks for every evaluated checkpoint. The improvement under graded rewards therefore accompanies more successful execution and a higher rate of satisfying the full specification.

The training dynamics are consistent with this interpretation. As shown in Figure~\ref{fig:training_dynamics}, mean reward increases from 0.8293 in the first epoch to 0.8844 in the third epoch. Over the same period, the full-acceptance rate rises from 73.12\% to 80.99\%, while intermediate outcomes decrease from 26.71\% to 19.01\%. These changes indicate that a larger proportion of programs sampled during training satisfy the complete engineering requirements. They describe training-set behaviour rather than independent generalisation; the separate benchmark results in Table~\ref{tab:main_results} establish the corresponding improvement on evaluation tasks.

\subsection{Failure modes and representative engineering cases}
\label{sec:results_failures}
\label{sec:results_cases}

\begin{figure}[htbp]
\centering
\includegraphics[width=\linewidth]{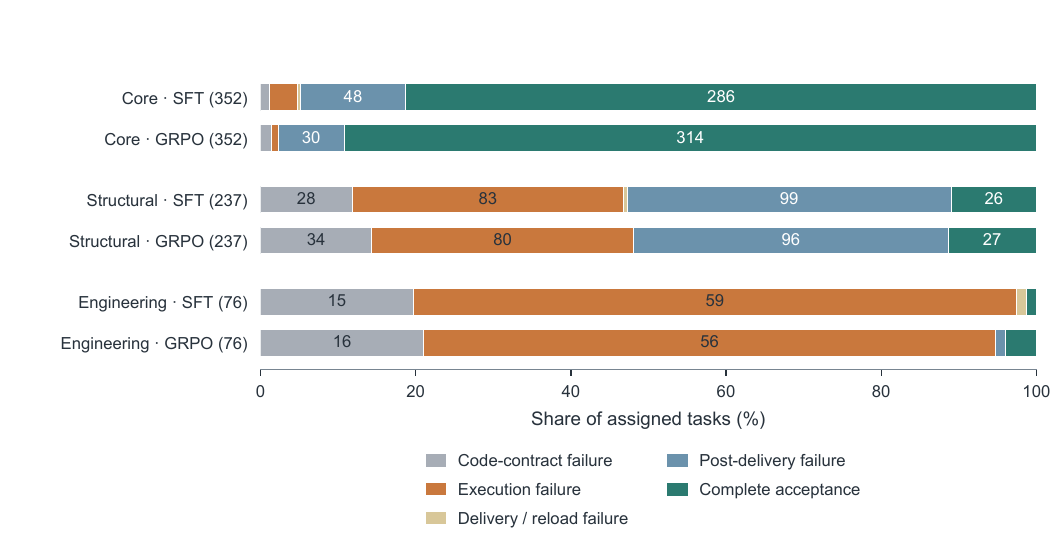}
\caption{Failure and acceptance outcomes for SFT and four-level GRPO across task strata in the first evaluation round. Each bar partitions the assigned tasks by the earliest failed stage or complete acceptance. The stages cover code-contract compliance, execution, delivery and model reloading, and subsequent engineering acceptance. Parentheses indicate stratum sizes.}
\label{fig:failure_reward_analysis}
\end{figure}

Despite the improvements in complete-task success, performance remains strongly dependent on task complexity. Figure~\ref{fig:failure_reward_analysis} shows the stage-wise outcomes from the first evaluation round. On structural variations, 123 GRPO programs execute successfully, but only 27 pass complete engineering acceptance. The remaining 96 executable programs fail subsequent acceptance checks, indicating persistent difficulties in preserving modified physical configurations or extracting the required responses. Engineering-context tasks exhibit a different failure pattern: only four programs execute successfully, three of which pass complete acceptance. Their principal difficulties therefore arise earlier, during program construction and execution. These patterns indicate that structural modifications and broader engineering scenarios challenge different stages of the modelling process.

These patterns give different meanings to failure in the two strata. A structural variant can retain most of a familiar program while changing a local segment, connection or restraint. The implementation must associate the modification with the correct object and preserve the requested response location. A solvable model can still contain an incorrect assignment. For engineering-context tasks, the low execution rate places the main observed difficulty earlier in the workflow. The model must assemble the required objects, API operations and analysis sequence into a complete executable program. The stage records consequently distinguish a failure to construct that workflow from a failure to implement its engineering details.

\begin{figure}[htbp]
\centering
\includegraphics[width=\linewidth]{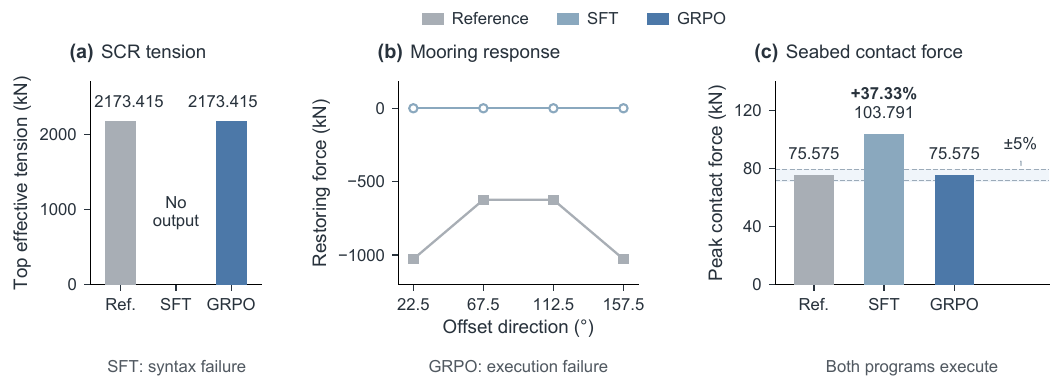}
\caption{Reference solutions and OceanCoder outputs for three representative engineering tasks. (a) Top effective tension in the steel catenary riser task. (b) Horizontal restoring forces at four vessel offset directions in the mooring task. (c) Peak upward seabed contact force in the two-point lowering task introduced in Figure~\ref{fig:motivation}. Syntax and execution failures are marked explicitly. The shaded band in panel (c) indicates the $\pm5\%$ task tolerance around the reference force.}
\label{fig:engineering_cases}
\end{figure}

Three representative cases from the first evaluation round illustrate these differences in greater detail (Appendix~\ref{app:case_evidence} provides the detailed case specifications and numerical comparisons). The static steel catenary riser task requires modelling a riser in 1,200~m water depth with a length of 2,220~m. The SFT-generated program contains a malformed statement and fails syntax validation. In contrast, GRPO constructs the riser--seabed system, completes the static equilibrium analysis and reports a top effective tension of 2,173.415~kN, matching the reference value to three decimal places (Figure~\ref{fig:engineering_cases}(a)). This case demonstrates an improvement in program construction and execution, where the SFT model fails before a valid analysis can begin.

Structural modifications present a different challenge. In the four-point mooring task, a short terminal line segment with reduced axial stiffness is added, and horizontal restoring forces are required at four vessel offsets of 12~m. The SFT program executes but incorrectly assigns the multi-segment line collection, producing restoring forces close to zero rather than the reference range of $-1{,}026.96$ to $-622.85$~kN. The GRPO program attempts to configure the required segment arrays but encounters a Python tuple-unpacking error during execution. As shown in Figure~\ref{fig:engineering_cases}(b), neither program completes the task, although their failures occur at different stages. This case illustrates the remaining difficulty of generating programs for modified structural configurations, consistent with the low acceptance rate observed in the structural-variation stratum.

The two-point lowering case further demonstrates why complete engineering verification is necessary even when programs execute successfully. A 65~t subsea payload is lowered onto an elastic seabed using two winch-controlled lines. The prescribed line diameter, unit mass and axial stiffness are 70~mm, 18~kg/m and 210~MN, respectively. Both SFT and GRPO programs complete execution and export the required models and results. However, the SFT program assigns 55~mm, 11~kg/m and 120~MN, altering the physical system and consequently its contact response. It predicts first seabed contact at 35.0825~s rather than the reference 36.0126~s, an error of 0.9301~s that exceeds the 0.010~s tolerance. Its peak upward contact force is 103.791~kN, compared with the reference 75.575~kN, corresponding to a 37.33\% overprediction against a 5\% tolerance, and it overpredicts maximum foot penetration (0.09041~m compared with the reference 0.05918~m). The GRPO program preserves the prescribed line properties and lowering schedule, reproducing the reference contact time, peak force and penetration (Figure~\ref{fig:engineering_cases}(c)). This example illustrates how engineering verification distinguishes executable but physically incorrect programs from those that satisfy the complete task specification.

\section{Conclusions}
\label{sec:conclusions}

We have developed OceanCoder to generate complete OrcaFlex programs from natural-language ocean engineering requirements. Engineering-constrained data construction and supervised fine-tuning establish program generation for model construction, analysis execution and result extraction. A three-layer verifier then assesses the constructed model, completed analyses and reported responses. Its outcomes provide graded feedback for online GRPO and common acceptance criteria for the 665-task benchmark.

Across the benchmark, Pass@1 increases from 46.77\% after SFT to 51.43\% after four-level GRPO, compared with 46.47\% under binary rewards. The full verifier agrees with manual inspection on all 100 audited programs, including cases in which execution succeeds despite engineering errors. These results support the complementary roles of reliable domain supervision and engineering verification feedback. Most of the performance gain occurs in core parameter variations, while structural modifications and broader engineering scenarios remain difficult. Complete task acceptance therefore requires more than successful program execution.

The findings are limited to the task categories, parameter ranges and modelling assumptions represented in the corpus and benchmark. Engineering acceptance establishes consistency with the specified numerical problem and its required outputs; it does not by itself validate predictions against unobserved field conditions. Future work should address structural and engineering-scenario failures by expanding the configurations and analysis procedures represented in training, together with their verification criteria. Such extensions would broaden the modelling scope while retaining explicit links between requirements, generated programs and accepted results.

\section*{Data availability}

Representative seed specifications, source mappings, summaries of the 100-task verification audit and evaluation statistics are provided in the manuscript and appendices. The trained OceanCoder weights and the specifications of the 665-task benchmark will be released publicly upon publication.

\par
\appendix

\numberwithin{equation}{section}
\numberwithin{figure}{section}
\numberwithin{table}{section}

\renewcommand{\theHequation}{appendix.\arabic{section}.\arabic{equation}}
\renewcommand{\theHfigure}{appendix.\arabic{section}.\arabic{figure}}
\renewcommand{\theHtable}{appendix.\arabic{section}.\arabic{table}}

\section{Data construction and training records}
\label{app:training_records}

\begin{figure}[htbp]
\centering
\includegraphics[width=\linewidth]{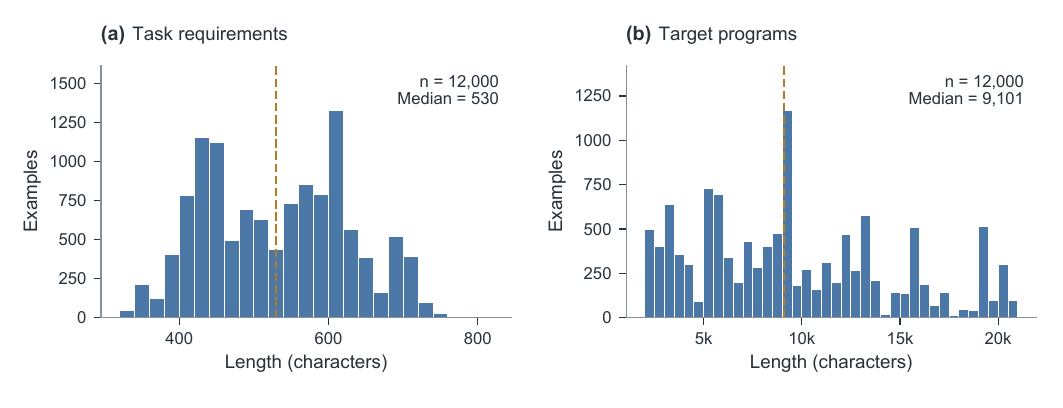}
\caption{Character-length distributions of the 12,000 SFT task requirements and target Python programs.}
\label{fig:data_lengths}
\end{figure}

\subsection{Corpus composition and example lengths}
\label{app:data_lengths}

The supervised corpus contains 12,000 instruction--program pairs spanning 12 task categories, with ten seed families per category and 100 parameter-expanded instances per seed. Each pair is generated from a distinct parameter set within its seed family.

Table~\ref{tab:app_data_lengths} and Figure~\ref{fig:data_lengths} report the character-length distributions of the task descriptions and target programs across the corpus.

\begin{table}[htbp]
\small
\caption{Corpus records and character lengths (mean; median; range). GRPO requirements contain no supervised response.}
\label{tab:app_data_lengths}
 {\centering
\begin{tabularx}{\linewidth}{@{}l r >{\raggedright\arraybackslash}X@{}}
\toprule
Resource & Records & Character lengths (mean; median; range) \\
\midrule
SFT requirements & 12,000 & 530.95; 530; 321--804 \\
SFT programs & 12,000 & 9,608.77; 9,101; 2,420--20,976 \\
GRPO requirements & 150 & 512.90; 502; 395--717 \\
\bottomrule
\end{tabularx}
\par}
\end{table}

The SFT responses contain 67--462 lines, with a mean of 206.57 and a median of 221. The 150 GRPO requirements are sampled from the supervised training sources for online policy exploration, focusing on multi-component mooring and staged offshore-operation tasks (A2: 35; A3: 2; A4: 2; B1: 6; B2: 4; B4: 1; C1: 32; C2: 32; and C4: 36) where multi-step configuration and response-extraction errors are most frequent after SFT.

\subsection{Category-specific admission verification}
\label{app:admission_verification}

Each of the 12 categories in the supervised corpus contributes 1,000 verified task--program pairs. Candidate pairs undergo category-specific verification tailored to their physical modelling and numerical analysis requirements, combining static AST checks, native OrcaFlex solver execution, saved-model reload and recomputation, and output verification. At the seed-family level, manual review of model geometry and response curves is further combined with mesh and time-step sensitivity checks to establish baseline discretisation settings. Table~\ref{tab:app_admission} summarises the primary verification checks across the 12 categories.

\begin{table}[htbp]
\small
\caption{Admission verification checks across the 12 SFT categories (1,000 pairs per category).}
\label{tab:app_admission}
 {\centering
\begin{tabularx}{\linewidth}{@{}l >{\raggedright\arraybackslash}X@{}}
\toprule
Cat. & Primary physical and numerical verification checks \\
\midrule
A1 & Single-catenary static equilibrium, line-property and boundary coordinates, end tensions, minimum elevation and maximum curvature. \\
A2 & Multi-line mooring equilibrium, fairlead/anchor layout, segment properties and directional offset restoring forces. \\
A3 & Deepwater SCR static profile, hang-off and touchdown zone (TDZ) arc lengths, bending moments and effective tensions. \\
A4 & Pipeline lateral stability across stable and breakout regimes, hydrodynamic/coating parameters and lateral displacement. \\
B1 & Regular-wave catenary dynamics, build-up/steady-cycle window alignment, peak tension and dynamic excursion. \\
B2 & Coupled four-point mooring dynamics under waves and current, floater trajectory and fairlead tension envelopes. \\
B3 & Dynamic SCR response under top motion, waves and current, TDZ curvature, minimum tension and stress ranges. \\
B4 & Free-span pipeline wave--current response, span geometry, end restraints, mid-span oscillation and bending moments. \\
C1 & Static S-lay installation across stinger supports, overbend/sagbend curvature, tensioner load and touchdown reaction. \\
C2 & Dynamic cable-laying under vessel transit and payout, catenary evolution, seabed contact and bottom tension bounds. \\
C3 & Single-point lifting and splash-zone lowering, winch payout kinematics, hydrodynamic loading and hoist-line tensions. \\
C4 & Two-point subsea lowering under implicit dynamics, synchronous payout, first-contact timing, peak force and penetration. \\
\bottomrule
\end{tabularx}
\par}
\end{table}

\subsection{Executed training settings and resources}
\label{app:executed_hyperparameters}

The principal hyperparameter settings for both stages are listed in Table~\ref{tab:training_settings}. During Stage I, the five-epoch SFT run (3,750 updates) required 44.92 h (161,724.67 s) in total, with a cumulative run-average training loss of 0.00807 across all updates. At the selected epoch-3 checkpoint (update 2,250, reached at 26.96 h; Figure~\ref{fig:sft_loss}), the logged step-level response-token cross-entropy loss had decreased from 0.8971 (at update 10) to 0.0022. Stage II (online GRPO) was executed for three epochs (57 updates; 19 updates per epoch) on two NVIDIA A800 (80 GB PCIe) GPUs running CUDA 12.4 and PyTorch 2.6.0.

\begin{figure}[htbp]
\centering
\includegraphics[width=\linewidth]{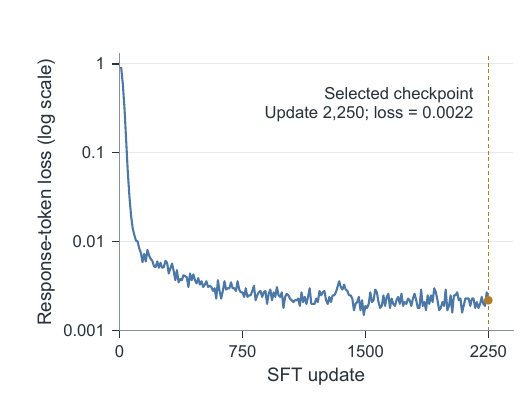}
\caption{SFT response-token loss trajectory through the selected epoch-3 checkpoint (update 2,250). All 225 logged loss values from updates 10 to 2,250 are shown on a logarithmic scale; the dashed line marks update 2,250, where the loss stabilises at 0.0022.}
\label{fig:sft_loss}
\end{figure}

\subsection{GRPO rewards, usable groups and partial progress}
\label{app:grpo_diagnostics}

Across three epochs, the training loop consumes 450 four-candidate groups and samples 1,800 nominal candidates. After the recorded verification exclusions, 438 groups and 1,752 candidates ($n=584$ per epoch) remain eligible for reward statistics and updates. Table~\ref{tab:app_epoch_rewards} retains the exact four-level counts. The mean training reward rises from 0.8293 to 0.8844, and the full-reward proportion rises from 73.12\% to 80.99\%. These quantities describe the reused training requirements and are distinct from success rates on evaluation tasks.

\begin{figure}[htbp]
\centering
\includegraphics[width=\linewidth]{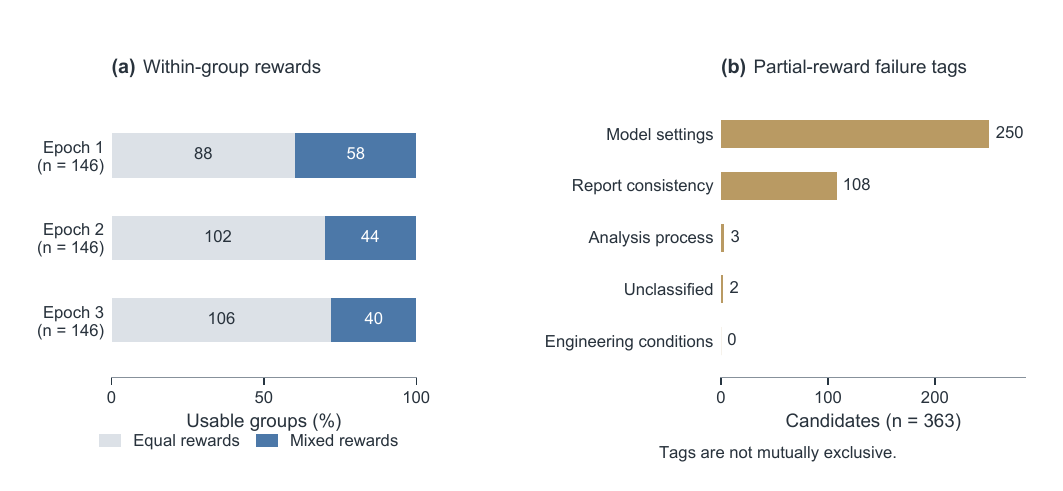}
\caption{Additional GRPO training diagnostics. The left panel separates equal-reward and mixed-reward usable groups by epoch; equal-reward groups have zero centred task advantage. The right panel shows recorded failure labels for the 363 reward-0.4 candidates.}
\label{fig:reward_groups}
\end{figure}

\begin{table}[htbp]
\small
\caption{GRPO candidate reward counts by training epoch ($n=584$ eligible per epoch; 1,752 total). Partial denotes $r\in\{0.1,0.4\}$.}
\label{tab:app_epoch_rewards}
 {\centering
\begin{tabularx}{\linewidth}{@{}l r r r r >{\raggedleft\arraybackslash}X r r@{}}
\toprule
Epoch & $r=0$ & $0.1$ & $0.4$ & $1$ & Mean & Full (\%) & Part. (\%) \\
\midrule
1 & 1 & 17 & 139 & 427 & 0.8293 & 73.12 & 26.71 \\
2 & 1 & 7 & 116 & 460 & 0.8683 & 78.77 & 21.06 \\
3 & 0 & 3 & 108 & 473 & 0.8844 & 80.99 & 19.01 \\
\midrule
Total & 2 & 27 & 363 & 1360 & 0.8607 & 77.63 & 22.26 \\
\bottomrule
\end{tabularx}
\par}
\end{table}

Four groups are excluded in each of the three epochs due to solver-licence or timeout interruptions. The 48 affected candidates are omitted from gradient updates rather than assigned zero rewards.

Among usable groups, 88 of 146 have identical candidate rewards in epoch 1, 102 of 146 in epoch 2, and 106 of 146 in epoch 3, corresponding to 58, 44 and 40 mixed-reward groups, respectively (Figure~\ref{fig:reward_groups}a). Under the group-centred advantage formulation in Eq.~\eqref{eq:group_advantage}, equal-reward groups yield zero relative task advantage while still contributing through reference-policy KL regularisation.

Table~\ref{tab:app_partial_failures} and Figure~\ref{fig:reward_groups}(b) summarise the verification gates failed by the 363 candidates receiving $r=0.4$. Because a single candidate can trigger multiple verification flags, these categories represent diagnostic failure tags rather than mutually exclusive root causes.

\begin{table}[htbp]
\small
\caption{Verification failure tags for the 363 GRPO training candidates with reward 0.4. Percentages use 363 as the denominator.}
\label{tab:app_partial_failures}
\begingroup
 {\centering
\begin{tabularx}{\linewidth}{@{}>{\raggedright\arraybackslash}X r r@{}}
\toprule
Recorded label & Candidates & \% \\
\midrule
Model settings & 250 & 68.87 \\
Analysis process & 3 & 0.83 \\
Engineering conditions & 0 & 0.00 \\
Result/report consistency & 108 & 29.75 \\
Unclassified numerical difference & 2 & 0.55 \\
\bottomrule
\end{tabularx}
\par}
\endgroup
\end{table}

Across epochs 1 to 3, physical model-setting violations decline steadily from 100 cases in epoch 1 to 80 in epoch 2 and 70 in epoch 3, indicating fewer violations of the specified engineering constraints on the sampled training candidates during reinforcement learning.

The training logs record reward summaries across all 57 updates on both GPU ranks. Figure~\ref{fig:training_dynamics}(a) plots their per-update means and exponential moving average with $\alpha=0.10$.

\section{Category-level evaluation outcomes}
\label{app:benchmark_provenance}
\label{app:category_results}

\begin{table}[htbp]
\small
\caption{Complete task acceptance by category and stratum in the first evaluation round (SFT vs.\ GRPO accepted counts).}
\label{tab:appendix_category_counts}
 {\centering
\begin{tabular}{@{}l r r r r r r r r r@{}}
\toprule
 & \multicolumn{3}{c}{Core} & \multicolumn{3}{c}{Structural} & \multicolumn{3}{c}{Engineering} \\
\cmidrule(lr){2-4}\cmidrule(lr){5-7}\cmidrule(l){8-10}
Cat. & $n$ & SFT & GRPO & $n$ & SFT & GRPO & $n$ & SFT & GRPO \\
\midrule
A1 & 28 & 25 & 27 & 20 & 0 & 0 & 10 & 0 & 0 \\
A2 & 30 & 10 & 14 & 20 & 0 & 0 & 11 & 0 & 0 \\
A3 & 30 & 18 & 27 & 20 & 0 & 0 & 10 & 1 & 3 \\
A4 & 30 & 27 & 27 & 17 & 9 & 9 & -- & -- & -- \\
B1 & 30 & 26 & 26 & 20 & 0 & 0 & 8 & 0 & 0 \\
B2 & 29 & 28 & 28 & 20 & 0 & 0 & 2 & 0 & 0 \\
B3 & 30 & 29 & 29 & 20 & 0 & 0 & 8 & 0 & 0 \\
B4 & 30 & 30 & 30 & 20 & 5 & 6 & 8 & 0 & 0 \\
C1 & 28 & 12 & 19 & 20 & 12 & 12 & 9 & 0 & 0 \\
C2 & 30 & 30 & 30 & 20 & 0 & 0 & 9 & 0 & 0 \\
C3 & 27 & 27 & 27 & 20 & 0 & 0 & 1 & 0 & 0 \\
C4 & 30 & 24 & 30 & 20 & 0 & 0 & -- & -- & -- \\
\midrule
Total & 352 & 286 & 314 & 237 & 26 & 27 & 76 & 1 & 3 \\
\bottomrule
\end{tabular}
\par}
\end{table}

Table~\ref{tab:appendix_category_counts} reports complete task acceptance counts across all category--stratum combinations on the 665-task benchmark for the first evaluation round; see Section~\ref{subsec:benchmark_tasks} for benchmark composition. A dash indicates that the benchmark contains no task in that cell. Comparing task-level outcomes between the SFT and four-level GRPO checkpoints in the first evaluation round, the net gain of 31 accepted tasks comprises 28 core parameter-variation tasks, one structural-variation task and two engineering-context tasks.

\section{Stage-wise verification outcomes}
\label{app:acceptance_records}

Table~\ref{tab:appendix_acceptance_stages} reports cumulative pass counts across the five sequential evaluation gates analysed in Section~\ref{sec:results_failures}: syntactic validity, strict code-contract compliance, native solver execution, model delivery/reloading, and complete engineering acceptance. Unlike the mutually exclusive failure breakdown in Figure~\ref{fig:failure_reward_analysis}, each column in Table~\ref{tab:appendix_acceptance_stages} records the total number of tasks passing up to that verification gate out of the full stratum denominator. For the base Qwen3-8B and Qwen3-235B models, cumulative pass counts reach 0/665 (0.00\%) at the solver execution stage and all subsequent verification tiers (delivery and complete engineering acceptance), as invalid OrcFxAPI object or data-item references prevent the generated scripts from executing under the single-attempt evaluation protocol.

\begin{table}[htbp]
\small
\caption{Cumulative stage pass counts in the first evaluation round across syntax (Syn.), strict code (Code), execution (Exec.), delivery/reload (Del.) and complete acceptance (Pass).}
\label{tab:appendix_acceptance_stages}
 {\centering
\begin{tabularx}{\linewidth}{@{}X l r r r r r r@{}}
\toprule
Stratum & Policy & $n$ & Syn. & Code & Exec. & Del. & Pass \\
\midrule
All & SFT & 665 & 618 & 618 & 463 & 460 & 313 \\
 & GRPO & 665 & 622 & 610 & 471 & 471 & 344 \\
Core & SFT & 352 & 348 & 348 & 335 & 334 & 286 \\
 & GRPO & 352 & 347 & 347 & 344 & 344 & 314 \\
Structural & SFT & 237 & 209 & 209 & 126 & 125 & 26 \\
 & GRPO & 237 & 211 & 203 & 123 & 123 & 27 \\
Engineering & SFT & 76 & 61 & 61 & 2 & 1 & 1 \\
 & GRPO & 76 & 64 & 60 & 4 & 4 & 3 \\
\bottomrule
\end{tabularx}
\par}
\end{table}

\section{Generation and verification protocol}
\label{app:generation_review}

All evaluated models use a decoding temperature of 0, with a maximum completion length of 16,384 tokens. Within each evaluation round, each model generates one program per task without simulator-feedback repair or access to reference solutions. Responses that exceed the token limit remain in the evaluation denominator and are recorded as code-check failures. The same model, analysis and output checks are applied to all evaluated models.

For the 100-task verification reliability audit (Section~\ref{sec:results_verifier_audit}, Table~\ref{tab:verification_audit}), tasks were selected prior to score inspection via proportional stratified sampling over the 665-task benchmark, producing 53 core parameter-variation tasks (four to five per category across A1--C4), 36 structural-variation tasks (three per category across A1--C4) and 11 engineering-context tasks (spanning A1, A2, A3, B1, B3, B4, C1 and C2). Across these three strata, solver execution succeeds on 51/53 (96.23\%), 19/36 (52.78\%) and 0/11 (0.00\%) tasks (70/100 overall), whereas complete engineering verification accepts 41/53 (77.36\%), 7/36 (19.44\%) and 0/11 (0.00\%) tasks (48/100 overall), matching the manual code-and-output audit on every stratum.

\section{Engineering sources, seeds and case evidence}
\label{app:case_sources}

\begin{table}[htbp]
\small
\caption{Archived source catalogue for the 76 engineering-context tasks. Counts indicate the number of tasks citing each source; one task may cite multiple sources.}
\label{tab:source_catalogue_1}
\label{tab:source_catalogue_2}
 {\centering
\begin{tabularx}{\linewidth}{@{}l >{\raggedright\arraybackslash}p{0.28\linewidth} r >{\raggedright\arraybackslash}X@{}}
\toprule
ID & Source title & Tasks & Public URL \\
\midrule
G03 & Line contact -- modelling & 1 & \url{https://www.orcina.com/webhelp/OrcaFlex/Content/html/Linecontact\%2CModelling.htm} \\
G04 & Modelling line feeding & 1 & \url{https://www.orcina.com/webhelp/OrcaFlex/Content/html/Modellinglinefeeding.htm} \\
G11 & C06 CALM buoy & 2 & \url{https://www.orcina.com/wp-content/uploads/examples/c/c06/C06\%20CALM\%20buoy.pdf} \\
G15 & Production riser examples & 1 & \url{https://www.orcina.com/Resources/examples/} \\
G16 & Mooring examples & 2 & \url{https://www.orcina.com/resources/examples/?key=c} \\
G26 & Line results -- contact and clearance & 2 & \url{https://www.orcina.com/webhelp/OrcaFlex/Content/html/Lineresults\%2CContact.htm} \\
G30 & OrcFxAPI and Python documentation & 2 & \url{https://www.orcina.com/resources/documentation/} \\
S01 & OrcaFlex examples -- Production risers & 24 & \url{https://www.orcina.com/Resources/examples/} \\
S03 & OrcaFlex examples -- Moorings & 23 & \url{https://www.orcina.com/resources/examples/?key=c} \\
S05 & OrcaFlex examples -- Pipelay and recovery & 9 & \url{https://www.orcina.com/resources/examples/?key=e} \\
S07 & OrcaFlex examples -- Deployment & 2 & \url{https://www.orcina.com/resources/examples/?key=g} \\
S13 & OrcaFlex User Group Meeting material & 9 & \url{https://www.orcina.com/Resources/ugm-material/} \\
S14 & OrcaFlex validation documents & 4 & \url{https://www.orcina.com/resources/validation/} \\
S15 & OrcaFlex QA, Testing and Validation & 3 & \url{https://www.orcina.com/wp-content/uploads/resources/validation/99-005-OrcaFlex-QA-Testing-and-Validation.pdf} \\
S17 & Browse Trunkline Installation Analysis & 5 & \url{https://atteris.com/resources/case-studies/concept-engineering-browse-pipelay/} \\
S18 & Veja Mate OWF cable installation study & 8 & \url{https://longitude-engineering.com/case-studies/veja-mate-owf-siem-aimery-cable-installation-study/} \\
S23 & Baltic Sea OWF inter-array cable study & 8 & \url{https://apollo.engineer/case-studies/installation-engineering-for-inter-array-cables-of-a-baltic-sea-offshore-wind-farm/} \\
S24 & The Fensfjorden Project & 5 & \url{https://bein.no/en/the-fensfjorden-project-nexans/} \\
S25 & Project Natick deployment and cable hook-up & 1 & \url{https://greenmarineuk.com/engineering-services/deployment-procedures-and-cable-hook-up/} \\
\bottomrule
\end{tabularx}
\par}
\end{table}

\subsection{Scope and source provenance}

The 76 engineering-context tasks comprise 18 tasks adapted from published offshore engineering project descriptions and 58 inspired by official examples, modelling documentation, validation reports and user-group materials. These two classes are mutually exclusive at the task level and span static moorings and risers, dynamic wave--current responses, free-span pipelines, S-lay and cable-laying operations, and subsea lowering scenarios.

To form self-contained benchmark problems suitable for automated verification, the task specifications standardise coordinate frames, representative load cases, structural properties and installation geometries so that each case defines a complete numerical problem. Table~\ref{tab:source_catalogue_1} lists the archived sources, and Table~\ref{tab:engineering_case_source_mapping} maps each task to its sources.

\begin{table}[htbp]
\small
\caption{Case-to-source mapping for the 76 engineering-context tasks (Table~\ref{tab:source_catalogue_1}).}
\label{tab:engineering_case_source_mapping}
 {\centering
\begin{tabularx}{\linewidth}{@{}
l >{\raggedright\arraybackslash}X
l >{\raggedright\arraybackslash}X
l >{\raggedright\arraybackslash}X@{}}
\toprule
Case ID & Source IDs & Case ID & Source IDs & Case ID & Source IDs \\
\midrule
RE2-001 & S03, S07 & RE2-027 & G11 & RE2-053 & S13, S24 \\
RE2-002 & S03 & RE2-028 & G11, G30 & RE2-054 & S13 \\
RE2-003 & S03 & RE2-029 & G15 & RE2-055 & S13 \\
RE2-004 & S03 & RE2-030 & G03, G26 & RE2-056 & S13 \\
RE2-005 & S03, S07 & RE2-031 & G16 & RE2-057 & S13 \\
RE2-006 & S03 & RE2-032 & G04 & RE2-058 & S13, S24 \\
RE2-007 & S03 & RE2-033 & S01, S03 & RE2-059 & S05, S17 \\
RE2-008 & S03 & RE2-034 & S01, S03 & RE2-060 & S05 \\
RE2-009 & S03 & RE2-035 & S01 & RE2-061 & S05, S17 \\
RE2-010 & S03 & RE2-036 & S01, S03 & RE2-062 & S05 \\
RE2-011 & S03 & RE2-037 & S01, S03 & RE2-063 & S05 \\
RE2-012 & S03 & RE2-038 & S01 & RE2-064 & S05, S17 \\
RE2-013 & S03 & RE2-039 & S01, S13 & RE2-065 & S05 \\
RE2-014 & S03 & RE2-040 & S01, S03 & RE2-066 & S05, S17 \\
RE2-015 & S03 & RE2-041 & S03 & RE2-067 & S05, S17 \\
RE2-016 & S03 & RE2-042 & S03 & RE2-068 & S18, S23 \\
RE2-017 & S01, S15 & RE2-043 & S01, S14 & RE2-069 & S18, S23 \\
RE2-018 & S01 & RE2-044 & S01 & RE2-070 & S18, S23 \\
RE2-019 & S01, S14 & RE2-045 & S01 & RE2-071 & S18, S24 \\
RE2-020 & S01 & RE2-046 & S01 & RE2-072 & S18, S23 \\
RE2-021 & S01 & RE2-047 & S01, S14 & RE2-073 & S18, S23 \\
RE2-022 & S01, S15 & RE2-048 & S01 & RE2-074 & S23, S25 \\
RE2-023 & S01 & RE2-049 & S01, S14 & RE2-075 & S18, S23 \\
RE2-024 & S01, S15 & RE2-050 & S01 & RE2-076 & S18, S23 \\
RE2-025 & G26 & RE2-051 & S13, S24 & & \\
RE2-026 & G16, G30 & RE2-052 & S13, S24 & & \\
\bottomrule
\end{tabularx}
\par}
\end{table}

\subsection{Archived seed examples and review coverage}
\label{app:seed_examples}

Table~\ref{tab:archived_seed_specifications} summarises three representative seed specifications: A1 single-catenary statics, B1 regular-wave catenary dynamics and C4 two-point subsea lowering. Each seed family undergoes manual modelling review, native solver execution, saved-model reload and independent output recomputation. For A1 \texttt{seed\_01} (73 nodes), static equilibrium yields effective tensions spanning 40.43--103.14~kN along the catenary profile.

To establish baseline discretisation settings at the seed-family level, mesh-refinement checks on B1 confirm a maximum tension variation of 0.035\% and a maximum displacement variation of 1.98\%, while time-step sensitivity tests on C4 across $\Delta t \in \{0.005, 0.0025, 0.00125\}$~s record a maximum first-contact timing variation of 0.00548~s. These discretisation baselines inform the numerical acceptance thresholds ($\pm 5\%$ for dynamic force and displacement quantities and $\pm 0.010$~s for contact timing) applied to expanded instances.

\begin{table}[htbp]
\small
\caption{Key specifications of three representative seed examples (A1, B1 and C4).}
\label{tab:archived_seed_specifications}
 {\centering
\begin{tabularx}{\linewidth}{@{}l >{\raggedright\arraybackslash}X@{}}
\toprule
Seed & Physical specification and requested analysis \\
\midrule
A1 \texttt{seed\_01} & Water depth 80~m; ends $(0,0,-8)$ and $(110,0,-80)$~m; length 145~m; $d=90$~mm; $m=95$~kg/m; $EA=800$~MN; 2~m segments. Still-water statics: end tensions, minimum elevation and maximum curvature with arc lengths. \\
B1 \texttt{seed\_01} & Water depth 80~m; 145~m chain ($d=80$~mm); regular wave ($H=2$~m, $T=8$~s, $0^\circ$); friction 0.50. Dynamic simulation (1 build-up + 15 steady periods): peak top tension and maximum horizontal excursion. \\
C4 \texttt{seed\_02} & Water depth 80~m; $6\times4\times3$~m payload (65~t, 24~m$^3$); two slings (2.5~m spacing); 3~s hover + 2~m payout at 0.06~m/s ($\Delta t=0.005$~s); seabed stiffness 1,000/300~kN/m$^3$. Outputs: first-contact time, peak force, penetration and line tensions. \\
\bottomrule
\end{tabularx}
\par}
\end{table}

For the explicit A1 expansion example, record \texttt{A1\_seed\_01\_v001} changes the water depth to 100~m, the upper end to $(0,0,-12)$~m, the anchor horizontal distance to 145~m and the line length to 200~m. The outer diameter, mass and axial stiffness become 160.7~mm, 158.2~kg/m and 729.7~MN; the outer contact diameter is 306~mm, and compression is limited. The 2~m target segment length and 1,000~kN/m$^3$ seabed normal stiffness are retained. The following lines are an excerpt of the archived expanded response; the surrounding object creation, connections and result extraction remain in the complete record.
\begin{verbatim}
model.environment.WaterDepth = 100.0
line_type_1.OD = 0.1607
line_type_1.MassPerUnitLength = 0.1582
line_type_1.EA = 729700.0
line.Length = (200.0,)
model.CalculateStatics()
\end{verbatim}

\subsection{Representative-case specifications and numerical evidence}
\label{app:case_evidence}

The case records retain the task specification, generated Python script, execution status, verification outcomes and saved simulation files. Reference values are obtained from verified OrcaFlex simulations. Table~\ref{tab:case_response_evidence} summarises the numerical comparisons for the three representative cases analysed in Section~\ref{sec:results_cases}.

\begin{table}[htbp]
\small
\caption{Recorded responses for the three representative cases (--- denotes failure; C4-039 denotes C4-FC1000-seed\_02-039).}
\label{tab:case_response_evidence}
 {\centering
\begin{tabularx}{\linewidth}{@{}l >{\raggedright\arraybackslash}X r r r@{}}
\toprule
Case & Quantity & Ref. & SFT & GRPO \\
\midrule
RE2-018 & Top tension (kN) & 2,173.42 & --- & 2,173.42 \\
A2-B06 & Force at $22.5^\circ$ (kN) & $-1,026.96$ & $\simeq0$ & --- \\
 & Force at $67.5^\circ$ (kN) & $-622.85$ & $\simeq0$ & --- \\
 & Force at $112.5^\circ$ (kN) & $-622.85$ & $\simeq0$ & --- \\
 & Force at $157.5^\circ$ (kN) & $-1,026.96$ & $\simeq0$ & --- \\
C4-039 & First contact (s) & 36.0126 & 35.0825 & 36.0126 \\
 & Peak force (kN) & 75.5751 & 103.7909 & 75.5751 \\
 & Penetration (m) & 0.05918 & 0.09041 & 0.05918 \\
\bottomrule
\end{tabularx}
\par}
\end{table}

RE2-018 specifies a 2,220~m uniform steel catenary riser in 1,200~m water depth, with a hinged top at $(0,0,-15)$~m and an anchored end at horizontal position $(1,350,0)$~m. The outer and inner diameters are 0.324 and 0.273~m, Young's modulus is 210~GPa, steel density is 7.85~t/m$^3$, and the internal and external water density is 1.025~t/m$^3$. The target segment length is 5~m, the seabed normal stiffness is 1,000~kN/m$^3$, and there is no environmental loading or static tangential friction. The GRPO policy's reported top tension is checked against independent recomputation from its reloaded model together with the model-input and applicable physical checks. The SFT response fails syntax validation at line 63 and supplies no numerical result.

A2-B06 specifies a four-point mooring in 130.6~m water depth, with fairleads at the four sign combinations of $(\pm52,\pm13.6,-9)$~m and anchors at $(\pm460,\pm300,-130.6)$~m. Each line has total unstretched length 529.5~m, outer diameter 0.0766~m, mass 0.1134~t/m and nominal axial stiffness 753,600~kN. Its terminal short section occupies 3.25\% of the specified last segment and has 0.925 times the parent axial stiffness. The calculation evaluates the origin and four offsets of radius 12~m at directions $22.5^\circ$, $67.5^\circ$, $112.5^\circ$ and $157.5^\circ$. Evaluation uses a 5\% relative tolerance and checks the saved model inputs. The SFT program fails both the response comparison and the input check because the line-segment list is misassigned. The GRPO policy raises a tuple-unpacking \texttt{ValueError} and does not deliver a valid restoring-force series.

For C4-FC1000-seed\_02-039, the payload dimensions, mass, volume and lifting-point separation match the C4 seed family, while the available payout is 5~m, the landing payout is 2~m and the reserved length is 2~m. The seabed normal/shear stiffness is 400/100~kN/m$^3$ and its friction coefficient is 0.60. The task prescribes a 0.005~s implicit time step, geometric zero-crossing for first contact, confirmation by contact force over 0.05~s, and a two-second post-contact peak window. The stored comparisons apply an absolute tolerance of 0.010~s to first-contact times; the peak contact force and penetration comparisons use the applicable 5\% relative rule. These tolerances are defined by the task specification.

All numerical comparisons in Table~\ref{tab:case_response_evidence} and Figure~\ref{fig:engineering_cases} are extracted directly from the saved native \texttt{.dat}/\texttt{.sim} simulation files and structured output summaries generated during benchmark evaluation.

\bibliographystyle{plainnat}
\bibliography{references}

\fi
\end{document}